\documentclass{article}

\usepackage{microtype}
\usepackage{graphicx}
\usepackage{subfigure}
\usepackage{booktabs} % for professional tables
\usepackage[dvipsnames]{xcolor}
\usepackage{tcolorbox}

\usepackage{wrapfig}

\usepackage{hyperref}

\usepackage[accepted]{icml2024}

\usepackage{amsmath}
\usepackage{amssymb}
\usepackage{mathtools}
\usepackage{amsthm}

\usepackage{soul}

\usepackage[capitalize,noabbrev]{cleveref}

\theoremstyle{plain}

\theoremstyle{definition}

\theoremstyle{remark}

\usepackage[textsize=tiny]{todonotes}

\hypersetup{colorlinks,linkcolor={blue},citecolor={magenta},urlcolor={blue}}

\icmltitlerunning{In {value-based} deep reinforcement learning, a pruned network is a good network}

\begin{document}

\twocolumn[
\icmltitle{In {value-based} deep reinforcement learning,\\a pruned network is a good network}

% It is OKAY to include author information, even for blind
% submissions: the style file will automatically remove it for you
% unless you've provided the [accepted] option to the icml2024
% package.

% List of affiliations: The first argument should be a (short)
% identifier you will use later to specify author affiliations
% Academic affiliations should list Department, University, City, Region, Country
% Industry affiliations should list Company, City, Region, Country

% You can specify symbols, otherwise they are numbered in order.
% Ideally, you should not use this facility. Affiliations will be numbered
% in order of appearance and this is the preferred way.
\icmlsetsymbol{equal}{*}

\begin{icmlauthorlist}
\icmlauthor{Johan Obando-Ceron}{dm,mila,udem}
\icmlauthor{Aaron Courville}{mila,udem}
\icmlauthor{Pablo Samuel Castro}{dm,mila,udem}
\end{icmlauthorlist}

\icmlaffiliation{mila}{Mila - Québec AI Institute}
\icmlaffiliation{udem}{Université de Montréal}
\icmlaffiliation{dm}{Google DeepMind}

\icmlcorrespondingauthor{Johan Obando-Ceron}{jobando0730@gmail.com}
\icmlcorrespondingauthor{Pablo Samuel Castro}{psc@google.com}

% You may provide any keywords that you
% find helpful for describing your paper; these are used to populate
% the "keywords" metadata in the PDF but will not be shown in the document
\icmlkeywords{Machine Learning, ICML}

\vskip 0.3in
]

% this must go after the closing bracket ] following \twocolumn[ ...

% This command actually creates the footnote in the first column
% listing the affiliations and the copyright notice.
% The command takes one argument, which is text to display at the start of the footnote.
% The \icmlEqualContribution command is standard text for equal contribution.
% Remove it (just {}) if you do not need this facility.

%\printAffiliationsAndNotice{}  % leave blank if no need to mention equal contribution
\printAffiliationsAndNotice{\icmlEqualContribution} % otherwise use the standard text.

\begin{abstract}
Recent work has shown that deep reinforcement learning agents have difficulty in effectively using their network parameters. We leverage prior insights into the advantages of sparse training techniques and demonstrate that gradual magnitude pruning enables {value-based} agents to maximize parameter effectiveness. This results in networks that yield dramatic performance improvements over traditional networks, using only a small fraction of the full network parameters. \textbf{Our code is publicly available,} see \autoref{apen:code_availability} for details.
\end{abstract}

\section{Introduction}
\label{sec:introduction}

Despite successful examples of deep reinforcement learning (RL) being applied to real-world problems \citep{mnih2015humanlevel, berner2019dota, vinyals2019grandmaster, fawzi2022discovering, Bellemare2020AutonomousNO}, there is growing evidence of challenges and pathologies arising when training these networks \citep{ostrovski2021tandem,kumar2021implicit, lyle2022understanding,graesser2022state,nikishin22primacy,sokar2023dormant,ceron2023small}. In particular, it has been shown that deep RL agents {\em under-utilize} their network's parameters: \citet{kumar2021implicit} demonstrated that there is an implicit underparameterization, \citet{sokar2023dormant} revealed that a large number of neurons go dormant during training, and \citet{graesser2022state} showed that sparse training methods can maintain performance with a very small fraction of the original network parameters.

One of the most surprising findings of this last work is that applying the gradual magnitude pruning technique proposed by \citet{zhu2017prune} on DQN \citep{mnih2015humanlevel} with a ResNet backbone (as introduced in Impala \citep{espeholt2018impala}), results in a 50\% performance improvement over the dense counterpart, with only 10\% of the original parameters (see the bottom right panel of Figure 1 of \citet{graesser2022state}). Curiously, when the same pruning technique is applied to the original CNN architecture there are no performance improvements, but no degradation either.

\begin{figure}[!t]
    \centering
    \includegraphics[width=0.47\textwidth]{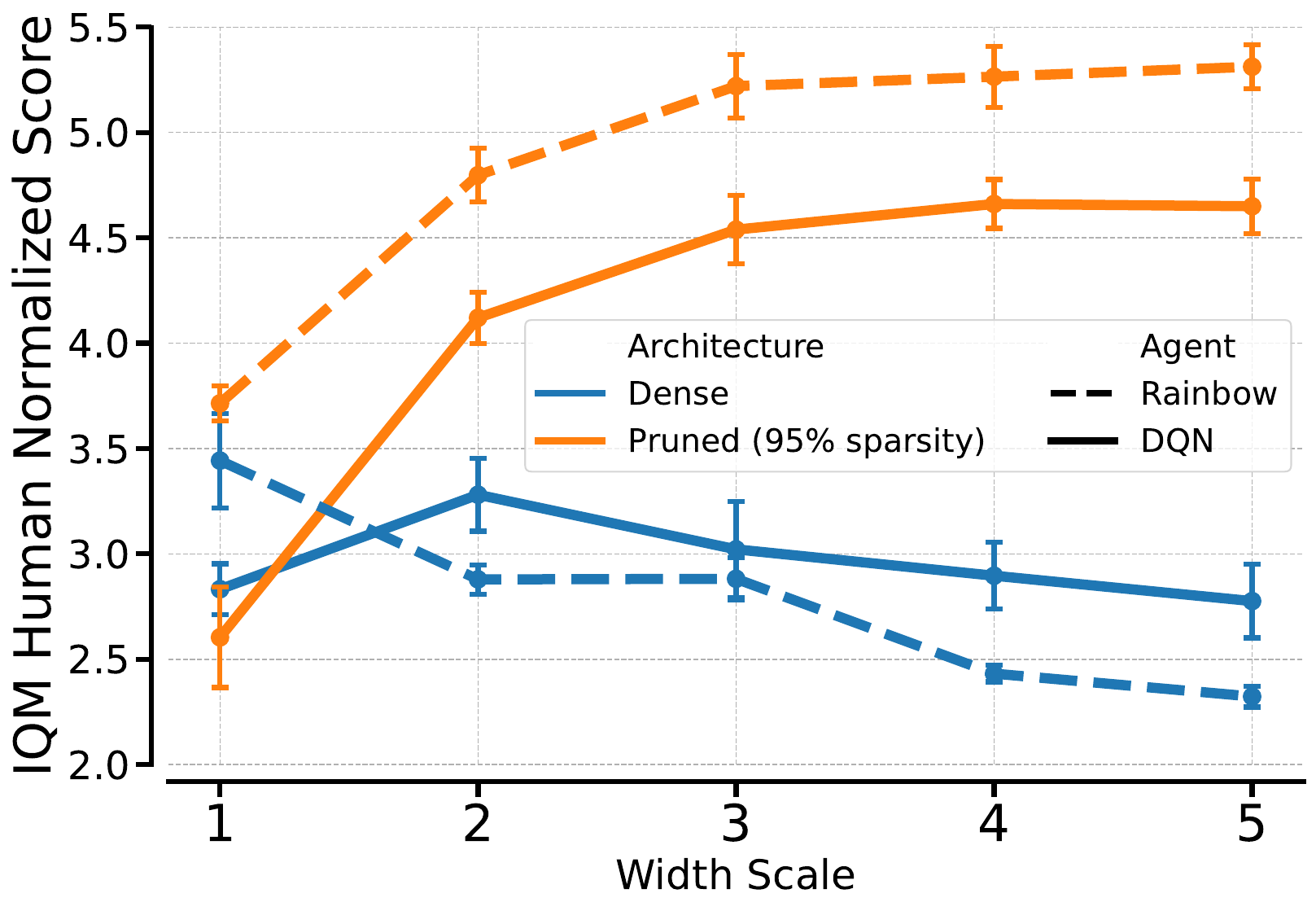}%
    \caption{\textbf{Scaling network widths for ResNet architecture}, for DQN and Rainbow with an Impala-based ResNet \citep{espeholt2018impala}. We report the interquantile mean after 40 million environment steps, aggregated over 15 games with 5 seeds each; error bars indicate 95\% stratified bootstrap confidence intervals. Replay ratio is fixed to the standard $0.25$.
    The default network is \textit{Dense}, which we indicate with a \textbf{{\color{NavyBlue} blue}} color in all the plots, for clarity. }
    \label{fig:topline}
    \vspace{-0.2cm}
\end{figure}

That the same pruning technique can have such qualitatively different, yet non-negative, results by simply changing the underlying architecture is interesting. It suggests that training deep RL agents with non-standard network topologies (as induced by techniques such as gradual magnitude pruning) may be generally useful, and warrants a more profound investigation.

In this paper we explore gradual magnitude pruning as a general technique for improving the performance of RL agents. We demonstrate that in addition to improving the performance of standard network architectures {for value-based agents}, the gains increase proportionally with the size of the base network architecture. This last point is significant, as deep RL networks are known to struggle with scaling architectures \citep{ota2021training,farebrother2023protovalue, taiga2023investigating, schwarzer2023bigger}.

\begin{tcolorbox}[colback=gray!10,
leftrule=0.5mm,top=0mm,bottom=0mm]
Our main contributions are as follows. We:
\begin{itemize}
    \item present gradual magnitude pruning as a general technique for maximizing parameter efficiency in {value-based} RL;
    \item demonstrate that networks trained with this technique produce stronger agents than their dense counterparts, and {continue improving} as we scale the size of the initial network; 
    \item {investigate} this technique across a varied set of agents and training regimes, {including actor-critic methods};
    \item present in-depth analyses to better understand the reasons behind their benefits.
\end{itemize}
\end{tcolorbox}

\section{Related Work}
\label{sec:related_work}

\paragraph{Scaling in Deep RL} Deep neural networks have been the driving factor behind many of the successful applications of reinforcement learning to real-world tasks. However, it has been historically difficult to scale these networks, in a manner similar to what has led to the ``scaling laws'' in supervised learning, without performance degradation; this is due in large part to exacerbated training instabilities that are endemic to reinforcement learning \citep{van2018deep, sinha2020d2rl, ota2021training}. Recent works that have been able to do so successfully have had to rely on a number of targeted techniques and careful hyper-parameter selection \citep{farebrother2023protovalue,taiga2023investigating,schwarzer2023bigger,ceron2023small}.

\citet{cobbe2020leveraging,farebrother2023protovalue} and \citet{schwarzer2023bigger} switched from the original CNN architecture of \citet{mnih2015humanlevel} to a ResNet based architecture, as proposed by \citet{espeholt2018impala}, which proved to be more amenable to scaling.  \citet{cobbe2020leveraging} and \citet{farebrother2023protovalue} observe advantages when increasing the number of features in each layer of the ResNet architecture.
\citet{schwarzer2023bigger} show that the performance of their agent (BBF) continues to grow proportionally with the width of their network. 
\citet{bjorck2021towards} propose spectral normalization to mitigate training instabilities and enable scaling of their architectures. 
\citet{ceron2023small} propose reducing batch sizes for improved performance, even when scaling networks.

\citet{obando2024mixtures} demonstrate that while parameter scaling with convolutional networks hurts single-task RL performance on Atari, incorporating Mixture-of-Expert (MoE) modules in such networks improves performance. \citet{farebrother24classification}  demonstrate that value functions trained with categorical cross-entropy significantly
improves performance and scalability in a variety of domains. 

\paragraph{Sparse Models in Deep RL} 

Previous studies \citep{schmitt2018kickstarting, zhang2019accelerating} have employed knowledge distillation with static data to mitigate instability, resulting in small, but dense, agents. \citet{Livne_2020} introduced policy pruning and shrinking, utilizing iterative policy pruning similar to iterative magnitude pruning \citep{han2015deep}, to obtain a sparse DRL agent. The exploration of the lottery ticket hypothesis in DRL was initially undertaken by \citet{yu2019playing}, and later \citet{vischer2021lottery} demonstrated that a sparse winning ticket can also be identified through behavior cloning. 
\citet{sokar2021dynamic} proposed the use of structural evolution of network topology in DRL, achieving 50\% sparsity with no performance degradation. \citet{arnob2021single} introduced single-shot pruning for offline Reinforcement Learning. 

\citet{graesser2022state} discovered that pruning often yields improved results, and {\em dynamic} sparse training methods, where the sparse topology changes throughout training~\citep{mocanu2018scalable,evci2020rigging}, can significantly outperform {\em static} sparse training, where the sparse topology remains fixed throughout training. \citet{tan2023rlx} enhance the efficacy of dynamic sparse training through the introduction of a novel delayed multi-step temporal difference target mechanism and a dynamic-capacity replay buffer. \citet{grooteautomatic23} proposed an automatic noise filtering method, which uses the principles of dynamic sparse training for adjusting the network topology to focus on task-relevant features.

\paragraph{Overparameterization in Deep RL}
\citet{song2019observational} and \citet{zhang2018study} highlighted and the tendency of RL networks to overfit, while \citet{nikishin22primacy} and \citet{sokar2023dormant} demonstrated the prevalence of plasticity loss in RL networks, leading to a decline in final performance. %yarats2021mastering
%For addressing this issue,
Several strategies have been proposed to mitigate this, such as data augmentation \citep{yarats2021image,cetin2022stabilizing},
dropout \citep{gal2016dropout}, and layer and batch normalization \citep{ba2016layer,ioffe2015batch}. 
%Notably, 
\citet{hiraoka2021dropout} demonstrated the success of employing dropout and layer normalization in Soft Actor-Critic \citep{haarnoja2018soft}, while \citet{liu2020regularization} identified that applying $\ell_2$ weight regularization on actors can enhance both on- and off-policy algorithms.

\citet{nikishin22primacy} identify a tendency of networks to overfit to early data (the primacy bias), which can hinder subsequent learning, and propose periodic network re-initialization as a means to mitigate it.
Similarly, \citet{sokar2023dormant} proposed re-initializing {\em dormant neurons} to improve network plasticity, while \citet{nikishin2023deep} propose plasticity injection by temporarily freezing the current network and utilizing newly initialized weights to facilitate continuous learning. 
\section{Background}
\label{sec:background}

\begin{figure}[!t]
    \centering
    \includegraphics[width=0.4\textwidth]{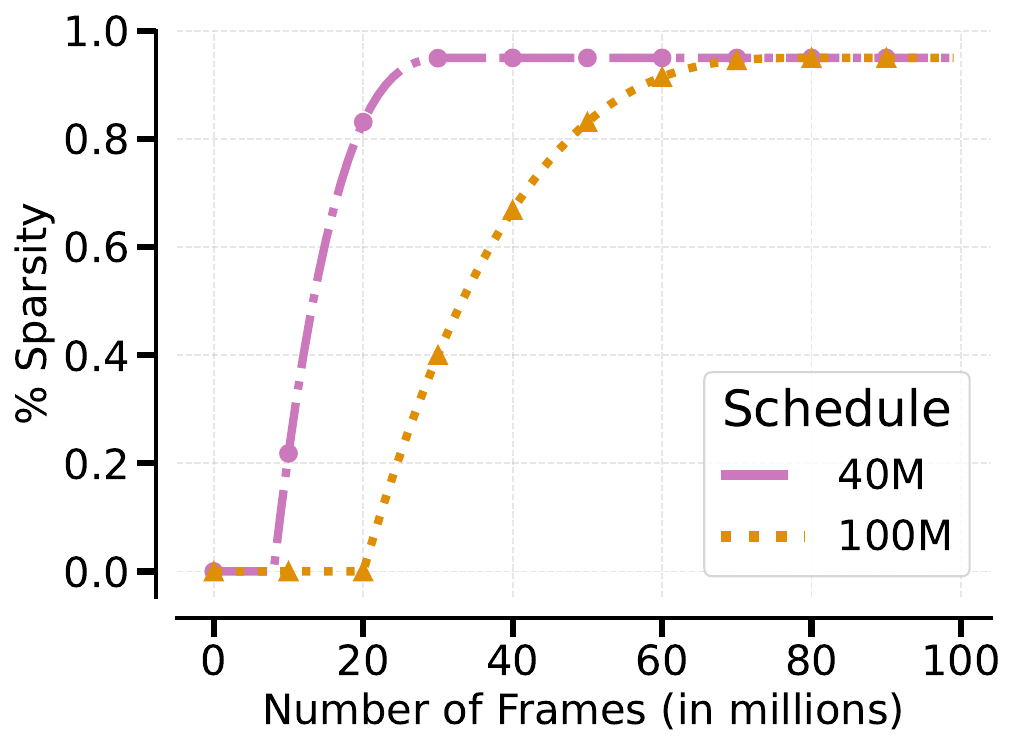}
    \vspace{-0.1cm}
    \caption{\textbf{Gradual magnitude pruning schedules} used in our experiments, to a target sparsity of 95\%, as specified in \autoref{eqn:polynomialSchedule}. Impact of varying pruning schedules, see \autoref{fig:varyingSchedules}.}
    \label{fig:pruningSchedule}
    %\vspace{-0.2cm}
\end{figure}

\paragraph{Deep reinforcement learning}
The goal in Reinforcement Learning is to optimize the cumulative discounted return over a long horizon, and is typically formulated as a Markov decision process (MDP) $(\mathcal{X}, \mathcal{A}, P, r, \gamma)$.
An MDP is comprised of a state space $\mathcal{X}$, an action space $\mathcal{A}$, a transition dynamics model $P:\mathcal{X}\times\mathcal{A}\rightarrow\Delta(\mathcal{X})$ (where $\Delta(X)$ is a distribution over a set $X$), a reward function $\mathcal{R}:\mathcal{X}\times\mathcal{A}\rightarrow\mathbb{R}$, and a discount factor  $\gamma \in[0,1)$. A policy $\pi:\mathcal{X}\rightarrow\Delta(\mathcal{A})$ formalizes an agent's behaviour.

For a policy $\pi$, $Q^\pi(\mathbf{x}, \mathbf{a})$ represents the expected discounted reward achieved by taking action $\mathbf{a}$ in state $\mathbf{x}$ and subsequently following the policy $\pi$: $Q^\pi(\mathbf{x}, \mathbf{a}):=\mathbb{E}_\pi\left[\sum_{t=0}^{\infty} \gamma^t \mathcal{R}\left(\mathbf{x}_t, \mathbf{a}_t\right) | \mathbf{x}_0 = x, \mathbf{a}_0 = a \right]$. The optimal Q-function, denoted as $Q^{\star}(\mathbf{x}, \mathbf{a})$, satisfies the Bellman recurrence \\
$Q^{\star}(\mathbf{x}, \mathbf{a})=\mathbb{E}_{\mathbf{x}^{\prime} \sim P\left(\mathbf{x}^{\prime} \mid \mathbf{x}, \mathbf{a}\right)}\left[\mathcal{R}(\mathbf{x}, \mathbf{a})+\gamma \max _{\mathbf{a}^{\prime}} Q^{\star}\left(\mathbf{x}^{\prime}, \mathbf{a}^{\prime}\right)\right]$.

Most modern value-based methods will approximate $Q$ via a neural network with parameters $\theta$, denoted as $Q_{\theta}$. This idea was introduced by \citet{mnih2015humanlevel} with their DQN agent, which has served as the basis for most modern deep RL algorithms. The network $Q_{\theta}$ is typically trained with a {\em temporal difference loss}, such as:
\begin{align*}
L(\theta) & =  \\
& \underset{\left(\mathbf{x}, \mathbf{a}, \mathbf{r}, \mathbf{x}^{\prime}\right) \sim \mathcal{D}}{\mathbb{E}}\left[\left(\mathbf{r} +\gamma \max_{\mathbf{a}^{\prime}\in\mathcal{A}}\bar{Q} \left(\mathbf{x}^{\prime}, \mathbf{a}^{\prime}\right)-Q_\theta(\mathbf{x}, \mathbf{a})\right)^2\right],
\end{align*}

Here $\mathcal{D}$ represents a stored collection of transitions $(\mathbf{x}_t, \mathbf{a}_t, \mathbf{r}_t, \mathbf{x}_{t+1})$ which the agent samples from for learning (known as the replay buffer). $\bar{Q}$ is a static network that infrequently copies its parameters from $Q_\theta$; its purpose is to produce stabler learning targets. 

Rainbow \citep{Hessel2018RainbowCI} extended, and improved, the original DQN algorithm with double Q-learning  \citep{hasselt2015doubledqn}, prioritized experience replay \citep{Schaul2016PrioritizedER}, dueling networks \citep{wang16dueling}, multi-step returns \citep{sutton88learning}, distributional reinforcement learning  \citep{Bellemare2017ADP}, and noisy networks \citep{fortunato18noisy}.

\begin{figure}[!t]
    \centering
    \includegraphics[width=0.39\textwidth]{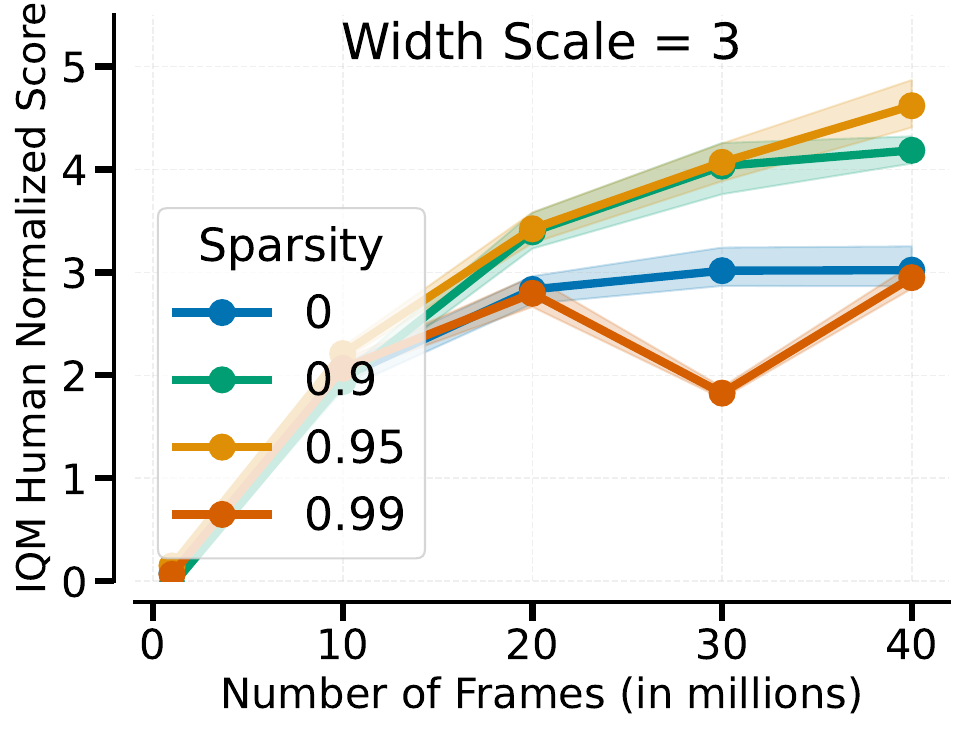}%
    \vspace{-0.1cm}
    \caption{\textbf{
    Evaluating how varying sparsity affects performance} for DQN with the ResNet architecture and a width multiplier of 3. See Section \ref{subsec:setup} for training details. 
    %We report results aggregated IQM of human-normalized scores over 15 games.
    }
    \label{fig:sparsity_values}
    %\vspace{-0.2cm}
\end{figure}
\paragraph{Gradual pruning}
In supervised learning settings there is a broad interest in sparse training techniques, whereby only a subset of the full network parameters are trained/used \citep{gale2019state}. This is motivated by computational and space efficiency, as well as speed of inference. \citet{zhu2017prune} proposed a polynomial schedule for gradually sparsifying a dense network over the course of training by {\em pruning} model parameters with low weight magnitudes.

Specifically, let $s_{F}\in [0, 1]$ denote the final desired sparsity level (e.g. $0.95$ in most of our experiments) and let $t_{\mathrm{start}}$ and $t_{\mathrm{end}}$ denote the start and end iterations of pruning, respectively; then the sparsity level at iteration $t$ is given by:
\begin{align}
    \label{eqn:polynomialSchedule}
    s_t & = & \\
    & s_{F} \left( 1 - \left( 1 - \frac{t - t_{\mathrm{start}}}{t_{\mathrm{end}} - t_{\mathrm{start}}} \right)^3 \right) & \textrm{if } t_{\mathrm{start}} \leq t \leq t_{\mathrm{end}} \notag \\
    & \qquad\qquad\qquad\qquad 0.0 & \textrm{if } t < t_{\mathrm{start}} \notag \\
    & \qquad\qquad\qquad\qquad s_F & \textrm{if } t > t_{\mathrm{end}} \notag
\end{align}

\citet{graesser2022state} applied this idea to deep RL networks, setting $t_{\mathrm{start}}$ at 20\% of training and $t_{\mathrm{end}}$ at 80\% of training.\\
\section{Pruning {can} boost deep RL performance}
\label{sec:experiments}

We investigate the general usefulness of gradual magnitude pruning in deep RL agents in both online and offline settings.

\begin{figure}[!t]
    \centering
    \includegraphics[width=0.475\textwidth]{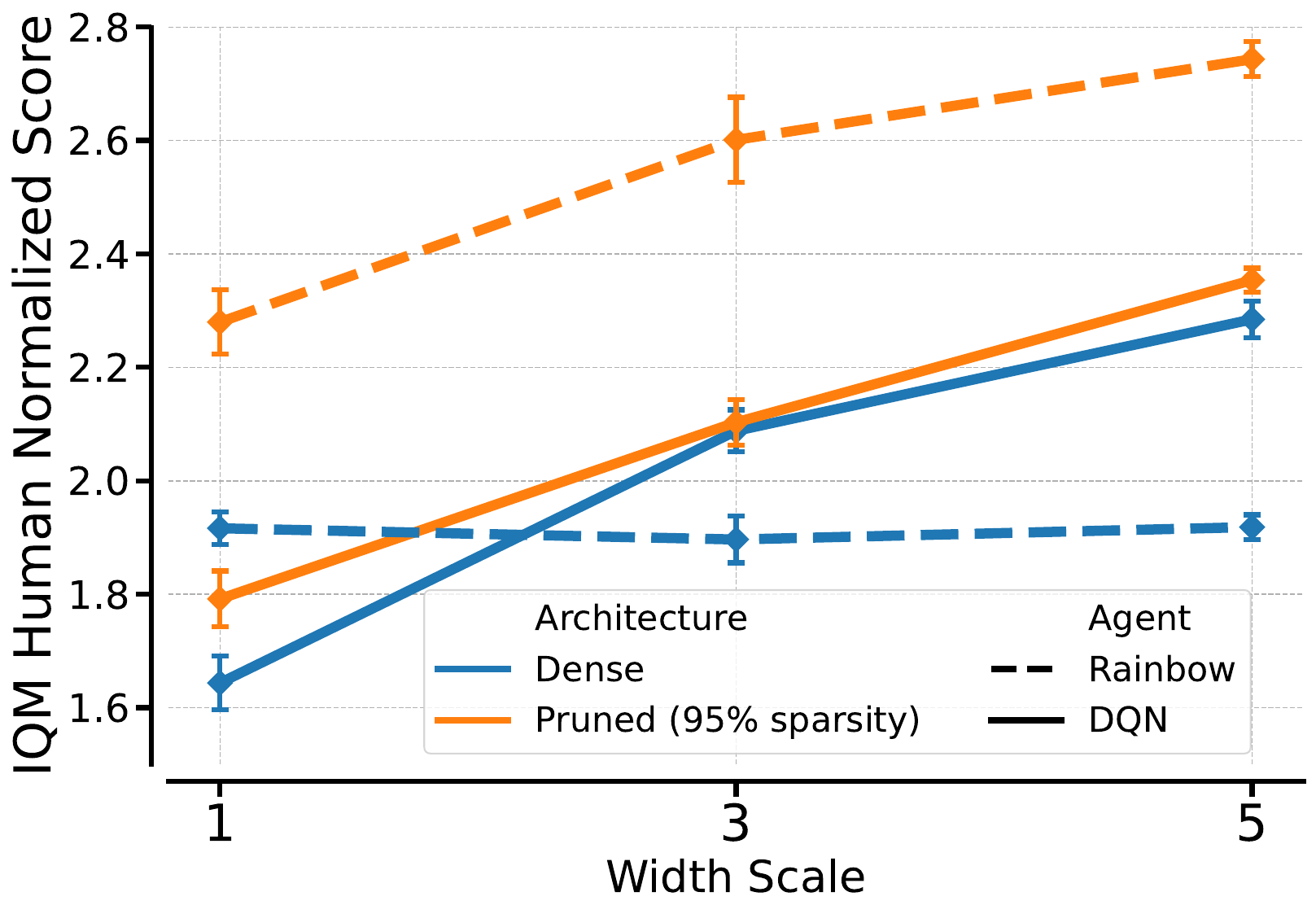}
    \vspace{-0.1cm}
    \caption{\textbf{Scaling network widths for the original CNN architecture of \citet{mnih2015humanlevel}}, for DQN \textbf{(left)} and Rainbow \textbf{(right)}. See Section \ref{subsec:setup} for training details. 
    }
    \label{fig:scalingWidths_cnn}
    \vspace{-0.2cm}
\end{figure}

\subsection{Implementation details}
\label{subsec:setup}

For the base DQN and Rainbow agents we use the Jax implementations of the Dopamine library\footnote{Dopamine code available at  \url{https://github.com/google/dopamine}.} \citep{castro2018dopamine} with their default values. It is worth noting that Dopamine provides a ``compact'' version of the original Rainbow agent, using only multi-step updates, prioritized replay, and distributional RL. For all experiments we use the Impala architecture introduced by \citet{espeholt2018impala}, which is a 15-layer ResNet,  {unless} otherwise specified. Our reasoning for this is not only because of the results from \citet{graesser2022state}, but also due to a number of recent works demonstrating that this architecture results in generally improved performance  \citep{schmidt2021fast, kumar2022offline,schwarzer2023bigger}.

We use the JaxPruner\footnote{JaxPruner code available at  \url{https://github.com/google-research/jaxpruner}.} \citep{lee2024jaxpruner} library for gradual magnitude pruning, as it already provides integration with Dopamine. We follow the schedule of \citet{graesser2022state}: begin pruning the network 20\% into training and stop at 80\%, keeping the final sparse network fixed for the rest of training. \autoref{fig:pruningSchedule} illustrates the pruning schedules used in our experiments (for 95\% sparsity). We evaluate our agents on the Arcade Learning Environment (ALE) \citep{Bellemare_2013} on the same 15 games used by \citet{graesser2022state}, chosen for their diversity\footnote{Discussed in A.4 in \citet{graesser2022state}.}. For computational considerations, most experiments were conducted over 40 million frames (as opposed to the standard 200 million); in our investigations we found the qualitative differences between algorithms at 40 million frames to be mostly consistent with those at 100 million (e.g. see \autoref{fig:varyingSchedules}).

We follow the guidelines outlined by \citet{agarwal2021deep} for evaluation: each experiment was run with 5 independent seeds, and we report the human-normalized interquantile mean (IQM), aggregated across the 15 games, configurations, and seeds, along with 95\% stratified bootstrap confidence intervals. All experiments were run on NVIDIA Tesla P100 GPUs, and each took approximately 2 days to complete. {All hyper-parameters are listed in \cref{sec:list_hyperparameters}.}

\begin{figure}[!t]
    \centering
    \includegraphics[width=0.455\textwidth]{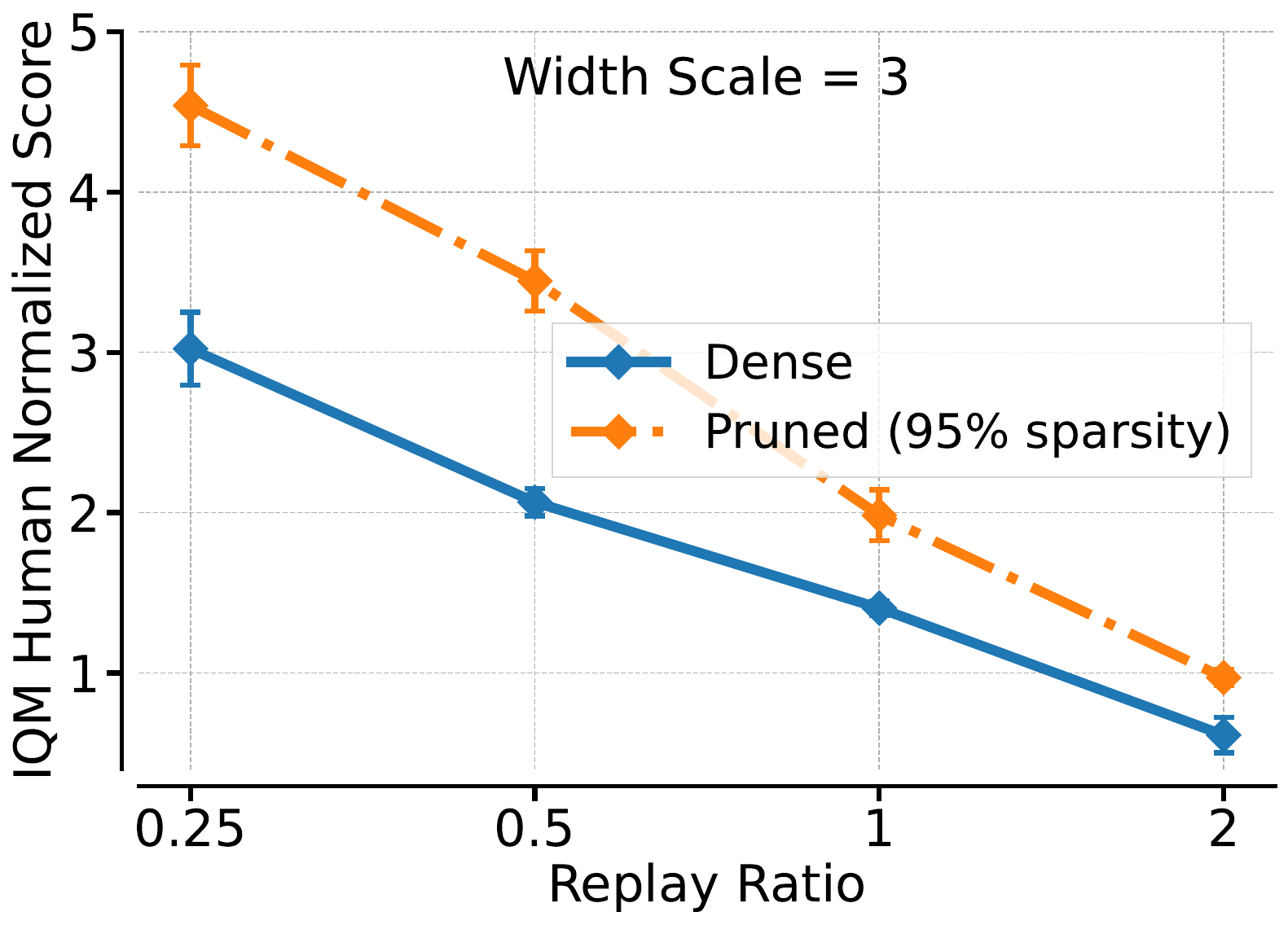}%
    \vspace{-0.1cm}
    \caption{\textbf{Scaling replay ratio for Rainbow with the ResNet architecture with a width multiplier of $3$.} 
    Default replaly ratio is $0.25$.
    See Section \ref{subsec:setup} for training details.}
    \label{fig:scalingReplayRatio}
    \vspace{-0.2cm}
\end{figure}

\subsection{Online RL}

While \citet{graesser2022state} demonstrates that sparse networks are capable of maintaining agent performance, if these levels of sparsity were too high, performance eventually degrades. This is intuitive, as with higher levels of sparsity, there are fewer active parameters left in the network. One natural question is whether {\em scaling} our initial network enables
high levels of sparsity. We thus begin our inquiry by applying gradual magnitude pruning on DQN with the Impala architecture, where we have scaled the convolutional layers by a factor of 3. \autoref{fig:sparsity_values} confirms that this is indeed the case: 90\% and 95\% sparsity produce a 33\% performance improvement, and 99\% sparsity maintains performance.

\begin{figure}[!t]
    \centering
    \includegraphics[width=0.247\textwidth]{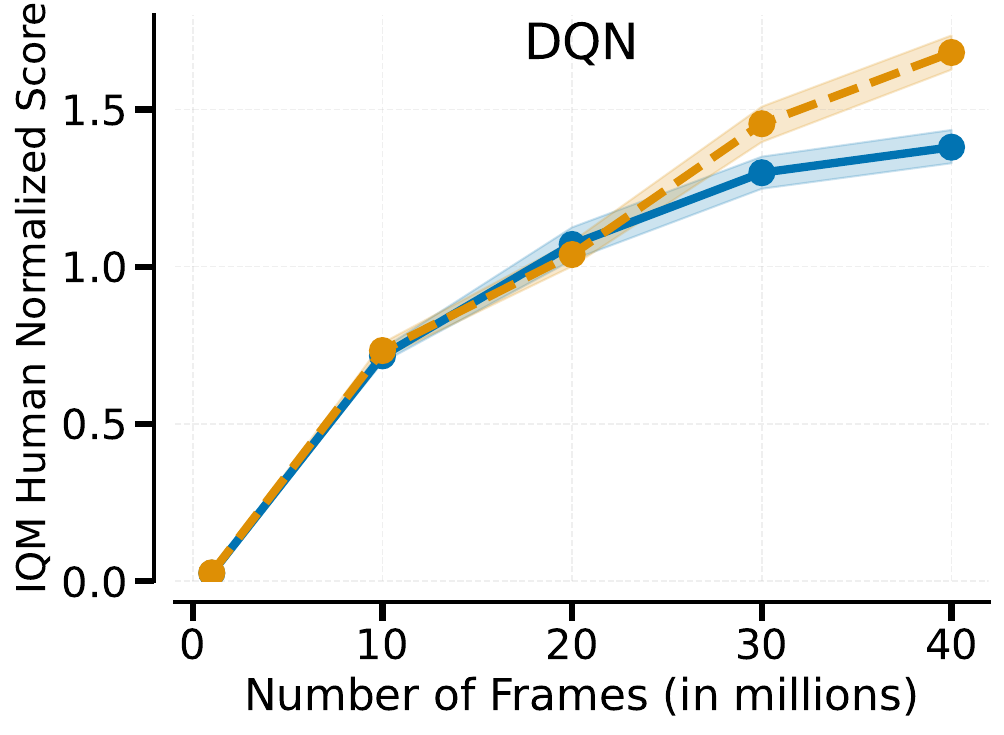}%
    \includegraphics[width=0.237\textwidth]{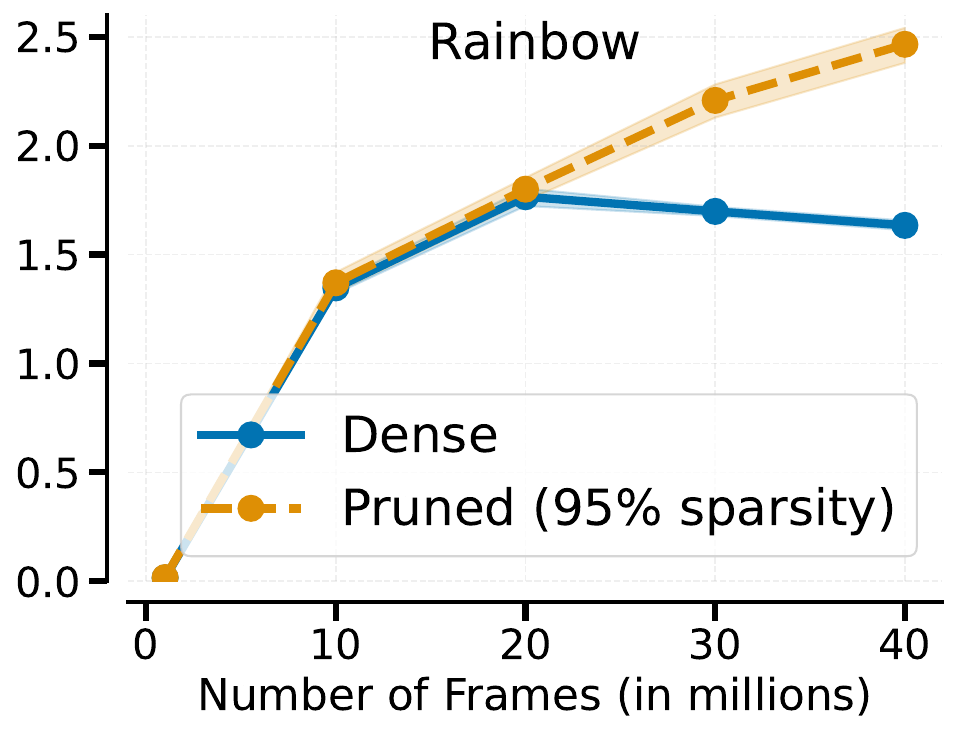}%
    \vspace{-0.1cm}
    \caption{\textbf{Evaluating performance on the full Atari 2600 suite.} DQN (left) and Rainbow (right), both using the ResNet architecture with a width of $3$. We report IQM performance with error bars indicating 95\% confidence interval. See Section \ref{subsec:setup} for training details.
    }
    \vspace{-0.1cm}
    \label{fig:allsuite}
\end{figure}

A sparsity of 95\% consistently yielded the best performance in our initial explorations, so we primarily focus on this sparsity level for our investigations. \autoref{fig:topline} is a striking result: we observe close to a 60\% (DQN) and 50\% (Rainbow) performance improvement over the original (unpruned and unscaled) architectures. Additionally, while the performance of the unpruned architectures decreases monotonically with increasing widths, the performance of the pruned counterparts {\em increases} monotonically. In \autoref{fig:allsuite} we evaluated pruning on DQN and Rainbow over all 60 Atari 2600 games, confirming our findings are not specific to the 15 games initially selected.

When switching both agents to using the original CNN architecture of \citet{mnih2015humanlevel} we see a similar trend with Rainbow, but see little improvement in DQN (\autoref{fig:scalingWidths_cnn}). This result is consistent with the findings of \citet{graesser2022state}, where no real improvements were observed with pruning on CNN architectures. An interesting observation is that, with this CNN architecture, the performance of DQN does seem to benefit from increased width, while the performance of Rainbow suffer from slight degradation.

{When evaluating on more modern value-based agents, specifically IQN \citep{dabney2018implicit} and Munchausen-IQN \citep{vieillard2020munchausen}, we observe the same advantages arising from pruning (see \cref{sec:iqn}).}

Our findings thus far suggest that the use of gradual magnitude pruning increases the parameter efficiency of these agents. If so, then these sparse networks should also be able to benefit from more gradient updates. The {\em replay ratio}\footnote{In the hyperparameters established in \cite{mnih2015humanlevel}, the policy is updated every 4 environment steps collected, resulting in a replay ratio of 0.25.}, which is the number of gradient updates per environment step, measures exactly this; it is well-known that it is difficult to increase this value without performance degradation \citep{fedus2020revisiting, nikishin22primacy,schwarzer2023bigger,doro2023sampleefficient}. 

In \autoref{fig:scalingReplayRatio} we can indeed confirm that the pruned architectures maintain a performance lead over the unpruned baseline even at high replay ratio values. {The sharper rate of decline with pruning may suggest that the pruning schedule needs to be re-tuned for these settings.}

\subsection{Low data regime}
\label{secc:samplefficient}
\citet{Kaiser2020Model} introduced the Atari 100k benchmark to evaluate RL agents in a sample-constrained setting, allowing agents only 100k\footnote{Here, 100k refers to agent steps, or 400k environment frames, due to skipping frames in the standard training setup.} environment interactions. For this regime, \citet{kostrikov2020image} introduced DrQ, a variant of DQN which makes use of data augmentation; the hyperparameters for this agent were further optimized by \citet{agarwal2021deep} in DrQ($\epsilon$).
Similarly, \citet{van2019use} introduced Data-Efficient Rainbow (DER), which optimized the hyperparameters of Rainbow \citep{Hessel2018RainbowCI} for this low data regime.

\begin{figure}[!t]
    \centering
    \includegraphics[width=0.247\textwidth]{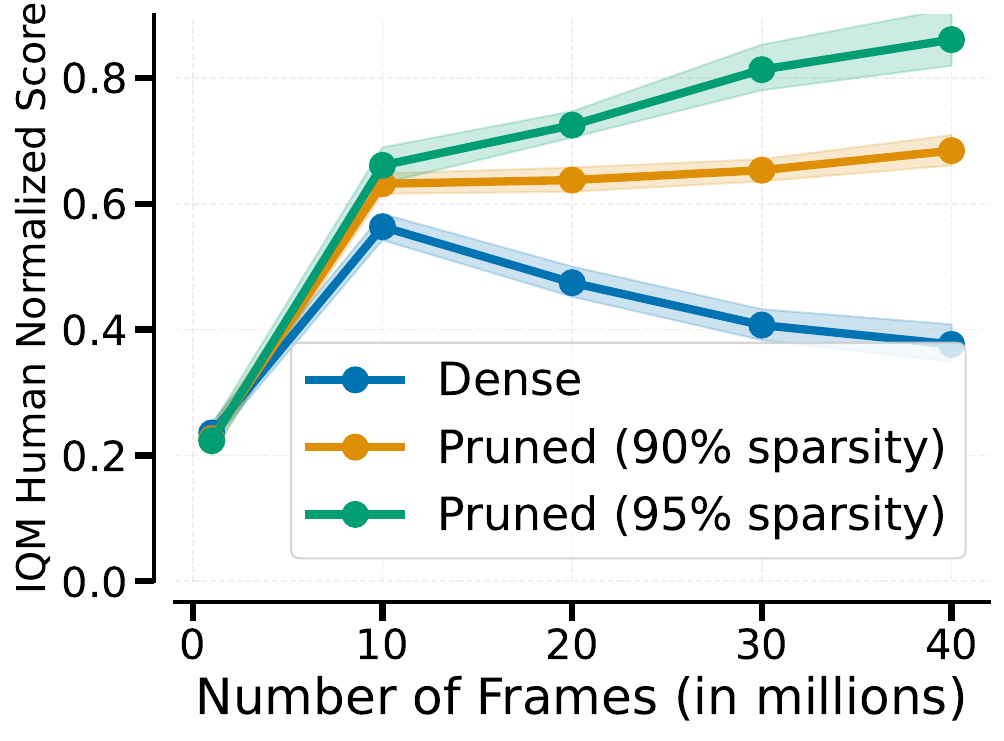}%
    \includegraphics[width=0.237\textwidth]{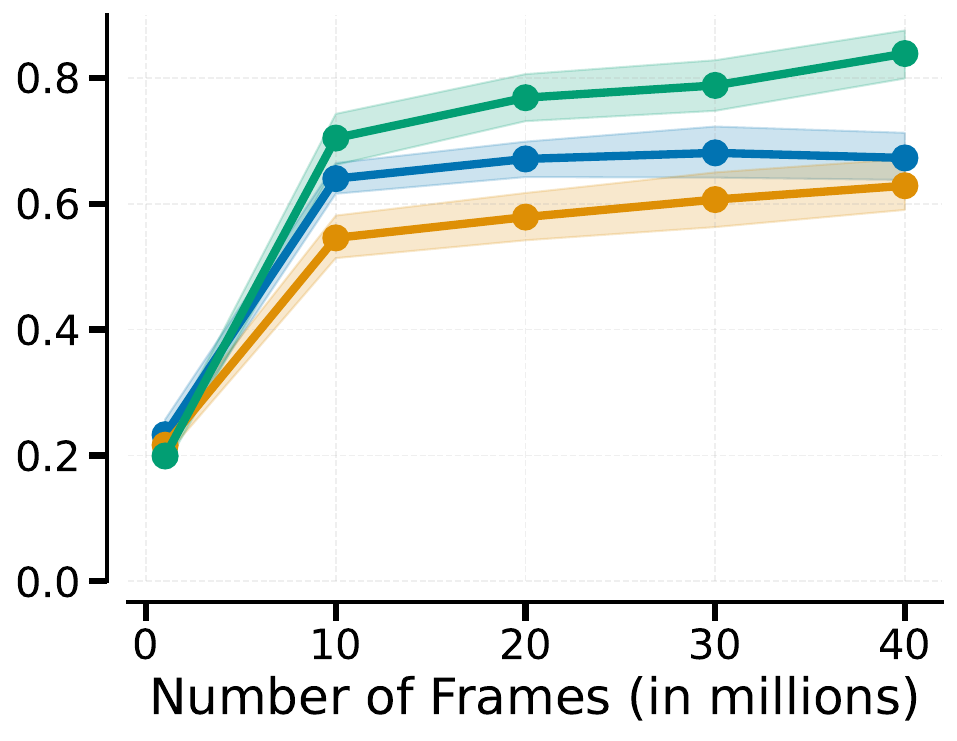}%
    \vspace{-0.2cm}
    \caption{\textbf{Performance of DrQ($\epsilon$) (left) and DER (right) when trained for 40M frames}. Both agents use a ResNet architecture with a width multiplier of $3$. See Section \ref{subsec:setup} for training details.}
    \vspace{-0.2cm}
    \label{fig:samplefficient_algo}
\end{figure}

When evaluated on this low data regime, our pruned agents demonstrated no gains.
%[\johan: Add 100k results in the appendix]). 
However, when we ran for 40M environment interactions (as suggested by \citet{ceron2023small}), we do observe significant gains when using gradual magnitude pruning, as shown in \autoref{fig:samplefficient_algo}. Interestingly, In DrQ($\epsilon$) the pruned agents avoid the performance degradation affecting the baseline when trained for longer.

\subsection{Offline RL}
\label{sec:offlineRL}

Offline reinforcement learning focuses on training an agent solely from a fixed dataset of samples without any environment interactions. We used two recent state of the art methods from the literature: CQL \citep{kumar2020conservative} and CQL+C51 \citep{kumar2022offline}, both with the ResNet architecture from \citet{espeholt2018impala}. Following \citet{kumar2021dr3}, we trained  these agents on 17 Atari games for 200 million frames iterations, where  where  1  iteration corresponds to 62,500 gradient updates. We assessed the agents by considering a dataset composed of a random 5\% sample of all the environment interactions collected by a DQN agent trained for 200M environment steps \citep{agarwal2020optimistic}.

\begin{figure*}[!t]
    \centering
    \includegraphics[width=0.259\textwidth]{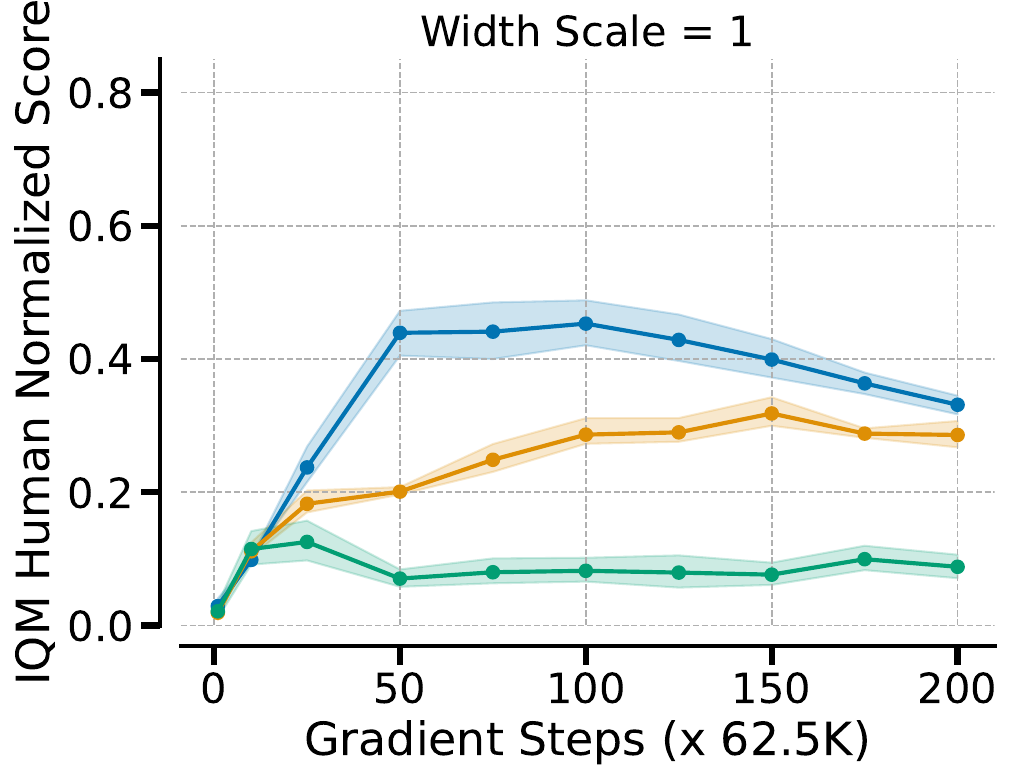}%
    \includegraphics[width=0.247\textwidth]{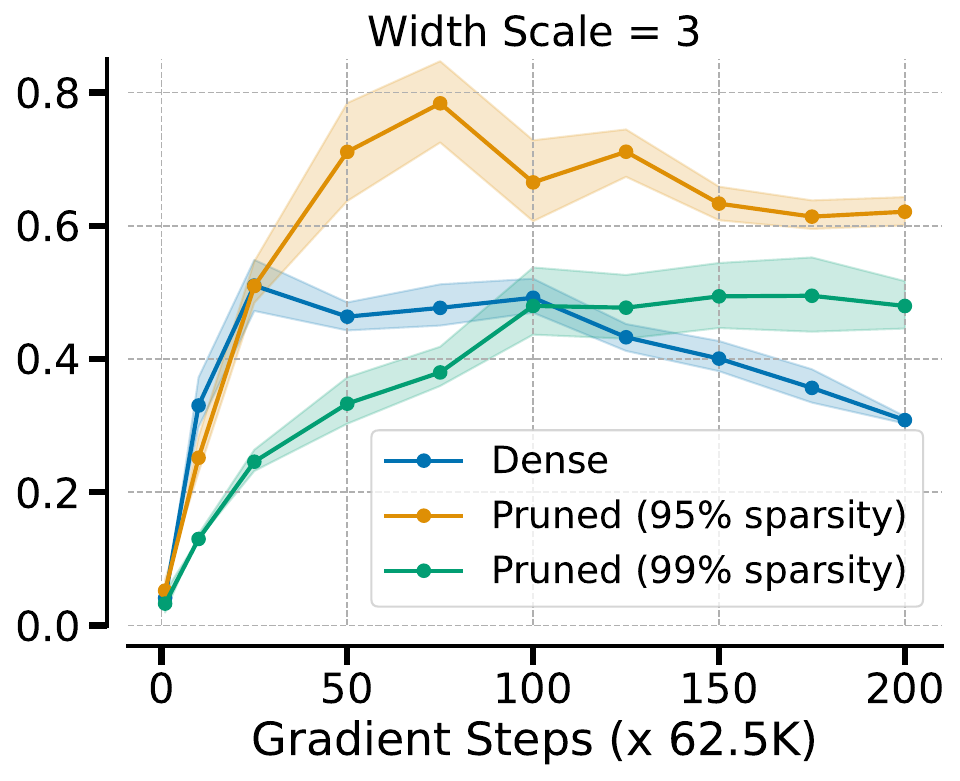}%
    \vline height 98pt depth 0 pt width 1 pt
    \includegraphics[width=0.247\textwidth]{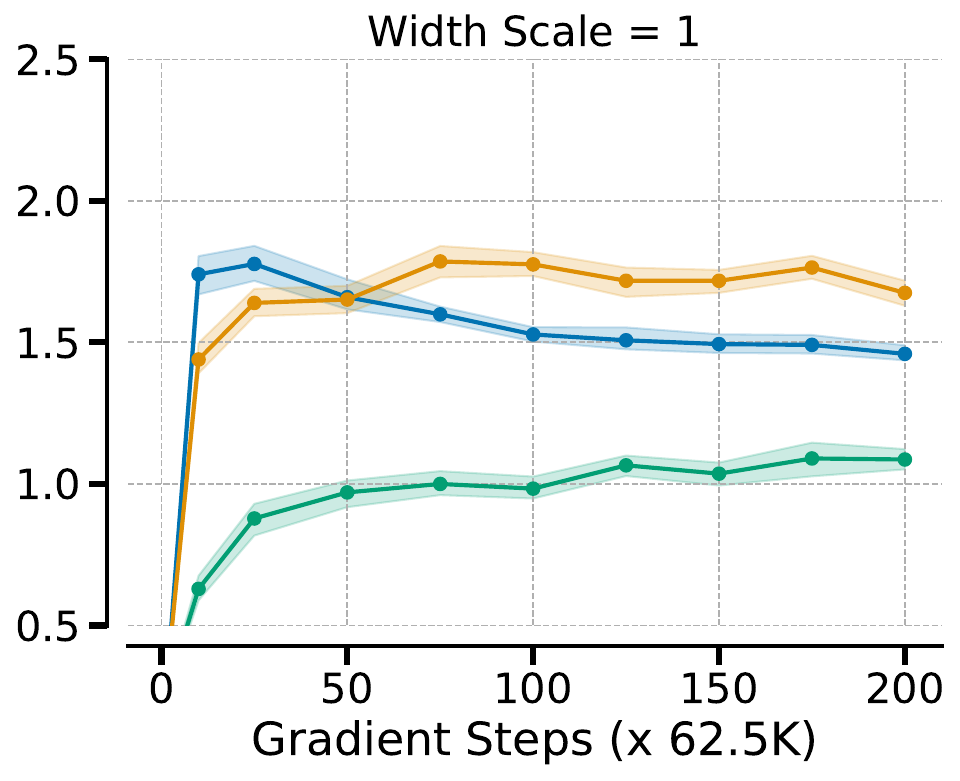}%
    \includegraphics[width=0.247\textwidth]{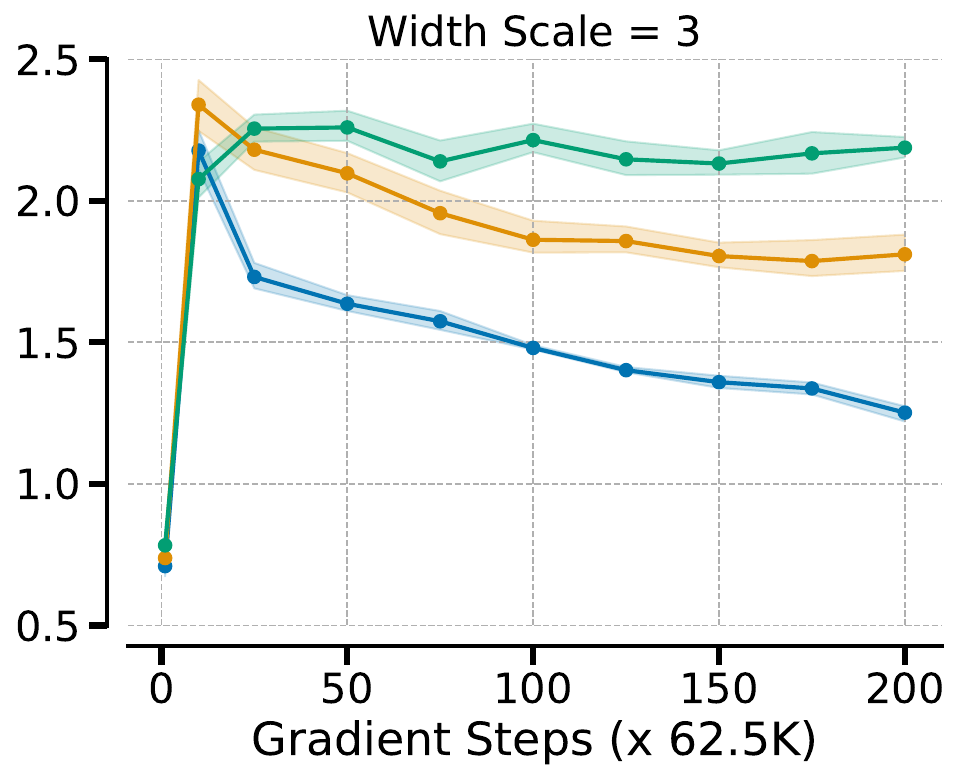}%
    %\vspace{-0.1cm}
    \caption{\textbf{Scaling network widths for offline agents} CQL \textbf{(left)} and CQL+C51 \textbf{(right)}, both using the ResNet architecture. We report interquantile mean performance with error bars indicating 95\% confidence intervals across 17 Atari games. x-axis represents gradient steps; no new data is collected.}
    \label{fig:scalingWidths_offline}
    %\vspace{-0.2cm}
\end{figure*}

\begin{figure*}[!t]\centering
   \begin{minipage}{0.59\textwidth}
    \includegraphics[width=\textwidth]{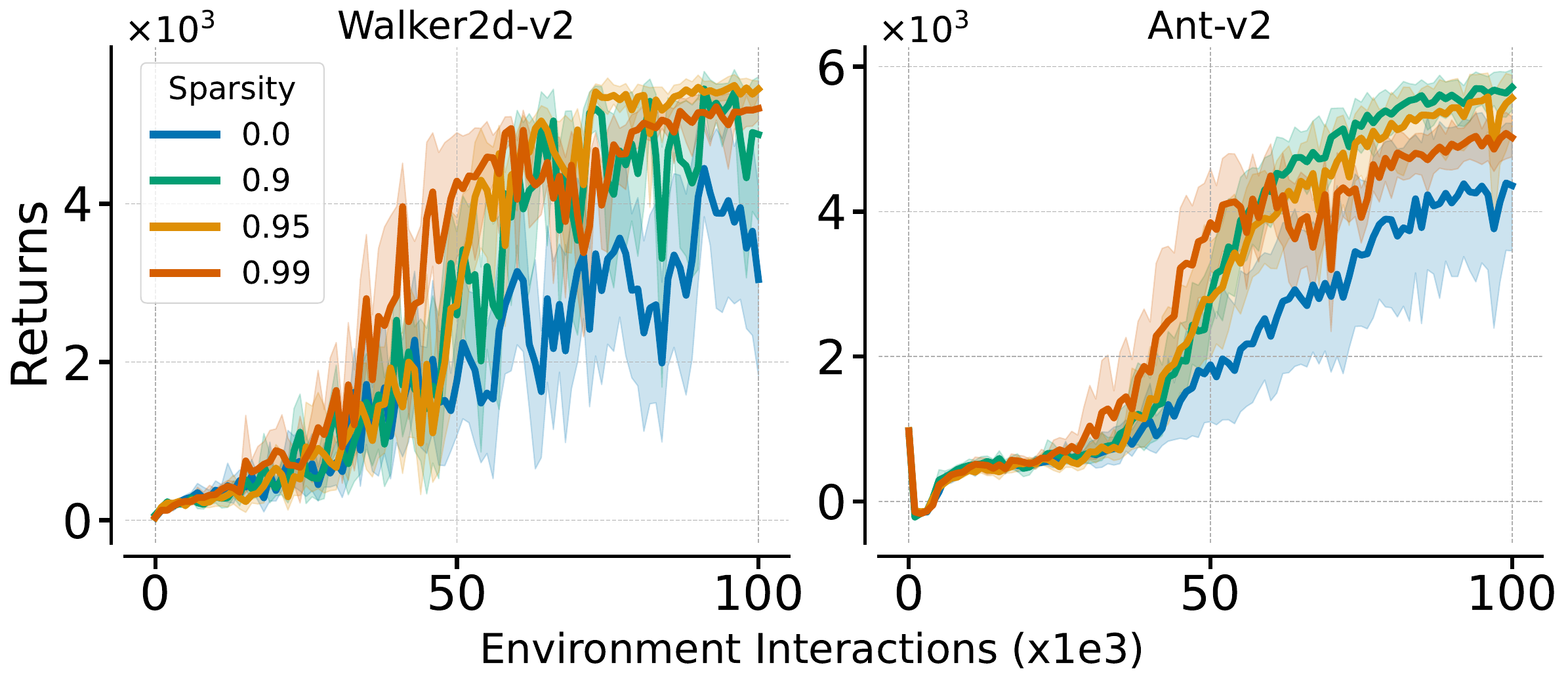}
     \vspace{-0.5cm}
     \caption{\textbf{Evaluating how varying the sparsity parameter affects performance of SAC} on two MuJoCo environments when increasing width x5. We report returns over 10 runs for each experiment. 
     }
    \label{fig:sacResults}
   \end{minipage}
    \hspace{0.6cm}
   \begin {minipage}{0.33\textwidth}
     \includegraphics[width=\textwidth]{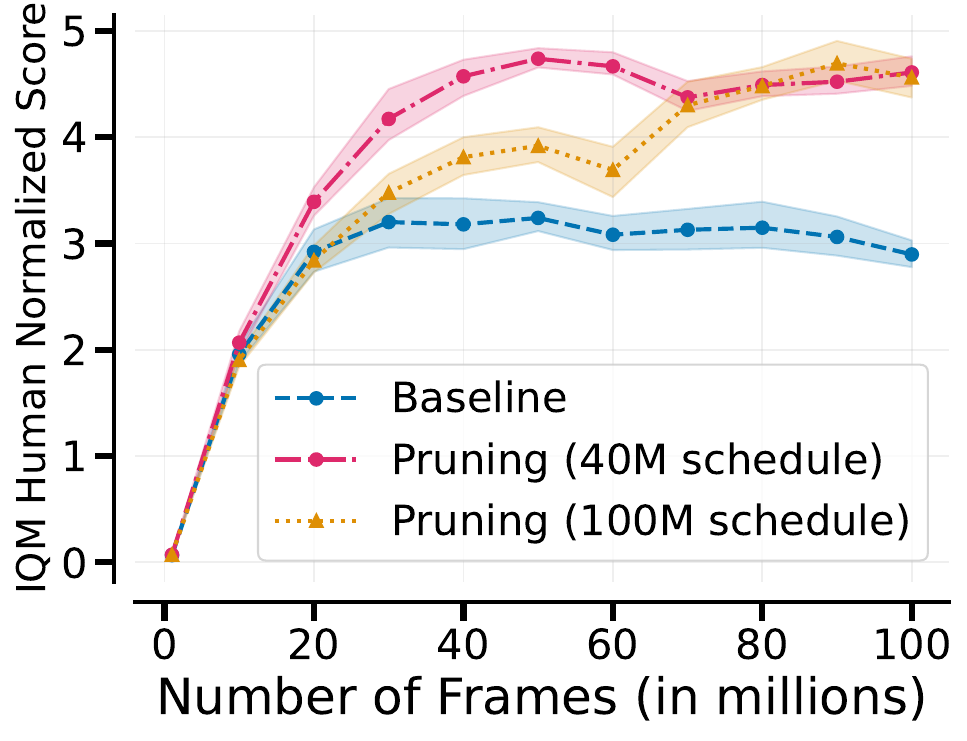}
     \vspace{-0.5cm}
     \caption{\textbf{Impact of varying pruning schedules}, for DQN with an Impala-based ResNet with a width multiplier of $3$.}
     \label{fig:varyingSchedules}
   \end{minipage}
\end{figure*}

Note that since we are training for a different number of steps than our previous experiments, we adjust the pruning schedule accordingly. As shown in  \autoref{fig:scalingWidths_offline}, both CQL and CQL+C51 observe significant gains when using pruned networks, in particular with wider networks. Interestingly, in the offline regime, pruning also helps to avoid performance collapse when using a shallow network (width scale equal to 1), or even improve final performance as in the case of CQL+C51.

\subsection{Actor-Critic methods}
\label{sec:sac}
\begin{figure*}[!t]
    \centering
   \includegraphics[width=\linewidth]{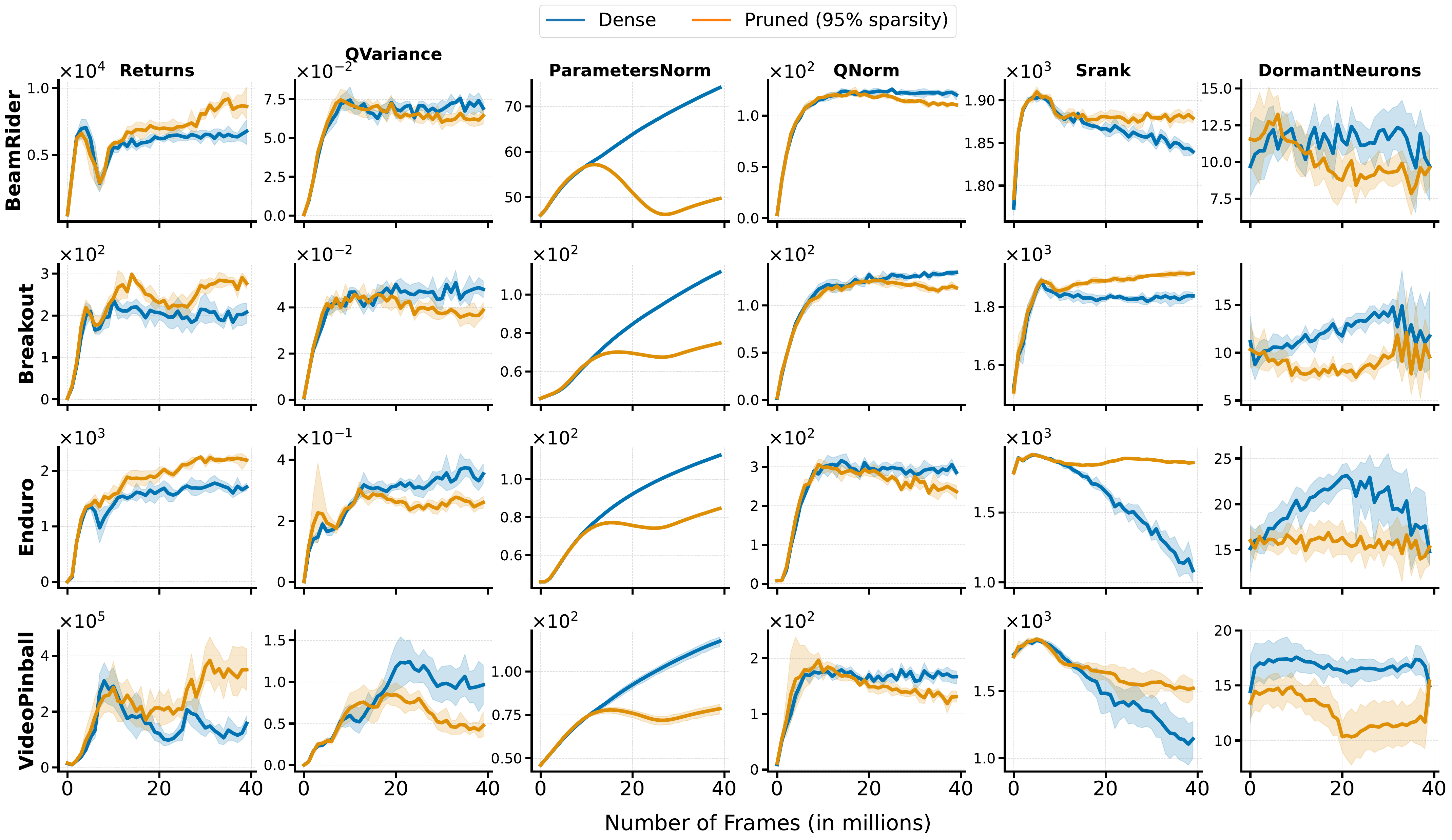}%

    \caption{\textbf{Empirical analyses for four representative games when applying pruning}. From left to right: training returns, average $Q$-target variance, average parameters norm, average $Q$-estimation norm, $srank$ \citep{kumar2021implicit}, and dormant neurons \citep{sokar2023dormant}. All results averaged over 5 seeds, shaded areas represent 95\% confidence intervals.}%
    \label{fig:aggregateAnalysesFourGames}%
    \vspace{1em}
    %\vspace{0.3cm}
\end{figure*}
Our investigation thus far has focused on value-based methods. Here we investigate if gradual magnitude pruning can yield performance gains for Soft Actor Critic \citep[SAC;][]{haarnoja2018soft}, a popular policy-gradient algorithm. 
We evaluated SAC on five continuous control environments from the MuJoCo suite \citep{todorov2012mujoco}, using 10 independent seeds for each.
In \autoref{fig:sacResults} we present the results for Walker2d-v2 and Ant-v2, where we see the advantages of gradual magnitude pruning persist; in the remaining three environments (see \autoref{appen:sac_threeEnv}) there are no real observable gains from pruning. {In \cref{sec:ppoExperiments} we see a similar trend with PPO \citep{schulman2017proximal}.}

\subsection{Stability of the pruned network}
\label{sec:stability}

We followed the pruning schedule proposed by \citet{graesser2022state}, which adapts naturally to differing training steps (as discussed above for the offline RL experiments). This schedule trains the final sparse network for only the final 20\% of training steps. A natural question is whether the resulting sparse network, when trained for longer, is still able to maintain its performance. To evaluate this, we trained DQN for 100 million frames and applied two pruning schedules: the regular schedule we would use for 100M as well as the schedule we would normally use for 40M training steps (see \autoref{fig:pruningSchedule}).

As \autoref{fig:varyingSchedules} shows, even with the compressed 40M schedule, the pruned network is able to maintains its strong performance. Interestingly, with the compressed schedule the agent achieves a higher performance {\em faster} than with the regular one. This suggests there is ample room for exploring alternate pruning schedules.

\subsection{Learning rate and Batch size scaling}
\label{sec:lr_batch}
%An important point to consider is that 
The default learning rate or batch size may not be optimal for large neural networks. The default learning for DQN is $6.25\times 10^{-5}$. We run experiments with a learning rate divided by the width scale factor (so $2.08\times 10^{-5}$ for a width factor of $3$, and $1.25\times 10^{-5}$ for a width factor of $5$). While these learning rates do improve the performance of the baseline, it is still surpassed by pruning (see \autoref{fig:learning_rate}). We observe a similar trend when evaluating different batch size values. The default batch size is 32 (for all value based agents used in this paper), and we ran experiments with batch sizes of 16 and 64. In all cases, pruning maintains its strong advantage (see \autoref{fig:batch_size}). These results are consistent with the thesis that \textit{pruning can serve as a drop-in mechanism for increasing agent performance}.

\section{Why is pruning so effective?}
\label{sec:analisis}

We focus our analyses on four games: BeamRider, Breakout, Enduro, and VideoPinball. For each, we measure the variance of the $Q$ estimates (QVariance); the average norm of the network parameters (ParametersNorm); the average norm of the $Q$-values (QNorm); the effective rank of the matrix (Srank) as suggested by \citet{kumar2021implicit}, and the fraction of dormant neurons as defined by \citet{sokar2023dormant}. 

We present our results in \autoref{fig:aggregateAnalysesFourGames}. What becomes evident from these figures is that pruning {\bf (i)} reduces variance, {\bf (ii)} reduces the norms of the parameters, {\bf (iii)} decreases the number of dormant neurons, and {\bf (iv)} increases the effective rank of the parameters. Some of these observations can be attributed to a form of normalization, whereas others may arise due to increased network plasticity. 

\citet{lyle2024disentangling} show that increased parameter norm accompanies plasticity loss in different neural architectures.
In \autoref{fig:aggregateAnalysesFourGames}, we observe a low parameter norm value when applying gradual pruning, which represent a high final performance return.

\subsection{Comparison to other methods}
In order to disentangle the impact of pruning from normalization and explicit plasticity injection, we compare against existing methods in the literature.

\begin{figure*}[!t]\centering
   \begin{minipage}{0.67\textwidth}
    \includegraphics[width=0.475\textwidth]{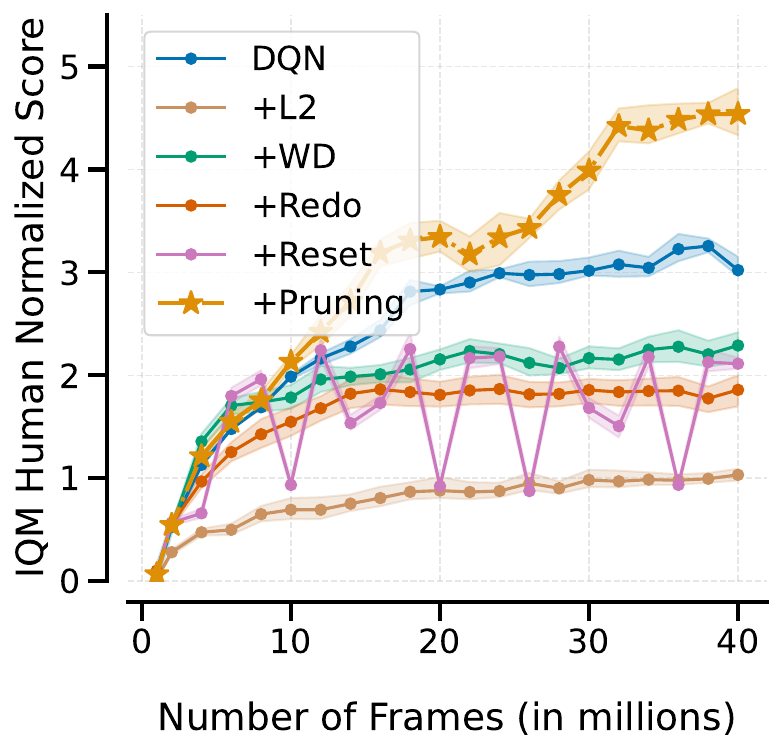}%
    \includegraphics[width=0.503\textwidth]{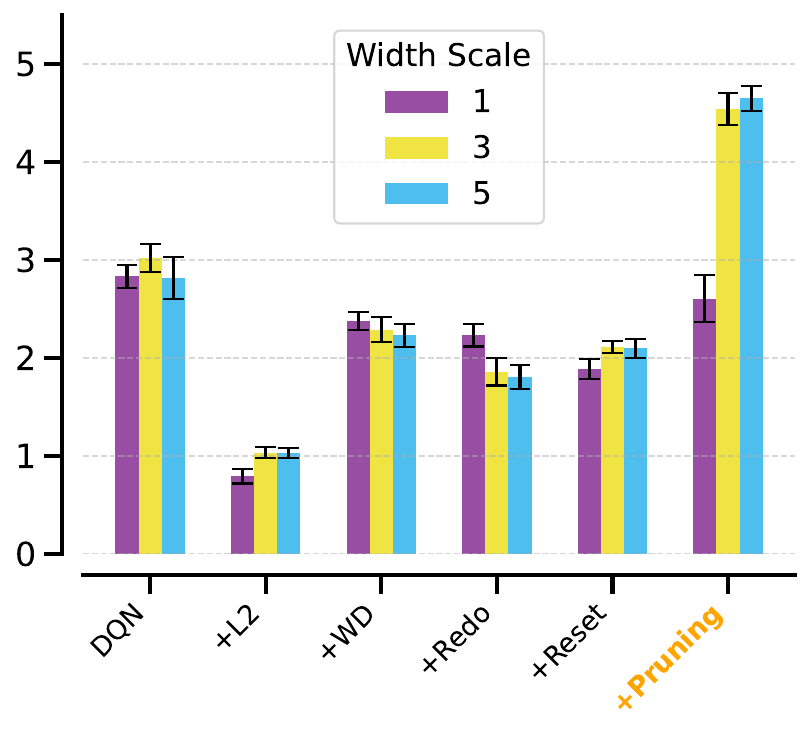}%
     \vspace{-0.3cm}
     \caption{{\bf Comparison against network resets \citep{nikishin22primacy}, weight decay, ReDo \citep{sokar2023dormant} and the normalization of \citep{kumar2022offline}.} {\bf Left:} Sample efficiency curves with a width factor of 3; {\bf Right: } final performance after 40M frames with varying widths (right panel). All experiments run on DQN with the ResNet architecture and a replay ratio of 0.25.}
    \label{fig:scalingWidths_learningCurves}
   \end{minipage}
    \hspace{0.18cm}
   \begin {minipage}{0.305\textwidth}
     \includegraphics[width=\textwidth]{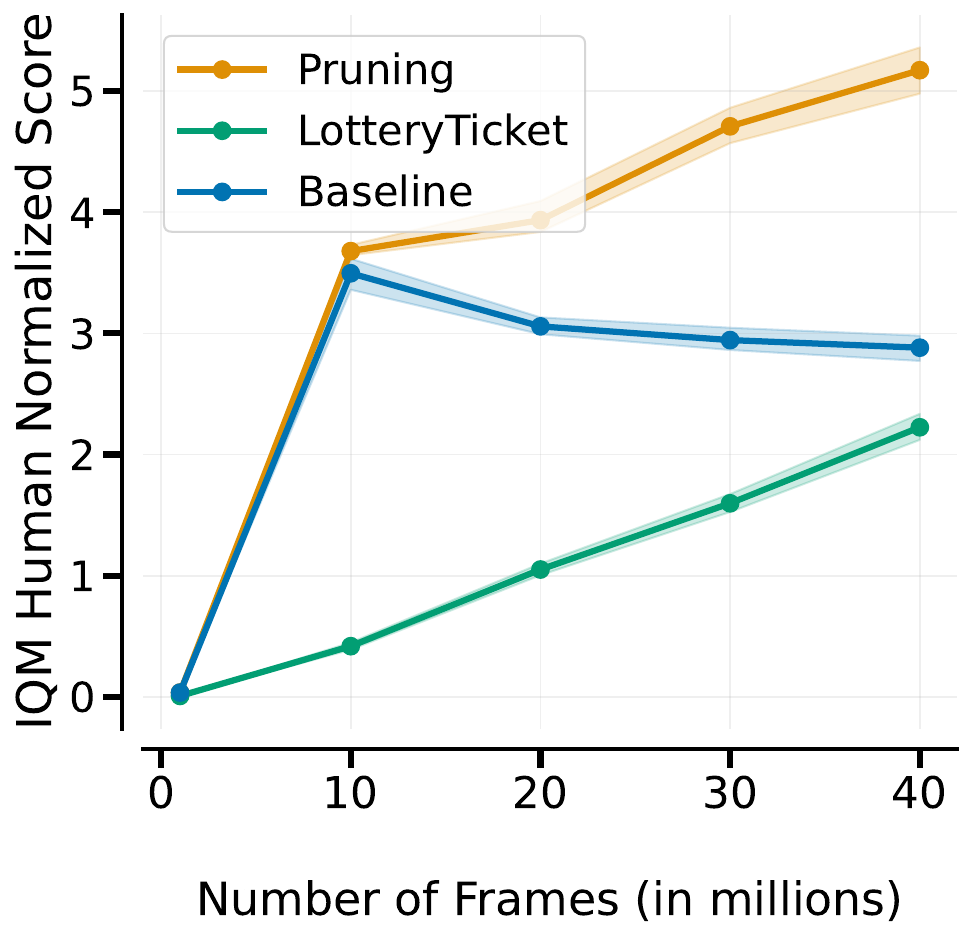}
     \vspace{-0.7cm}
     \caption{\textbf{Lottery ticket hypothesis experiment}. Taking the final pruned network (with a width factor of 3) and retraining with the original initialization results in worse performance.}
     \label{fig:lottery_ticket}
   \end{minipage}
   \vspace{0.3cm}
\end{figure*}

\paragraph{Lottery ticket baseline}
\citet{frankle2018lottery} argued that neural networks contain sparse sub-networks that can be trained at high levels of sparsity without gradual pruning; the authors provide an algorithm for finding these {\em winning tickets}. After training a network with pruning, we train a new network with the final mask {\em fixed} (i.e. not adjusted during training) and with the parameters initialized as in the original dense network. We found that the proposal under-performs both the pruning approach and the unpruned baseline. It is interesting to observe that both the pruning approach and the lottery ticket experiment seem to still be progressing at 40M, whereas the baseline seems to start deteriorating (\autoref{fig:lottery_ticket}).

\paragraph{Dynamic sparse training baselines}
\citet{evci2020rigging} proposed RigL as a dynamic sparse training mechanism that maintains a sparse network throughout the entirety of training. In \cref{sec:rigl} we evaluated RigL at various sparsity levels and found that, while effective, RigL is unable to match the performance of gradual magnitude pruning; these results are consistent with those of \citet{graesser2022state}.

\paragraph{Normalization baselines} 
To investigate the role normalization plays on the performance gains produced by pruning, we consider two types of $\ell_2$ normalization that have proven effective in the literature. The first is weight decay ({\bf WD}), a standard technique that adds an extra term to the loss that penalizes $\ell_2$ norm of the weights, thereby discouraging network parameters from growing too large. The second is {\bf L2}, the regularization approach proposed by \citet{kumar2022offline}, which is designed to enforce an $\ell_2$ norm of 1 for the parameters. 

\paragraph{Plasticity injection baselines}
We compare against two recent works that proposed methods for directly dealing with loss of plasticity. \citet{nikishin22primacy} observed a decline in performance with an increased replay ratio, attributing it to overfitting on early samples, an effect they termed the ``primacy bias''. The authors suggested periodically resetting the network and demonstrated that it proved very effective at mitigating the primacy bias, and overfitting in general (this is labeled as {\bf Reset} in our results). 

\citet{sokar2023dormant} demonstrated that most deep RL agents suffer from the {\em dormant neuron phenomenon}, whereby neurons increasingly ``turn off'' during training of deep RL agents, thus reducing network expressivity. To mitigate this, they proposed a simple and effective method that Recycles
Dormant neurons ({\bf ReDo}) throughout training.

As \autoref{fig:scalingWidths_learningCurves} illustrates,  gradual magnitude pruning surpasses all the other regularization methods at all levels of scale, and throughout the entirety of training. Interestingly, most of the regularization methods suffer a degradation when increasing network width. This suggests that the effect of pruning cannot be solely attributed to either a form of normalization or plasticity injection. However, as we will see below, increased plasticity does seem to arise out of its use. {We provide sweeps over various baseline hyperparameters in \cref{sec:redoSweep,sec:freqResetSweep,sec:layerResetSweep,sec:weightDecaySweep}}.

\subsection{Impact on plasticity}
Plasticity is a neural network's capacity to rapidly adjust in response to shifting data distributions \citep{lyle2022understanding,clare2023understanding, lewandowski2023curvature}; given the non-stationarity of reinforcement learning, it is crucial to maintain to ensure good performance. However, RL networks are known to lose plasticity over the course of training \citep{nikishin22primacy,sokar2023dormant,lee2023plastic}.

\begin{figure*}[!t]
    \centering
    \includegraphics[width=0.48\textwidth]{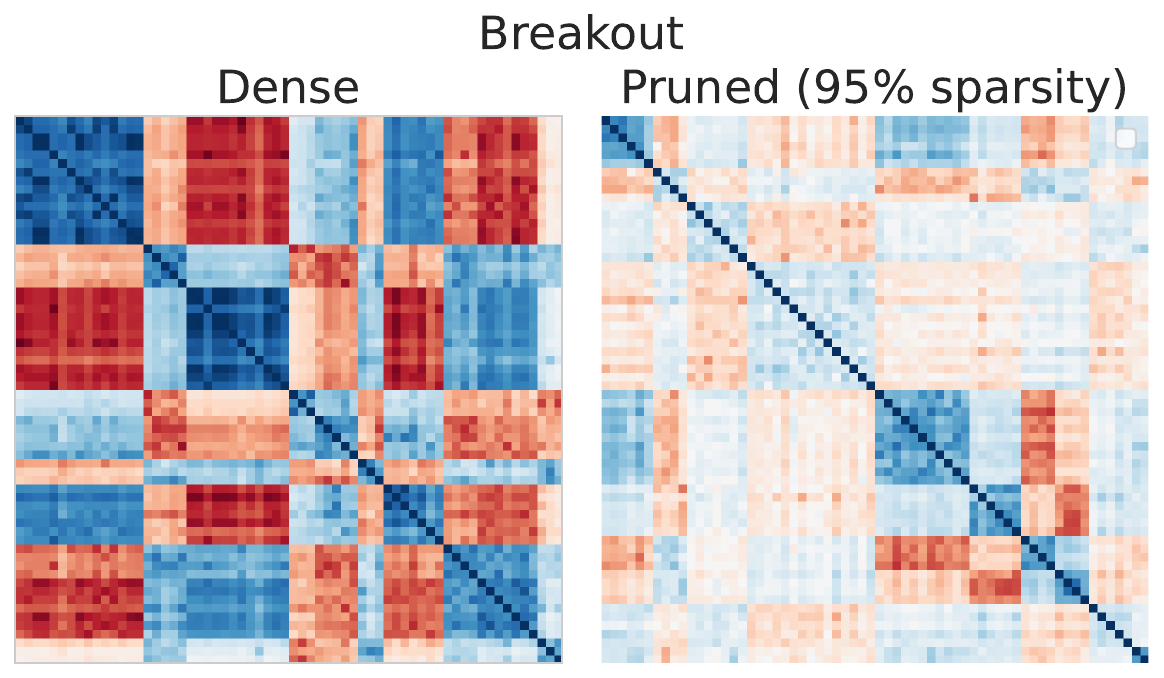}%
    \includegraphics[width=0.48\textwidth]{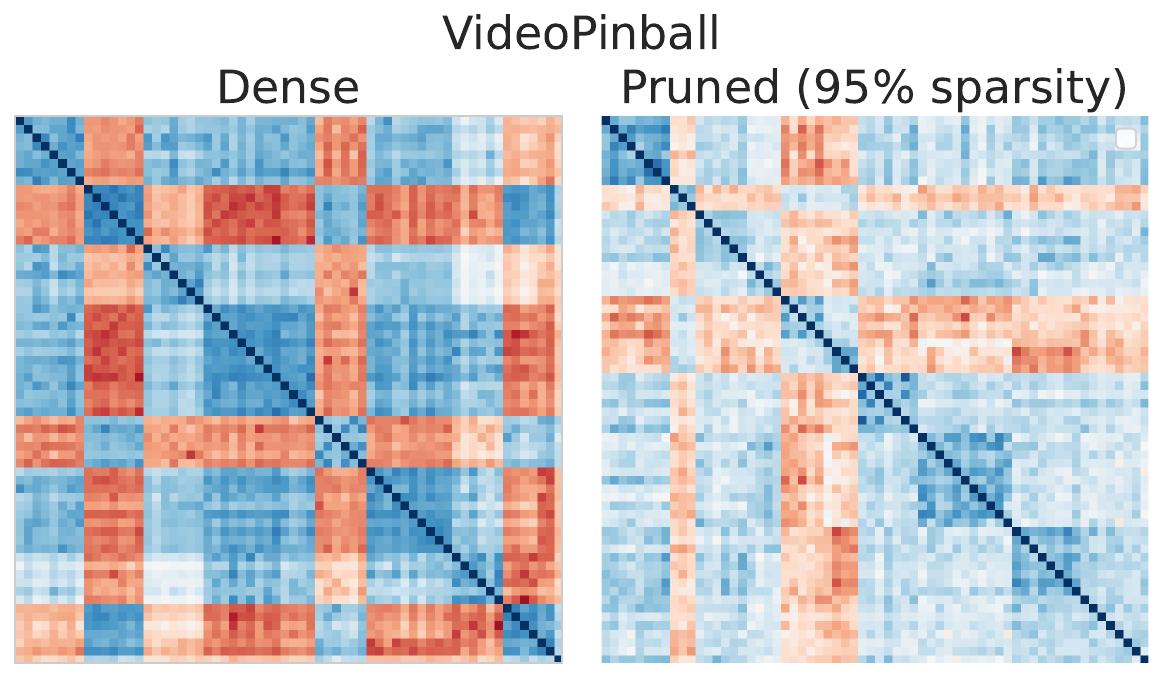}%
    \vspace{-0.1cm}
    \caption{\textbf{Gradient covariance matrices} for Breakout (\textbf{left}) and VideoPinball (\textbf{right}) atari games. \textcolor{red}{\textbf{Dark red}} denotes high negative correlation, while \textcolor{blue}{\textbf{dark blue}} indicates high positive correlation. The use of pruning induces weaker gradient correlation and less gradient interference, as evidenced by the paler hues in the heatmaps for the sparse networks.
    }
    \vspace{-0.2cm}
    \label{fig:gradientCovariance}
\end{figure*}

\citet{clare2023understanding} conducted an assessment of the covariance structure of gradients to examine the loss landscape of the network, and argued that improved performance, and increased plasticity, is often associated with weaker gradient correlation and reduced gradient interference. Our observations align with these findings, as illustrated in the gradient covariance heat maps in \autoref{fig:gradientCovariance}. In dense networks, gradients exhibit a notable colinearity, whereas this colinearity is dramatically reduced in the pruned networks.

\section{Discussion and Conclusion}
\label{sec:Discussion and Conclusion}

Prior work has demonstrated that reinforcement learning agents have a tendency to under-utilize its available parameters, and that this under-utilization increases throughout training and is amplified as network sizes increase \citep{graesser2022state,nikishin22primacy,sokar2023dormant,schwarzer2023bigger}. RL agents achieve strong performance in the majority of the established benchmarks with small networks (relative to those used in language models, for instance), so this evident parameter-inefficiency may be brushed off as being less critical than other, more algorithmic, considerations.

As RL continues to grow outside of academic benchmarks and into more complex tasks, it is almost surely going to necessitate larger, and more expressive, networks. In this case parameter efficiency becomes crucial to avoid the performance collapse prior works have shown, as well as for reducing computational costs \citep{ceron2021revisiting}.

Our work provides convincing evidence that sparse training techniques such as gradual magnitude pruning can be effective at maximizing network utilization{, especially as the initial networks are scaled (see \cref{fig:topline,fig:scalingWidths_cnn})}. The results in Figures~\ref{fig:scalingWidths_offline}, \ref{fig:samplefficient_algo}, and \ref{fig:varyingSchedules} all demonstrate that the sparse networks produced by pruning are better at maintaining stable performance when trained for longer. {The advantages of pruning remain even when sweeping over various baseline hyper-parameters (see \cref{sec:learningRateSweep,sec:epsilonSweep,sec:updateHorizonSweep,sec:batchSizeSweep}). It is worth noting that the performance of the dense baselines does improve when adjusting the learning rate based on the width multiplier (\cref{sec:learningRateSweep}); however, pruning is still the most performant in these settings.}

Collectively, our results demonstrate that, by meaningfully removing network parameters throughout training, we can outperform traditional dense counterparts and continue to improve performance as we grow the initial network architectures. Our results with varied agents and training regimes imply gradual magnitude pruning is a generally useful technique which can be used as a ``drop-in'' for maximizing agent performance.

\paragraph{Future work} It would be natural to explore incorporating gradual magnitude pruning into recent agents that were designed for multi-task generalization \citep{taiga2023investigating, kumar2022offline}, sample efficiency \citep{schwarzer2023bigger,doro2023sampleefficient}, and generalizability \citep{hafner2023dreamerv3}.
Further, the observed stability of the pruned networks may have implications for methods which rely on fine-tuning or reincarnation \citep{agarwal2022reincarnating}.

Recent advances in hardware accelerators for training sparse networks may result in faster training times, and serve as an incentive for further research in methods for sparse network training.
Further, the fact that a consequence of this approach is a network with fewer parameters than when initialized renders it appealing for downstream applications on edge devices.
 
At a minimum, we hope this work serves as an invitation to explore non-standard network architectures and topologies as an effective mechanism for maximizing the performance of reinforcement learning agents. Reinforcement learning agents typically use networks originally designed for stationary problems; therefore, other topologies might be better suited to the non-stationary nature of RL.
%\paragraph{Acknowledgements}
\section*{Acknowledgements}

The authors would like to thank Laura Graesser, Utku Evci, Gopeshh Subbaraj, Evgenii Nikishin, Hugo Larochelle, Ayoub Echchahed, Zhixuan Lin and the rest of the Google DeepMind Montreal team for valuable discussions during the preparation of this work. 

Laura Graesser deserves a special mention for providing us valuable feed-back on an early draft of the paper. We thank the anonymous reviewers for their valuable help in improving our manuscript. We would also like to thank the Python community \cite{van1995python, 4160250} for developing tools that enabled this work, including NumPy \cite{harris2020array}, Matplotlib \cite{hunter2007matplotlib}, Jupyter \cite{2016ppap}, Pandas \cite{McKinney2013Python} and JAX \cite{bradbury2018jax}.
%\paragraph{Impact statement}
\section*{Impact statement}
This paper presents work whose goal is to advance the field of Machine Learning, and reinforcement learning in particular. There are many potential societal consequences of our work, none which we feel must be specifically highlighted here.

%\newpage
\bibliography{references}

\begin{thebibliography}{89}
\providecommand{\natexlab}[1]{#1}
\providecommand{\url}[1]{\texttt{#1}}
\expandafter\ifx\csname urlstyle\endcsname\relax
  \providecommand{\doi}[1]{doi: #1}\else
  \providecommand{\doi}{doi: \begingroup \urlstyle{rm}\Url}\fi

\bibitem[Agarwal et~al.(2020)Agarwal, Schuurmans, and
  Norouzi]{agarwal2020optimistic}
Agarwal, R., Schuurmans, D., and Norouzi, M.
\newblock An optimistic perspective on offline reinforcement learning.
\newblock In III, H.~D. and Singh, A. (eds.), \emph{Proceedings of the 37th
  International Conference on Machine Learning}, volume 119 of
  \emph{Proceedings of Machine Learning Research}, pp.\  104--114. PMLR, 13--18
  Jul 2020.
\newblock URL \url{https://proceedings.mlr.press/v119/agarwal20c.html}.

\bibitem[Agarwal et~al.(2021)Agarwal, Schwarzer, Castro, Courville, and
  Bellemare]{agarwal2021deep}
Agarwal, R., Schwarzer, M., Castro, P.~S., Courville, A.~C., and Bellemare, M.
\newblock Deep reinforcement learning at the edge of the statistical precipice.
\newblock \emph{Advances in neural information processing systems},
  34:\penalty0 29304--29320, 2021.

\bibitem[Agarwal et~al.(2022)Agarwal, Schwarzer, Castro, Courville, and
  Bellemare]{agarwal2022reincarnating}
Agarwal, R., Schwarzer, M., Castro, P.~S., Courville, A.~C., and Bellemare, M.
\newblock Reincarnating reinforcement learning: Reusing prior computation to
  accelerate progress.
\newblock In Koyejo, S., Mohamed, S., Agarwal, A., Belgrave, D., Cho, K., and
  Oh, A. (eds.), \emph{Advances in Neural Information Processing Systems},
  volume~35, pp.\  28955--28971. Curran Associates, Inc., 2022.

\bibitem[Arnob et~al.(2021)Arnob, Ohib, Plis, and Precup]{arnob2021single}
Arnob, S.~Y., Ohib, R., Plis, S., and Precup, D.
\newblock Single-shot pruning for offline reinforcement learning.
\newblock \emph{arXiv preprint arXiv:2112.15579}, 2021.

\bibitem[Ba et~al.(2016)Ba, Kiros, and Hinton]{ba2016layer}
Ba, J.~L., Kiros, J.~R., and Hinton, G.~E.
\newblock Layer normalization.
\newblock \emph{arXiv preprint arXiv:1607.06450}, 2016.

\bibitem[Bellemare et~al.(2013)Bellemare, Naddaf, Veness, and
  Bowling]{Bellemare_2013}
Bellemare, M.~G., Naddaf, Y., Veness, J., and Bowling, M.
\newblock The arcade learning environment: An evaluation platform for general
  agents.
\newblock \emph{Journal of Artificial Intelligence Research}, 47:\penalty0
  253--279, jun 2013.
\newblock \doi{10.1613/jair.3912}.

\bibitem[Bellemare et~al.(2017)Bellemare, Dabney, and Munos]{Bellemare2017ADP}
Bellemare, M.~G., Dabney, W., and Munos, R.
\newblock A distributional perspective on reinforcement learning.
\newblock In \emph{ICML}, 2017.

\bibitem[Bellemare et~al.(2020)Bellemare, Candido, Castro, Gong, Machado,
  Moitra, Ponda, and Wang]{Bellemare2020AutonomousNO}
Bellemare, M.~G., Candido, S., Castro, P.~S., Gong, J., Machado, M.~C., Moitra,
  S., Ponda, S.~S., and Wang, Z.
\newblock Autonomous navigation of stratospheric balloons using reinforcement
  learning.
\newblock \emph{Nature}, 588:\penalty0 77 -- 82, 2020.

\bibitem[Berner et~al.(2019)Berner, Brockman, Chan, Cheung, D{e}biak, Dennison,
  Farhi, Fischer, Hashme, Hesse, et~al.]{berner2019dota}
Berner, C., Brockman, G., Chan, B., Cheung, V., D{e}biak, P., Dennison, C.,
  Farhi, D., Fischer, Q., Hashme, S., Hesse, C., et~al.
\newblock Dota 2 with large scale deep reinforcement learning.
\newblock \emph{arXiv preprint arXiv:1912.06680}, 2019.

\bibitem[Bjorck et~al.(2021)Bjorck, Gomes, and Weinberger]{bjorck2021towards}
Bjorck, N., Gomes, C.~P., and Weinberger, K.~Q.
\newblock Towards deeper deep reinforcement learning with spectral
  normalization.
\newblock \emph{Advances in Neural Information Processing Systems},
  34:\penalty0 8242--8255, 2021.

\bibitem[Bradbury et~al.(2018)Bradbury, Frostig, Hawkins, Johnson, Leary,
  Maclaurin, Necula, Paszke, VanderPlas, Wanderman-Milne,
  et~al.]{bradbury2018jax}
Bradbury, J., Frostig, R., Hawkins, P., Johnson, M.~J., Leary, C., Maclaurin,
  D., Necula, G., Paszke, A., VanderPlas, J., Wanderman-Milne, S., et~al.
\newblock Jax: composable transformations of python+ numpy programs.
\newblock 2018.

\bibitem[Castro et~al.(2018)Castro, Moitra, Gelada, Kumar, and
  Bellemare]{castro2018dopamine}
Castro, P.~S., Moitra, S., Gelada, C., Kumar, S., and Bellemare, M.~G.
\newblock Dopamine: A research framework for deep reinforcement learning.
\newblock \emph{arXiv preprint arXiv:1812.06110}, 2018.

\bibitem[Ceron \& Castro(2021)Ceron and Castro]{ceron2021revisiting}
Ceron, J. S.~O. and Castro, P.~S.
\newblock Revisiting rainbow: Promoting more insightful and inclusive deep
  reinforcement learning research.
\newblock In \emph{International Conference on Machine Learning}, pp.\
  1373--1383. PMLR, 2021.

\bibitem[Ceron et~al.(2023)Ceron, Bellemare, and Castro]{ceron2023small}
Ceron, J. S.~O., Bellemare, M.~G., and Castro, P.~S.
\newblock Small batch deep reinforcement learning.
\newblock In \emph{Thirty-seventh Conference on Neural Information Processing
  Systems}, 2023.
\newblock URL \url{https://openreview.net/forum?id=wPqEvmwFEh}.

\bibitem[Ceron et~al.(2024)Ceron, Sokar, Willi, Lyle, Farebrother, Foerster,
  Dziugaite, Precup, and Castro]{obando2024mixtures}
Ceron, J. S.~O., Sokar, G., Willi, T., Lyle, C., Farebrother, J., Foerster,
  J.~N., Dziugaite, G.~K., Precup, D., and Castro, P.~S.
\newblock Mixtures of experts unlock parameter scaling for deep {RL}.
\newblock In \emph{Forty-first International Conference on Machine Learning},
  2024.
\newblock URL \url{https://openreview.net/forum?id=X9VMhfFxwn}.

\bibitem[Cetin et~al.(2022)Cetin, Ball, Roberts, and
  Celiktutan]{cetin2022stabilizing}
Cetin, E., Ball, P.~J., Roberts, S., and Celiktutan, O.
\newblock Stabilizing off-policy deep reinforcement learning from pixels.
\newblock In \emph{International Conference on Machine Learning}, pp.\
  2784--2810. PMLR, 2022.

\bibitem[Cobbe et~al.(2020)Cobbe, Hesse, Hilton, and
  Schulman]{cobbe2020leveraging}
Cobbe, K., Hesse, C., Hilton, J., and Schulman, J.
\newblock Leveraging procedural generation to benchmark reinforcement learning.
\newblock In \emph{International conference on machine learning}, pp.\
  2048--2056. PMLR, 2020.

\bibitem[Dabney et~al.(2018)Dabney, Ostrovski, Silver, and
  Munos]{dabney2018implicit}
Dabney, W., Ostrovski, G., Silver, D., and Munos, R.
\newblock Implicit quantile networks for distributional reinforcement learning.
\newblock In \emph{International conference on machine learning}, pp.\
  1096--1105. PMLR, 2018.

\bibitem[D'Oro et~al.(2023)D'Oro, Schwarzer, Nikishin, Bacon, Bellemare, and
  Courville]{doro2023sampleefficient}
D'Oro, P., Schwarzer, M., Nikishin, E., Bacon, P.-L., Bellemare, M.~G., and
  Courville, A.
\newblock Sample-efficient reinforcement learning by breaking the replay ratio
  barrier.
\newblock In \emph{The Eleventh International Conference on Learning
  Representations}, 2023.
\newblock URL \url{https://openreview.net/forum?id=OpC-9aBBVJe}.

\bibitem[Espeholt et~al.(2018)Espeholt, Soyer, Munos, Simonyan, Mnih, Ward,
  Doron, Firoiu, Harley, Dunning, et~al.]{espeholt2018impala}
Espeholt, L., Soyer, H., Munos, R., Simonyan, K., Mnih, V., Ward, T., Doron,
  Y., Firoiu, V., Harley, T., Dunning, I., et~al.
\newblock Impala: Scalable distributed deep-rl with importance weighted
  actor-learner architectures.
\newblock In \emph{International conference on machine learning}, pp.\
  1407--1416. PMLR, 2018.

\bibitem[Evci et~al.(2020)Evci, Gale, Menick, Castro, and
  Elsen]{evci2020rigging}
Evci, U., Gale, T., Menick, J., Castro, P.~S., and Elsen, E.
\newblock Rigging the lottery: Making all tickets winners.
\newblock In \emph{International Conference on Machine Learning}, pp.\
  2943--2952. PMLR, 2020.

\bibitem[Farebrother et~al.(2023)Farebrother, Greaves, Agarwal, Lan, Goroshin,
  Castro, and Bellemare]{farebrother2023protovalue}
Farebrother, J., Greaves, J., Agarwal, R., Lan, C.~L., Goroshin, R., Castro,
  P.~S., and Bellemare, M.~G.
\newblock Proto-value networks: Scaling representation learning with auxiliary
  tasks.
\newblock In \emph{Submitted to The Eleventh International Conference on
  Learning Representations}, 2023.
\newblock URL \url{https://openreview.net/forum?id=oGDKSt9JrZi}.
\newblock under review.

\bibitem[Farebrother et~al.(2024)Farebrother, Orbay, Vuong, Ta{\"i}ga,
  Chebotar, Xiao, Irpan, Levine, Castro, Faust, Kumar, and
  Agarwal]{farebrother24classification}
Farebrother, J., Orbay, J., Vuong, Q., Ta{\"i}ga, A.~A., Chebotar, Y., Xiao,
  T., Irpan, A., Levine, S., Castro, P.~S., Faust, A., Kumar, A., and Agarwal,
  R.
\newblock Stop regressing: Training value functions via classification for
  scalable deep rl.
\newblock In \emph{Forty-first International Conference on Machine Learning}.
  PMLR, 2024.

\bibitem[Fawzi et~al.(2022)Fawzi, Balog, Huang, Hubert, Romera-Paredes,
  Barekatain, Novikov, R~Ruiz, Schrittwieser, Swirszcz,
  et~al.]{fawzi2022discovering}
Fawzi, A., Balog, M., Huang, A., Hubert, T., Romera-Paredes, B., Barekatain,
  M., Novikov, A., R~Ruiz, F.~J., Schrittwieser, J., Swirszcz, G., et~al.
\newblock Discovering faster matrix multiplication algorithms with
  reinforcement learning.
\newblock \emph{Nature}, 610\penalty0 (7930):\penalty0 47--53, 2022.

\bibitem[Fedus et~al.(2020)Fedus, Ramachandran, Agarwal, Bengio, Larochelle,
  Rowland, and Dabney]{fedus2020revisiting}
Fedus, W., Ramachandran, P., Agarwal, R., Bengio, Y., Larochelle, H., Rowland,
  M., and Dabney, W.
\newblock Revisiting fundamentals of experience replay.
\newblock In \emph{International Conference on Machine Learning}, pp.\
  3061--3071. PMLR, 2020.

\bibitem[Fortunato et~al.(2018)Fortunato, Azar, Piot, Menick, Osband, Graves,
  Mnih, Munos, Hassabis, Pietquin, Blundell, and Legg]{fortunato18noisy}
Fortunato, M., Azar, M.~G., Piot, B., Menick, J., Osband, I., Graves, A., Mnih,
  V., Munos, R., Hassabis, D., Pietquin, O., Blundell, C., and Legg, S.
\newblock Noisy networks for exploration.
\newblock 2018.

\bibitem[Frankle \& Carbin(2018)Frankle and Carbin]{frankle2018lottery}
Frankle, J. and Carbin, M.
\newblock The lottery ticket hypothesis: Finding sparse, trainable neural
  networks.
\newblock In \emph{International Conference on Learning Representations}, 2018.

\bibitem[Gal \& Ghahramani(2016)Gal and Ghahramani]{gal2016dropout}
Gal, Y. and Ghahramani, Z.
\newblock Dropout as a bayesian approximation: Representing model uncertainty
  in deep learning.
\newblock In \emph{international conference on machine learning}, pp.\
  1050--1059. PMLR, 2016.

\bibitem[Gale et~al.(2019)Gale, Elsen, and Hooker]{gale2019state}
Gale, T., Elsen, E., and Hooker, S.
\newblock The state of sparsity in deep neural networks.
\newblock \emph{arXiv preprint arXiv:1902.09574}, 2019.

\bibitem[Graesser et~al.(2022)Graesser, Evci, Elsen, and
  Castro]{graesser2022state}
Graesser, L., Evci, U., Elsen, E., and Castro, P.~S.
\newblock The state of sparse training in deep reinforcement learning.
\newblock In \emph{International Conference on Machine Learning}, pp.\
  7766--7792. PMLR, 2022.

\bibitem[Grooten et~al.(2023)Grooten, Sokar, Dohare, Mocanu, Taylor,
  Pechenizkiy, and Mocanu]{grooteautomatic23}
Grooten, B., Sokar, G., Dohare, S., Mocanu, E., Taylor, M.~E., Pechenizkiy, M.,
  and Mocanu, D.~C.
\newblock Automatic noise filtering with dynamic sparse training in deep
  reinforcement learning.
\newblock In \emph{Proceedings of the 2023 International Conference on
  Autonomous Agents and Multiagent Systems}, AAMAS '23, pp.\  1932–1941,
  Richland, SC, 2023. International Foundation for Autonomous Agents and
  Multiagent Systems.
\newblock ISBN 9781450394321.

\bibitem[Haarnoja et~al.(2018)Haarnoja, Zhou, Abbeel, and
  Levine]{haarnoja2018soft}
Haarnoja, T., Zhou, A., Abbeel, P., and Levine, S.
\newblock Soft actor-critic: Off-policy maximum entropy deep reinforcement
  learning with a stochastic actor.
\newblock In \emph{International conference on machine learning}, pp.\
  1861--1870. PMLR, 2018.

\bibitem[Hafner et~al.(2023)Hafner, Pasukonis, Ba, and
  Lillicrap]{hafner2023dreamerv3}
Hafner, D., Pasukonis, J., Ba, J., and Lillicrap, T.
\newblock Mastering diverse domains through world models.
\newblock \emph{arXiv preprint arXiv:2301.04104}, 2023.

\bibitem[Han et~al.(2015)Han, Mao, and Dally]{han2015deep}
Han, S., Mao, H., and Dally, W.~J.
\newblock Deep compression: Compressing deep neural networks with pruning,
  trained quantization and huffman coding.
\newblock \emph{arXiv preprint arXiv:1510.00149}, 2015.

\bibitem[Harris et~al.(2020)Harris, Millman, Van Der~Walt, Gommers, Virtanen,
  Cournapeau, Wieser, Taylor, Berg, Smith, et~al.]{harris2020array}
Harris, C.~R., Millman, K.~J., Van Der~Walt, S.~J., Gommers, R., Virtanen, P.,
  Cournapeau, D., Wieser, E., Taylor, J., Berg, S., Smith, N.~J., et~al.
\newblock Array programming with numpy.
\newblock \emph{Nature}, 585\penalty0 (7825):\penalty0 357--362, 2020.

\bibitem[Hessel et~al.(2018)Hessel, Modayil, Hasselt, Schaul, Ostrovski,
  Dabney, Horgan, Piot, Azar, and Silver]{Hessel2018RainbowCI}
Hessel, M., Modayil, J., Hasselt, H.~V., Schaul, T., Ostrovski, G., Dabney, W.,
  Horgan, D., Piot, B., Azar, M.~G., and Silver, D.
\newblock Rainbow: Combining improvements in deep reinforcement learning.
\newblock In \emph{AAAI}, 2018.

\bibitem[Hiraoka et~al.(2021)Hiraoka, Imagawa, Hashimoto, Onishi, and
  Tsuruoka]{hiraoka2021dropout}
Hiraoka, T., Imagawa, T., Hashimoto, T., Onishi, T., and Tsuruoka, Y.
\newblock Dropout q-functions for doubly efficient reinforcement learning.
\newblock In \emph{International Conference on Learning Representations}, 2021.

\bibitem[Hunter(2007)]{hunter2007matplotlib}
Hunter, J.~D.
\newblock Matplotlib: A 2d graphics environment.
\newblock \emph{Computing in science \& engineering}, 9\penalty0 (03):\penalty0
  90--95, 2007.

\bibitem[Ioffe \& Szegedy(2015)Ioffe and Szegedy]{ioffe2015batch}
Ioffe, S. and Szegedy, C.
\newblock Batch normalization: Accelerating deep network training by reducing
  internal covariate shift.
\newblock In \emph{International conference on machine learning}, pp.\
  448--456. pmlr, 2015.

\bibitem[Kaiser et~al.(2020)Kaiser, Babaeizadeh, Milos, Osinski, Campbell,
  Czechowski, Erhan, Finn, Kozakowski, Levine, et~al.]{Kaiser2020Model}
Kaiser, L., Babaeizadeh, M., Milos, P., Osinski, B., Campbell, R.~H.,
  Czechowski, K., Erhan, D., Finn, C., Kozakowski, P., Levine, S., et~al.
\newblock Model-based reinforcement learning for atari.
\newblock \emph{International Conference on Learning Representations}, 2020.

\bibitem[Kingma \& Ba(2014)Kingma and Ba]{kingma2014adam}
Kingma, D.~P. and Ba, J.
\newblock Adam: A method for stochastic optimization.
\newblock \emph{arXiv preprint arXiv:1412.6980}, 2014.

\bibitem[{Kluyver} et~al.(2016){Kluyver}, {Ragan-Kelley}, {P{\'e}rez},
  {Granger}, {Bussonnier}, {Frederic}, {Kelley}, {Hamrick}, {Grout}, {Corlay},
  {Ivanov}, {Avila}, {Abdalla}, {Willing}, and {Jupyter Development
  Team}]{2016ppap}
{Kluyver}, T., {Ragan-Kelley}, B., {P{\'e}rez}, F., {Granger}, B.,
  {Bussonnier}, M., {Frederic}, J., {Kelley}, K., {Hamrick}, J., {Grout}, J.,
  {Corlay}, S., {Ivanov}, P., {Avila}, D., {Abdalla}, S., {Willing}, C., and
  {Jupyter Development Team}.
\newblock {Jupyter Notebooks{\textemdash}a publishing format for reproducible
  computational workflows}.
\newblock In \emph{IOS Press}, pp.\  87--90. 2016.
\newblock \doi{10.3233/978-1-61499-649-1-87}.

\bibitem[Kostrikov et~al.(2020)Kostrikov, Yarats, and
  Fergus]{kostrikov2020image}
Kostrikov, I., Yarats, D., and Fergus, R.
\newblock Image augmentation is all you need: Regularizing deep reinforcement
  learning from pixels.
\newblock \emph{arXiv preprint arXiv:2004.13649}, 2020.

\bibitem[Kumar et~al.(2020)Kumar, Zhou, Tucker, and
  Levine]{kumar2020conservative}
Kumar, A., Zhou, A., Tucker, G., and Levine, S.
\newblock Conservative q-learning for offline reinforcement learning.
\newblock \emph{Advances in Neural Information Processing Systems},
  33:\penalty0 1179--1191, 2020.

\bibitem[Kumar et~al.(2021{\natexlab{a}})Kumar, Agarwal, Ghosh, and
  Levine]{kumar2021implicit}
Kumar, A., Agarwal, R., Ghosh, D., and Levine, S.
\newblock Implicit under-parameterization inhibits data-efficient deep
  reinforcement learning.
\newblock In \emph{International Conference on Learning Representations},
  2021{\natexlab{a}}.
\newblock URL \url{https://openreview.net/forum?id=O9bnihsFfXU}.

\bibitem[Kumar et~al.(2021{\natexlab{b}})Kumar, Agarwal, Ma, Courville, Tucker,
  and Levine]{kumar2021dr3}
Kumar, A., Agarwal, R., Ma, T., Courville, A., Tucker, G., and Levine, S.
\newblock Dr3: Value-based deep reinforcement learning requires explicit
  regularization.
\newblock \emph{arXiv preprint arXiv:2112.04716}, 2021{\natexlab{b}}.

\bibitem[Kumar et~al.(2022)Kumar, Agarwal, Geng, Tucker, and
  Levine]{kumar2022offline}
Kumar, A., Agarwal, R., Geng, X., Tucker, G., and Levine, S.
\newblock Offline q-learning on diverse multi-task data both scales and
  generalizes.
\newblock In \emph{The Eleventh International Conference on Learning
  Representations}, 2022.

\bibitem[Lee et~al.(2023)Lee, Cho, Kim, Gwak, Kim, Choo, Yun, and
  Yun]{lee2023plastic}
Lee, H., Cho, H., Kim, H., Gwak, D., Kim, J., Choo, J., Yun, S.-Y., and Yun, C.
\newblock Plastic: Improving input and label plasticity for sample efficient
  reinforcement learning.
\newblock In \emph{Thirty-seventh Conference on Neural Information Processing
  Systems}, 2023.

\bibitem[Lee et~al.(2024)Lee, Park, Mitchell, Pilault, Ceron, Kim, Lee,
  Frantar, Long, Yazdanbakhsh, et~al.]{lee2024jaxpruner}
Lee, J.~H., Park, W., Mitchell, N.~E., Pilault, J., Ceron, J. S.~O., Kim,
  H.-B., Lee, N., Frantar, E., Long, Y., Yazdanbakhsh, A., et~al.
\newblock Jaxpruner: A concise library for sparsity research.
\newblock In \emph{Conference on Parsimony and Learning}, pp.\  515--528. PMLR,
  2024.

\bibitem[Lewandowski et~al.(2023)Lewandowski, Tanaka, Schuurmans, and
  Machado]{lewandowski2023curvature}
Lewandowski, A., Tanaka, H., Schuurmans, D., and Machado, M.~C.
\newblock Curvature explains loss of plasticity.
\newblock \emph{arXiv preprint arXiv:2312.00246}, 2023.

\bibitem[Liu et~al.(2020)Liu, Li, Kang, and Darrell]{liu2020regularization}
Liu, Z., Li, X., Kang, B., and Darrell, T.
\newblock Regularization matters in policy optimization-an empirical study on
  continuous control.
\newblock In \emph{International Conference on Learning Representations}, 2020.

\bibitem[Livne \& Cohen(2020)Livne and Cohen]{Livne_2020}
Livne, D. and Cohen, K.
\newblock Pops: Policy pruning and shrinking for deep reinforcement learning.
\newblock \emph{IEEE Journal of Selected Topics in Signal Processing},
  14\penalty0 (4):\penalty0 789–801, May 2020.
\newblock ISSN 1941-0484.
\newblock \doi{10.1109/jstsp.2020.2967566}.
\newblock URL \url{http://dx.doi.org/10.1109/JSTSP.2020.2967566}.

\bibitem[Lyle et~al.(2022)Lyle, Rowland, and Dabney]{lyle2022understanding}
Lyle, C., Rowland, M., and Dabney, W.
\newblock Understanding and preventing capacity loss in reinforcement learning.
\newblock In \emph{International Conference on Learning Representations}, 2022.
\newblock URL \url{https://openreview.net/forum?id=ZkC8wKoLbQ7}.

\bibitem[Lyle et~al.(2023)Lyle, Zheng, Nikishin, Pires, Pascanu, and
  Dabney]{clare2023understanding}
Lyle, C., Zheng, Z., Nikishin, E., Pires, B.~A., Pascanu, R., and Dabney, W.
\newblock Understanding plasticity in neural networks.
\newblock In \emph{Proceedings of the 40th International Conference on Machine
  Learning}, ICML'23. JMLR.org, 2023.

\bibitem[Lyle et~al.(2024)Lyle, Zheng, Khetarpal, van Hasselt, Pascanu,
  Martens, and Dabney]{lyle2024disentangling}
Lyle, C., Zheng, Z., Khetarpal, K., van Hasselt, H., Pascanu, R., Martens, J.,
  and Dabney, W.
\newblock Disentangling the causes of plasticity loss in neural networks.
\newblock \emph{arXiv preprint arXiv:2402.18762}, 2024.

\bibitem[McKinney(2013)]{McKinney2013Python}
McKinney, W.
\newblock \emph{Python for Data Analysis: Data Wrangling with Pandas, {NumPy},
  and {IPython}}.
\newblock O'Reilly Media, 1 edition, February 2013.
\newblock ISBN 9789351100065.
\newblock URL
  \url{http://www.amazon.com/exec/obidos/redirect?tag=citeulike07-20\&path=ASIN/1449319793}.

\bibitem[Mnih et~al.(2015)Mnih, Kavukcuoglu, Silver, Rusu, Veness, Bellemare,
  Graves, Riedmiller, Fidjeland, Ostrovski, Petersen, Beattie, Sadik,
  Antonoglou, King, Kumaran, Wierstra, Legg, and Hassabis]{mnih2015humanlevel}
Mnih, V., Kavukcuoglu, K., Silver, D., Rusu, A.~A., Veness, J., Bellemare,
  M.~G., Graves, A., Riedmiller, M., Fidjeland, A.~K., Ostrovski, G., Petersen,
  S., Beattie, C., Sadik, A., Antonoglou, I., King, H., Kumaran, D., Wierstra,
  D., Legg, S., and Hassabis, D.
\newblock Human-level control through deep reinforcement learning.
\newblock \emph{Nature}, 518\penalty0 (7540):\penalty0 529--533, February 2015.

\bibitem[Mocanu et~al.(2018)Mocanu, Mocanu, Stone, Nguyen, Gibescu, and
  Liotta]{mocanu2018scalable}
Mocanu, D.~C., Mocanu, E., Stone, P., Nguyen, P.~H., Gibescu, M., and Liotta,
  A.
\newblock Scalable training of artificial neural networks with adaptive sparse
  connectivity inspired by network science.
\newblock \emph{Nature communications}, 9\penalty0 (1):\penalty0 2383, 2018.

\bibitem[Nikishin et~al.(2022)Nikishin, Schwarzer, D'Oro, Bacon, and
  Courville]{nikishin22primacy}
Nikishin, E., Schwarzer, M., D'Oro, P., Bacon, P.-L., and Courville, A.
\newblock The primacy bias in deep reinforcement learning.
\newblock In Chaudhuri, K., Jegelka, S., Song, L., Szepesvari, C., Niu, G., and
  Sabato, S. (eds.), \emph{Proceedings of the 39th International Conference on
  Machine Learning}, volume 162 of \emph{Proceedings of Machine Learning
  Research}, pp.\  16828--16847. PMLR, 17--23 Jul 2022.

\bibitem[Nikishin et~al.(2023)Nikishin, Oh, Ostrovski, Lyle, Pascanu, Dabney,
  and Barreto]{nikishin2023deep}
Nikishin, E., Oh, J., Ostrovski, G., Lyle, C., Pascanu, R., Dabney, W., and
  Barreto, A.
\newblock Deep reinforcement learning with plasticity injection.
\newblock In \emph{Thirty-seventh Conference on Neural Information Processing
  Systems}, 2023.
\newblock URL \url{https://openreview.net/forum?id=jucDLW6G9l}.

\bibitem[Oliphant(2007)]{4160250}
Oliphant, T.~E.
\newblock Python for scientific computing.
\newblock \emph{Computing in Science \& Engineering}, 9\penalty0 (3):\penalty0
  10--20, 2007.
\newblock \doi{10.1109/MCSE.2007.58}.

\bibitem[Ostrovski et~al.(2021)Ostrovski, Castro, and
  Dabney]{ostrovski2021tandem}
Ostrovski, G., Castro, P.~S., and Dabney, W.
\newblock The difficulty of passive learning in deep reinforcement learning.
\newblock In Beygelzimer, A., Dauphin, Y., Liang, P., and Vaughan, J.~W.
  (eds.), \emph{Advances in Neural Information Processing Systems}, 2021.
\newblock URL \url{https://openreview.net/forum?id=nPHA8fGicZk}.

\bibitem[Ota et~al.(2021)Ota, Jha, and Kanezaki]{ota2021training}
Ota, K., Jha, D.~K., and Kanezaki, A.
\newblock Training larger networks for deep reinforcement learning.
\newblock \emph{arXiv preprint arXiv:2102.07920}, 2021.

\bibitem[Schaul et~al.(2016)Schaul, Quan, Antonoglou, and
  Silver]{Schaul2016PrioritizedER}
Schaul, T., Quan, J., Antonoglou, I., and Silver, D.
\newblock Prioritized experience replay.
\newblock \emph{CoRR}, abs/1511.05952, 2016.

\bibitem[Schmidt \& Schmied(2021)Schmidt and Schmied]{schmidt2021fast}
Schmidt, D. and Schmied, T.
\newblock Fast and data-efficient training of rainbow: an experimental study on
  atari.
\newblock \emph{arXiv preprint arXiv:2111.10247}, 2021.

\bibitem[Schmitt et~al.(2018)Schmitt, Hudson, Zidek, Osindero, Doersch,
  Czarnecki, Leibo, Kuttler, Zisserman, Simonyan,
  et~al.]{schmitt2018kickstarting}
Schmitt, S., Hudson, J.~J., Zidek, A., Osindero, S., Doersch, C., Czarnecki,
  W.~M., Leibo, J.~Z., Kuttler, H., Zisserman, A., Simonyan, K., et~al.
\newblock Kickstarting deep reinforcement learning.
\newblock \emph{arXiv preprint arXiv:1803.03835}, 2018.

\bibitem[Schulman et~al.(2017)Schulman, Wolski, Dhariwal, Radford, and
  Klimov]{schulman2017proximal}
Schulman, J., Wolski, F., Dhariwal, P., Radford, A., and Klimov, O.
\newblock Proximal policy optimization algorithms.
\newblock \emph{arXiv preprint arXiv:1707.06347}, 2017.

\bibitem[Schwarzer et~al.(2023)Schwarzer, Ceron, Courville, Bellemare, Agarwal,
  and Castro]{schwarzer2023bigger}
Schwarzer, M., Ceron, J. S.~O., Courville, A., Bellemare, M.~G., Agarwal, R.,
  and Castro, P.~S.
\newblock Bigger, better, faster: Human-level atari with human-level
  efficiency.
\newblock In \emph{International Conference on Machine Learning}, pp.\
  30365--30380. PMLR, 2023.

\bibitem[Sinha et~al.(2020)Sinha, Bharadhwaj, Srinivas, and
  Garg]{sinha2020d2rl}
Sinha, S., Bharadhwaj, H., Srinivas, A., and Garg, A.
\newblock D2rl: Deep dense architectures in reinforcement learning.
\newblock \emph{arXiv preprint arXiv:2010.09163}, 2020.

\bibitem[Sokar et~al.(2021)Sokar, Mocanu, Mocanu, Pechenizkiy, and
  Stone]{sokar2021dynamic}
Sokar, G., Mocanu, E., Mocanu, D.~C., Pechenizkiy, M., and Stone, P.
\newblock Dynamic sparse training for deep reinforcement learning.
\newblock \emph{arXiv preprint arXiv:2106.04217}, 2021.

\bibitem[Sokar et~al.(2023)Sokar, Agarwal, Castro, and Evci]{sokar2023dormant}
Sokar, G., Agarwal, R., Castro, P.~S., and Evci, U.
\newblock The dormant neuron phenomenon in deep reinforcement learning.
\newblock In Krause, A., Brunskill, E., Cho, K., Engelhardt, B., Sabato, S.,
  and Scarlett, J. (eds.), \emph{Proceedings of the 40th International
  Conference on Machine Learning}, volume 202 of \emph{Proceedings of Machine
  Learning Research}, pp.\  32145--32168. PMLR, 23--29 Jul 2023.
\newblock URL \url{https://proceedings.mlr.press/v202/sokar23a.html}.

\bibitem[Song et~al.(2019)Song, Jiang, Tu, Du, and
  Neyshabur]{song2019observational}
Song, X., Jiang, Y., Tu, S., Du, Y., and Neyshabur, B.
\newblock Observational overfitting in reinforcement learning.
\newblock \emph{arXiv preprint arXiv:1912.02975}, 2019.

\bibitem[Sutton(1988)]{sutton88learning}
Sutton, R.~S.
\newblock Learning to predict by the methods of temporal differences.
\newblock \emph{Machine Learning}, 3\penalty0 (1):\penalty0 9–44, August
  1988.

\bibitem[Taiga et~al.(2023)Taiga, Agarwal, Farebrother, Courville, and
  Bellemare]{taiga2023investigating}
Taiga, A.~A., Agarwal, R., Farebrother, J., Courville, A., and Bellemare, M.~G.
\newblock Investigating multi-task pretraining and generalization in
  reinforcement learning.
\newblock In \emph{The Eleventh International Conference on Learning
  Representations}, 2023.

\bibitem[Tan et~al.(2023)Tan, Hu, Pan, Huang, and Huang]{tan2023rlx}
Tan, Y., Hu, P., Pan, L., Huang, J., and Huang, L.
\newblock {RL}x2: Training a sparse deep reinforcement learning model from
  scratch.
\newblock In \emph{The Eleventh International Conference on Learning
  Representations}, 2023.
\newblock URL \url{https://openreview.net/forum?id=DJEEqoAq7to}.

\bibitem[Todorov et~al.(2012)Todorov, Erez, and Tassa]{todorov2012mujoco}
Todorov, E., Erez, T., and Tassa, Y.
\newblock Mujoco: A physics engine for model-based control.
\newblock In \emph{2012 IEEE/RSJ international conference on intelligent robots
  and systems}, pp.\  5026--5033. IEEE, 2012.

\bibitem[van Hasselt et~al.(2016)van Hasselt, Guez, and
  Silver]{hasselt2015doubledqn}
van Hasselt, H., Guez, A., and Silver, D.
\newblock Deep reinforcement learning with double q-learning.
\newblock In \emph{Proceedings of the Thirthieth AAAI Conference On Artificial
  Intelligence (AAAI), 2016}, 2016.
\newblock cite arxiv:1509.06461Comment: AAAI 2016.

\bibitem[Van~Hasselt et~al.(2018)Van~Hasselt, Doron, Strub, Hessel, Sonnerat,
  and Modayil]{van2018deep}
Van~Hasselt, H., Doron, Y., Strub, F., Hessel, M., Sonnerat, N., and Modayil,
  J.
\newblock Deep reinforcement learning and the deadly triad.
\newblock \emph{arXiv preprint arXiv:1812.02648}, 2018.

\bibitem[Van~Hasselt et~al.(2019)Van~Hasselt, Hessel, and
  Aslanides]{van2019use}
Van~Hasselt, H.~P., Hessel, M., and Aslanides, J.
\newblock When to use parametric models in reinforcement learning?
\newblock \emph{Advances in Neural Information Processing Systems}, 32, 2019.

\bibitem[Van~Rossum \& Drake~Jr(1995)Van~Rossum and Drake~Jr]{van1995python}
Van~Rossum, G. and Drake~Jr, F.~L.
\newblock \emph{Python reference manual}.
\newblock Centrum voor Wiskunde en Informatica Amsterdam, 1995.

\bibitem[Vieillard et~al.(2020)Vieillard, Pietquin, and
  Geist]{vieillard2020munchausen}
Vieillard, N., Pietquin, O., and Geist, M.
\newblock Munchausen reinforcement learning.
\newblock In Larochelle, H., Ranzato, M., Hadsell, R., Balcan, M., and Lin, H.
  (eds.), \emph{Advances in Neural Information Processing Systems}, volume~33,
  pp.\  4235--4246. Curran Associates, Inc., 2020.

\bibitem[Vinyals et~al.(2019)Vinyals, Babuschkin, Czarnecki, Mathieu, Dudzik,
  Chung, Choi, Powell, Ewalds, Georgiev, et~al.]{vinyals2019grandmaster}
Vinyals, O., Babuschkin, I., Czarnecki, W.~M., Mathieu, M., Dudzik, A., Chung,
  J., Choi, D.~H., Powell, R., Ewalds, T., Georgiev, P., et~al.
\newblock Grandmaster level in starcraft ii using multi-agent reinforcement
  learning.
\newblock \emph{Nature}, 575\penalty0 (7782):\penalty0 350--354, 2019.

\bibitem[Vischer et~al.(2021)Vischer, Lange, and Sprekeler]{vischer2021lottery}
Vischer, M., Lange, R.~T., and Sprekeler, H.
\newblock On lottery tickets and minimal task representations in deep
  reinforcement learning.
\newblock In \emph{International Conference on Learning Representations}, 2021.

\bibitem[Wang et~al.(2016)Wang, Schaul, Hessel, Hasselt, Lanctot, and
  Freitas]{wang16dueling}
Wang, Z., Schaul, T., Hessel, M., Hasselt, H., Lanctot, M., and Freitas, N.
\newblock Dueling network architectures for deep reinforcement learning.
\newblock In \emph{Proceedings of the 33rd International Conference on Machine
  Learning}, volume~48, pp.\  1995--2003, 2016.

\bibitem[Yarats et~al.(2021)Yarats, Fergus, and Kostrikov]{yarats2021image}
Yarats, D., Fergus, R., and Kostrikov, I.
\newblock Image augmentation is all you need: Regularizing deep reinforcement
  learning from pixels.
\newblock In \emph{9th International Conference on Learning Representations,
  ICLR 2021}, 2021.

\bibitem[Yu et~al.(2019)Yu, Edunov, Tian, and Morcos]{yu2019playing}
Yu, H., Edunov, S., Tian, Y., and Morcos, A.~S.
\newblock Playing the lottery with rewards and multiple languages: lottery
  tickets in rl and nlp.
\newblock In \emph{International Conference on Learning Representations}, 2019.

\bibitem[Zhang et~al.(2018)Zhang, Vinyals, Munos, and Bengio]{zhang2018study}
Zhang, C., Vinyals, O., Munos, R., and Bengio, S.
\newblock A study on overfitting in deep reinforcement learning.
\newblock \emph{arXiv preprint arXiv:1804.06893}, 2018.

\bibitem[Zhang et~al.(2019)Zhang, He, and Li]{zhang2019accelerating}
Zhang, H., He, Z., and Li, J.
\newblock Accelerating the deep reinforcement learning with neural network
  compression.
\newblock In \emph{2019 International Joint Conference on Neural Networks
  (IJCNN)}, pp.\  1--8. IEEE, 2019.

\bibitem[Zhu \& Gupta(2017)Zhu and Gupta]{zhu2017prune}
Zhu, M. and Gupta, S.
\newblock To prune, or not to prune: exploring the efficacy of pruning for
  model compression, 2017.

\end{thebibliography}
\bibliographystyle{icml2024}

%%%%%%%%%%%%%%%%%%%%%%%%%%%%%%%%%%%%%%%%%%%%%%%%%%%%%%%%%%%%%%%%%%%%%%%%%%%%%%%
%%%%%%%%%%%%%%%%%%%%%%%%%%%%%%%%%%%%%%%%%%%%%%%%%%%%%%%%%%%%%%%%%%%%%%%%%%%%%%%
% APPENDIX
%%%%%%%%%%%%%%%%%%%%%%%%%%%%%%%%%%%%%%%%%%%%%%%%%%%%%%%%%%%%%%%%%%%%%%%%%%%%%%%
%%%%%%%%%%%%%%%%%%%%%%%%%%%%%%%%%%%%%%%%%%%%%%%%%%%%%%%%%%%%%%%%%%%%%%%%%%%%%%%
\newpage
\appendix
\onecolumn

% \section{You \emph{can} have an appendix here.}
\section{Code availability}
\label{apen:code_availability}

Our experiments were built on open source code, mostly from the Dopamine repository. The root
directory for these is \href{https://github.com/google/dopamine/tree/master/dopamine/}{https://github.com/google/dopamine/tree/master/dopamine/}, and we specify the subdirectories below (with clickable links):

\begin{itemize}
        \item DQN and Rainbow agents from \href{https://github.com/google/dopamine/tree/master/dopamine/jax/agents}{/jax/agents/} 
        \item Atari-100k agents from \href{https://github.com/google/dopamine/tree/master/dopamine/labs/atari_100k}{/labs/atari-100k/}
        \item Sparsity scripts from JaxPruner \href{https://github.com/google-research/jaxpruner/tree/main/jaxpruner}{/jaxpruner/baselines/dopamine/}
        \item Resnet architecture from \href{https://github.com/google/dopamine/blob/master/dopamine/labs/offline_rl/jax/networks.py#L108}{/labs/offline-rl/jax/networks.py} \textbf{(line 108)}
        \item Dormant neurons metric, Reset, ReDo and Weight Decay from \href{https://github.com/google/dopamine/tree/master/dopamine/labs/redo}{/labs/redo/}
        {\item SAC and PPO experiments from \href{https://github.com/google-research/rigl/tree/master/rigl/rl}{/google-research/rigl/rl/}}
\end{itemize}

For the srank metric experiments we used code from: \\ \href{https://github.com/google-research/google-research/blob/master/generalization_representations_rl_aistats22/coherence/coherence_compute.py}{https://github.com/google-research/google-research/blob/master/\\     generalization\_representations\_rl\_aistats22/coherence/coherence\_compute.py}

\section{Atari Game Selection}
\label{apen:game_selection}
Most of our experiments were run with $15$ games from the ALE suite \citep{Bellemare_2013}, as suggested by \citet{graesser2022state}. However, for the Atari $100$k agents (\autoref{secc:samplefficient}), we used the standard set of $26$ games \citep{Kaiser2020Model} to be consistent with the benchmark. We also ran some experiments with the full set of $60$ games. The specific games are detailed below.

\textbf{15 game subset:} MsPacman, Pong, Qbert, (Assault, Asterix, BeamRider, Boxing,
Breakout, CrazyClimber, DemonAttack, Enduro, FishingDerby, SpaceInvaders, Tutankham, VideoPinball. According to \cite{graesser2022state}, these games were selected to be roughly evenly distributed amongst the games ranked by DQN’s human normalized score in \citep{mnih2015humanlevel} with a lower cut off of approximately $100\%$ of human performance.

\textbf{26 game subset:} Alien, Amidar, Assault, Asterix, BankHeist, BattleZone, Boxing, Breakout, ChopperCommand, CrazyClimber, DemonAttack, Freeway, Frostbite, Gopher, Hero, Jamesbond, Kangaroo, Krull, KungFuMaster, MsPacman, Pong, PrivateEye, Qbert, RoadRunner, Seaquest, UpNDown.

\textbf{60 game set:} The 26 games above in addition to: AirRaid, Asteroids, Atlantis, BeamRider, Berzerk, Bowling, Carnival, Centipede, DoubleDunk, ElevatorAction, Enduro, FishingDerby, Gravitar, IceHockey, JourneyEscape, MontezumaRevenge, NameThisGame, Phoenix, Pitfall, Pooyan, Riverraid, Robotank, Skiing, Solaris, SpaceInvaders, StarGunner, Tennis, TimePilot, Tutankham, Venture, VideoPinball, WizardOfWor, YarsRevenge, Zaxxon.

\section{Sparsity Levels}
\label{appen:sparsitylevels}

\begin{figure}[h]
    \centering
    \includegraphics[width=0.25\textwidth]{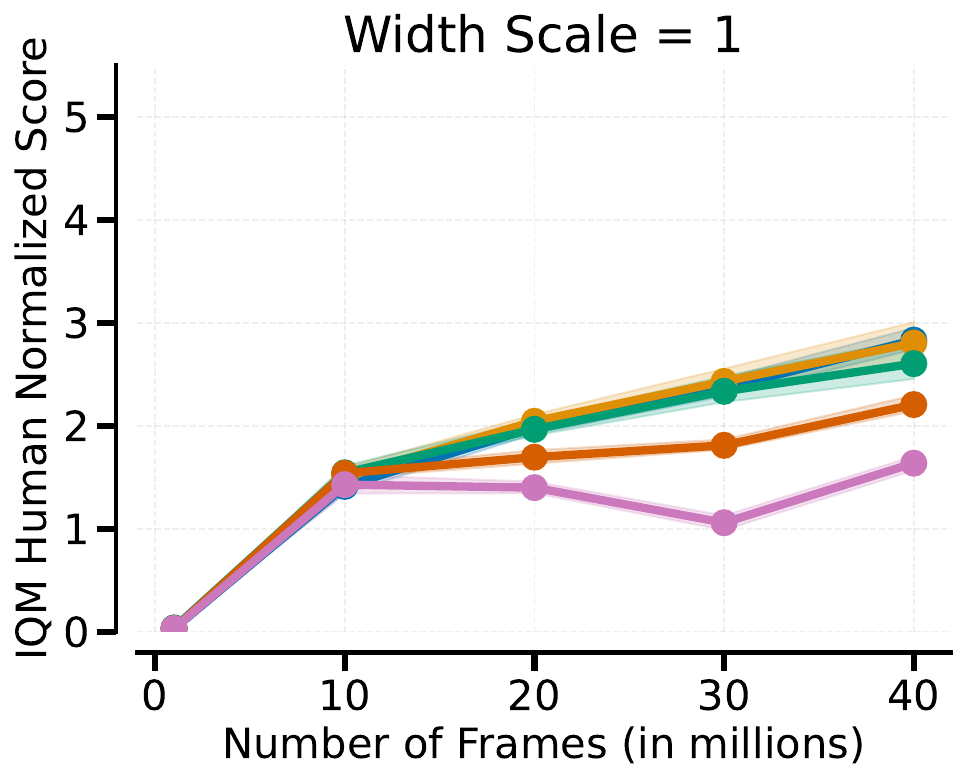}%
    \includegraphics[width=0.235\textwidth]{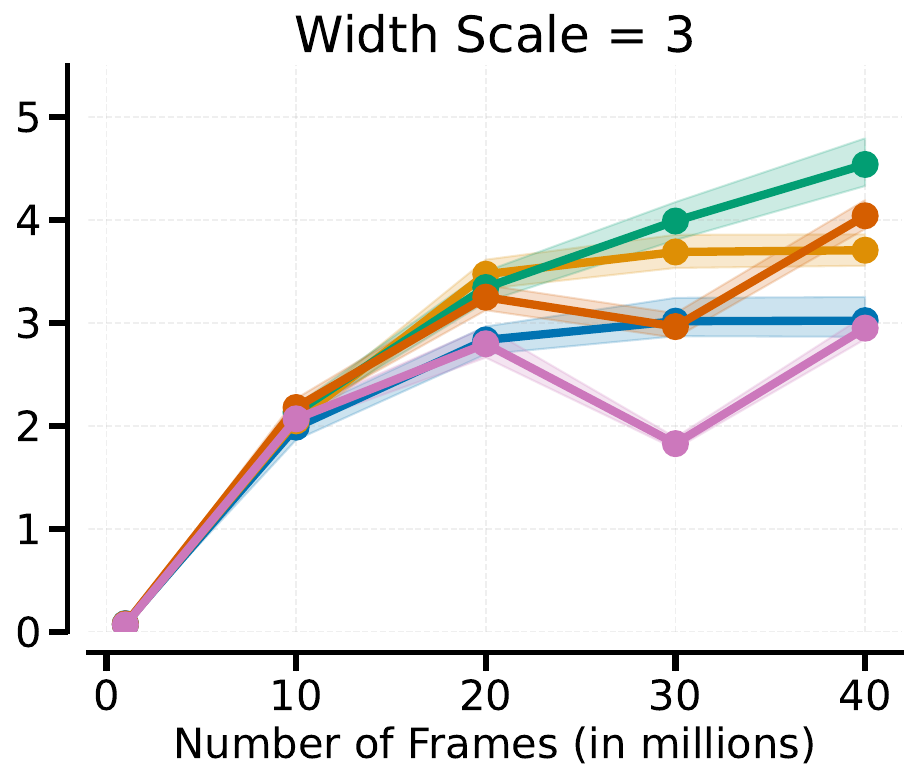}%
    \vline height 100pt depth 0 pt width 1.2 pt
    \includegraphics[width=0.24\textwidth]{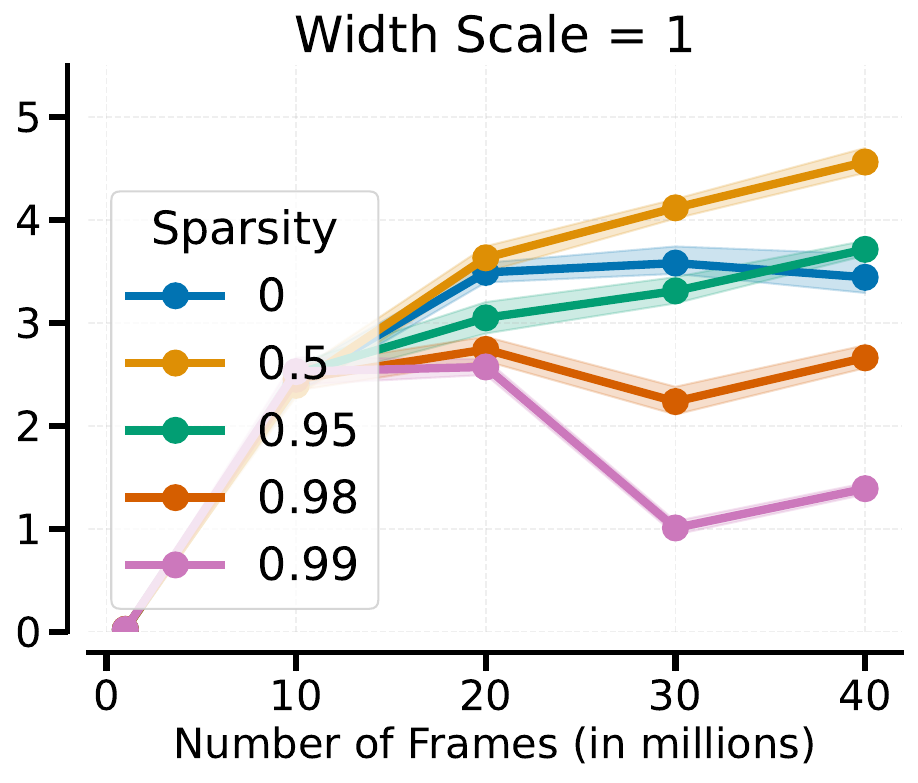}%
    \includegraphics[width=0.24\textwidth]{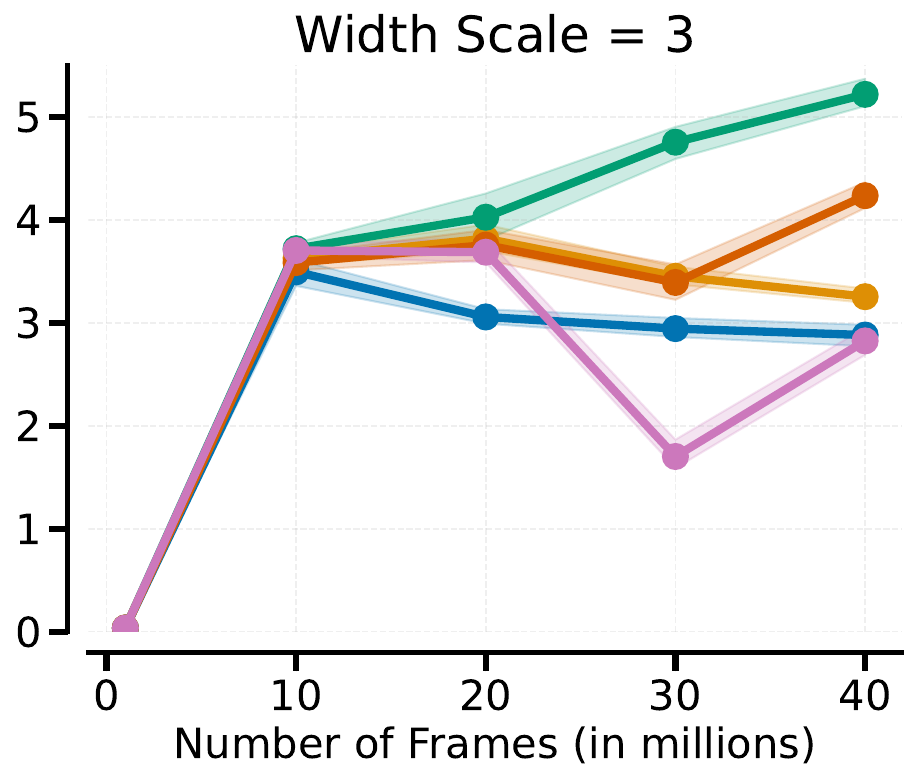}%
    \vspace{-0.1cm}
    \caption{\textbf{
    Evaluating how varying the sparsity parameter affects performance} for a given architecture on DQN \citep{mnih2015humanlevel} and Rainbow agent. We report results aggregated IQM of human-normalized scores over $15$ games.}
    \label{fig:sparisty_values}
    \vspace{-0.2cm}
\end{figure}

\begin{figure}[!h]
    \centering
    \includegraphics[width=0.45\textwidth]{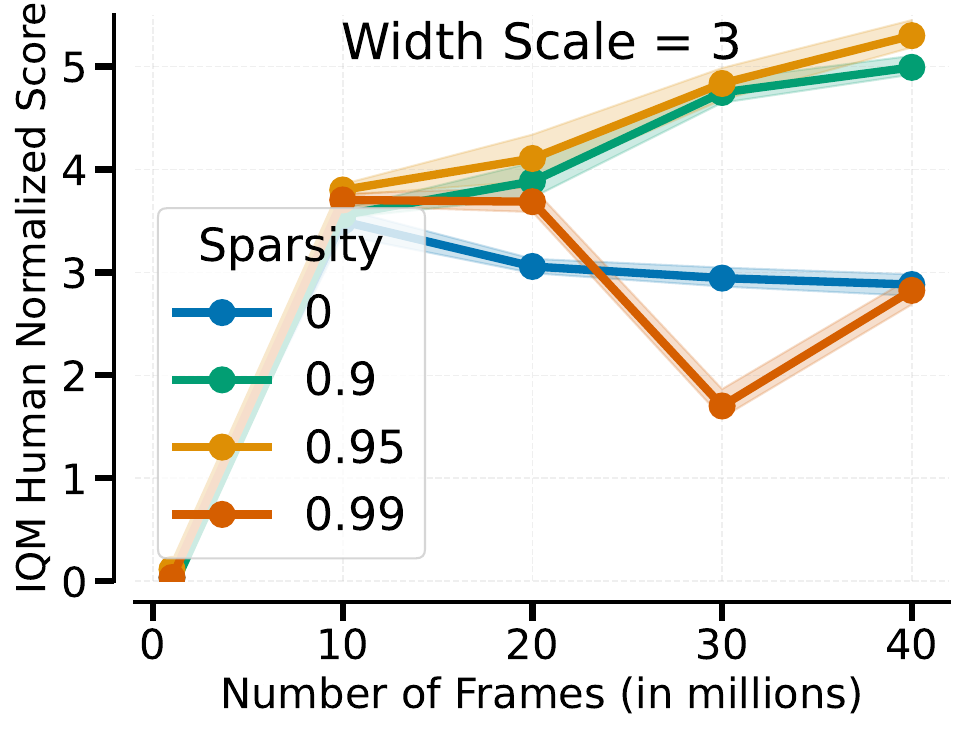}%
    \vspace{-0.1cm}
    \caption{\textbf{
    Evaluating how varying the sparsity parameter affects performance} for a given architecture, resnet with a width multiplier of $3$, on Rainbow agent. We report results aggregated IQM of human-normalized scores over $15$ games.}
    \label{fig:sparsity_values_rainbow}
    \vspace{-0.2cm}
\end{figure}

\newpage
\section{Scaling Replay Ratios}
\label{appen:scalingreplayratios}

\begin{figure}[!h]
    \centering
    \includegraphics[width=0.48\textwidth]{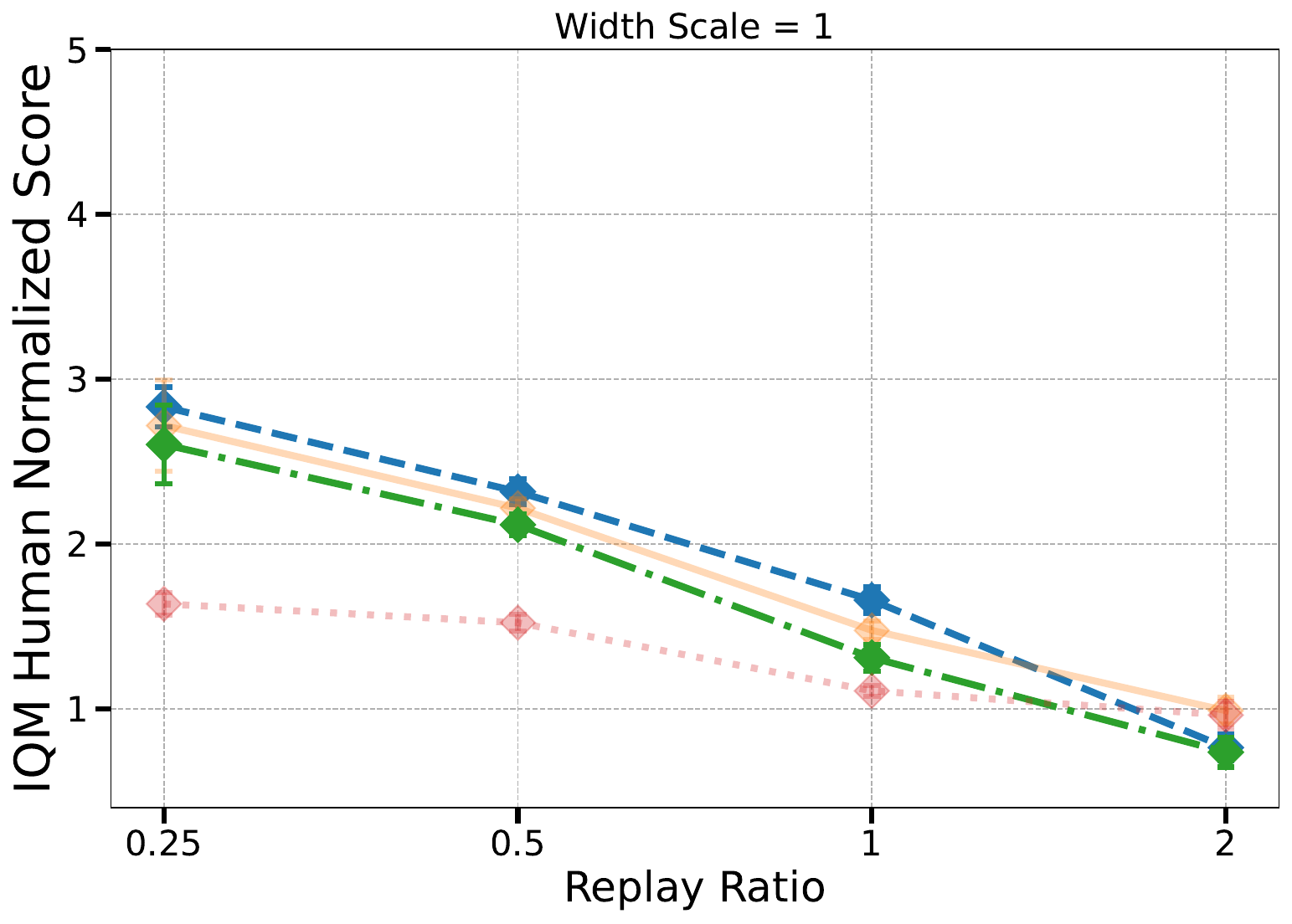}%
    \includegraphics[width=0.46\textwidth]{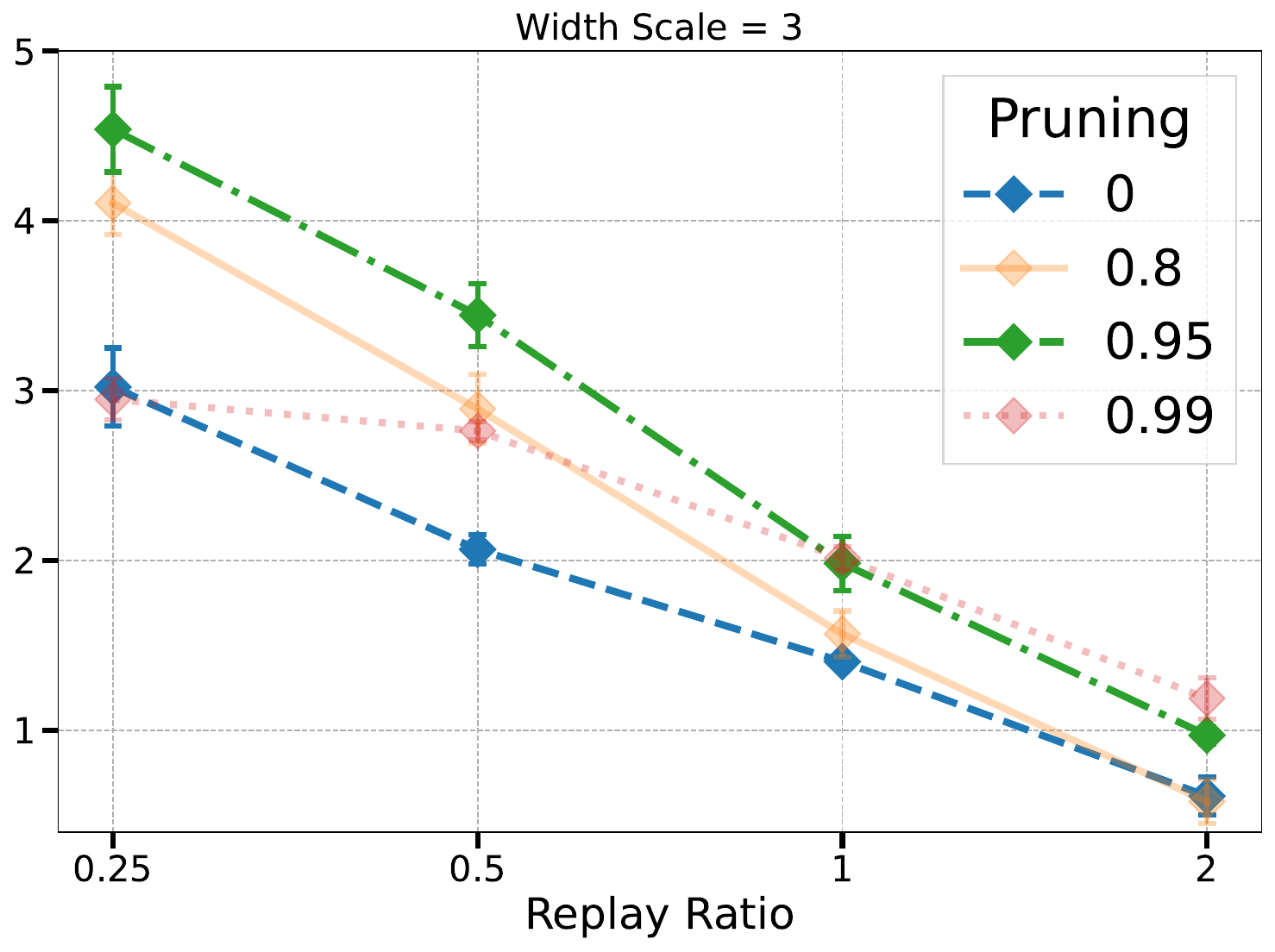}%
    \vspace{-0.1cm}
    \caption{\textbf{Scaling replay ratio for resnet architecture}  (default is $0.25$), for a width factor of $1$ \textbf{(left)} and a width factor of $3$ \textbf{(right)} using DQN agent. We report interquantile mean performance with error bars indicating 95\% confidence intervals. On the x-axis we report the replay ratio value.}
    \label{fig:scalingReplayRatioBoth}
    \vspace{-0.2cm}
\end{figure}

\section{MuJoCo environments}
\label{appen:sac_threeEnv}

\begin{figure}[!h]
    \includegraphics[width=\textwidth]{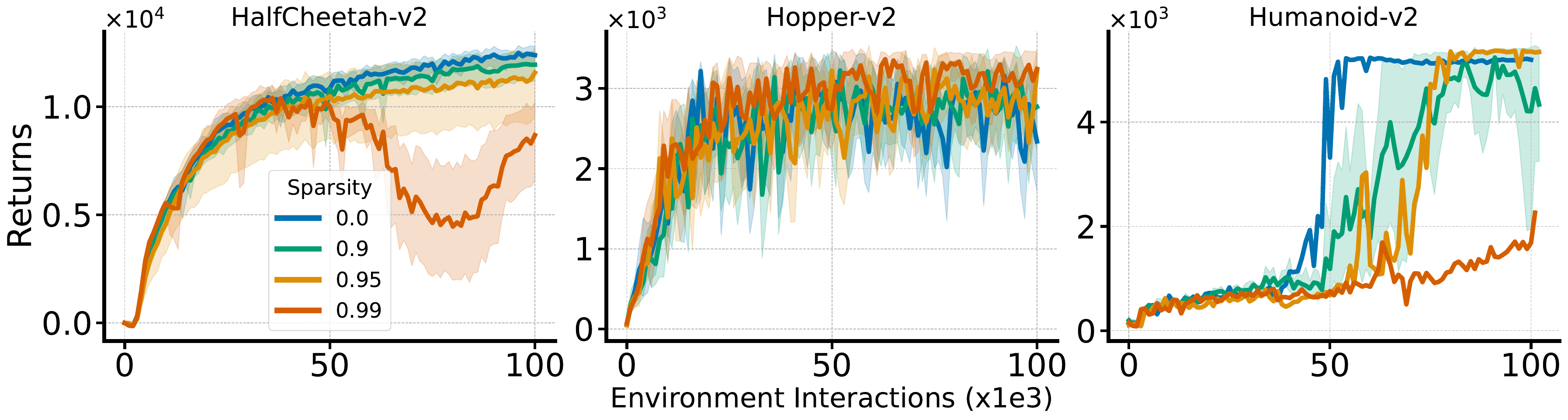}
     \vspace{-0.7cm}
     \caption{\textbf{Evaluating how varying the sparsity parameter affects performance of SAC} on three MuJoCo environments when increasing width x$5$. We report returns over 10 runs for each experiment.}
    \label{fig:scalingReplayRatioBoth}
    \vspace{-0.2cm}
\end{figure}

\newpage
{
\section{Hyper-parameters list}
\label{sec:list_hyperparameters}

Default hyper-parameter settings for DER \citep{van2019use} and DrQ($\epsilon$) \citep{Kaiser2020Model, agarwal2021deep}. \autoref{tbl:defaultvalues} shows the default values for each hyper-parameter across all the Atari games.

\begin{table}[!h]
 \centering
  \caption{Default hyper-parameters setting for DER and DrQ($\epsilon$) agents.}
  \label{tbl:defaultvalues}
 \begin{tabular}{@{} ccc @{}}
  %\toprule
    \toprule
    & \multicolumn{2}{c}{Atari}\\
    \cmidrule(lr){2-3}
  Hyper-parameter &  DER & DrQ($\epsilon$) \\
    \midrule
     Adam's($\epsilon$) & 0.00015 & 0.00015\\
     Adam's learning rate & 0.0001 & 0.0001 \\
     Batch Size & 32 & 32\\
     Conv. Activation Function & ReLU & ReLU \\
     Convolutional Width & 1& 1\\
     Dense Activation Function & ReLU & ReLU\\
     Dense Width & 512 & 512 \\
     Normalization & None & None \\
     Discount Factor & 0.99 & 0.99 \\
     Exploration $\epsilon$ & 0.01 & 0.01\\
     Exploration $\epsilon$ decay & 2000 & 5000\\
     Minimum Replay History & 1600 & 1600\\
     Number of Atoms & 51 & 0 \\
     Number of Convolutional Layers & 3 & 3\\
     Number of Dense Layers & 2 & 2\\
     Replay Capacity & 1000000 & 1000000 \\
     Reward Clipping & True & True \\
     Update Horizon & 10 & 10 \\
     Update Period & 1& 1\\
     Weight Decay & 0 & 0\\
     Sticky Actions & False & False \\
     \bottomrule
  \end{tabular}
\end{table}

\newpage
Default hyper-parameter settings for DQN \citep{mnih2015humanlevel}, Rainbow \citep{Hessel2018RainbowCI}, IQN \citep{dabney2018implicit}, Munchausen-IQN \citep{vieillard2020munchausen}. \autoref{tbl:defaultvalues_40M} shows the default values for each hyper-parameter across all the Atari games.

\begin{table}[!h]
 \centering
  \caption{Default hyper-parameters setting for DQN, Rainbow, IQN, Munchausen-IQN agents.}
  \label{tbl:defaultvalues_40M}
 \begin{tabular}{@{} ccccc @{}}
  %\toprule
    \toprule
    & \multicolumn{2}{c}{Atari}\\
    \cmidrule(lr){2-5}
  Hyper-parameter &  DQN & Rainbow & IQN & M-IQN\\
    \midrule
     Adam's ($\epsilon$) & 1.5e-4 & 1.5e-4 & 3.125e-4 & 3.125e-4 \\
     Adam's learning rate &  6.25e-5 & 6.25e-5 & 5e-5 & 5e-5 \\
     Batch Size & 32 & 32 & 32 & 32\\
     Conv. Activation Function & ReLU & ReLU & ReLU & ReLU \\
     Convolutional Width & 1 & 1 & 1 & 1\\
     Dense Activation Function & ReLU & ReLU & ReLU & ReLU\\
     Dense Width & 512 & 512 & 512 & 512 \\
     Normalization & None & None & None & None \\
     Discount Factor & 0.99 & 0.99 & 0.99 & 0.99 \\
     Exploration $\epsilon$ & 0.01 & 0.01 & 0.01 & 0.01 \\
     Exploration $\epsilon$ decay & 250000 & 250000 & 250000 & 250000\\
     Minimum Replay History & 20000 & 20000 & 20000 & 20000\\
     Number of Atoms & 0 & 51 &- &- \\
     Kappa & - & - & 1.0 &1.0\\
     Num tau samples &  -& - & 64 & 64\\
     Num tau prime samples  & - & -& 64 & 64\\
     Num quantile samples & - &  -& 32 & 32\\
     Number of Convolutional Layers & 3 & 3 & 3 & 3 \\
     Number of Dense Layers & 2 & 2 & 2 & 2 \\
     Replay Capacity & 1000000 & 1000000 & 1000000 & 1000000 \\
     Reward Clipping & True & True & True & True \\
     Update Horizon & 1 & 3 & 3 & 3 \\
     Update Period & 4 & 4  & 4 & 4 \\
     Weight Decay & 0 & 0 & 0 & 0\\
     Sticky Actions & True & True & True & True \\
     Tau & - & - & 0 & 0.03 \\
     \bottomrule
  \end{tabular}
\end{table}

\newpage
Default hyper-parameter settings for CQL \citep{kumar2020conservative} and CQL+C51 \citep{kumar2022offline} offline agents. \autoref{tbl:defaultvalues_offline} shows the default values for each hyper-parameter across all the Atari games.

\begin{table}[!h]
 \centering
  \caption{Default hyper-parameters setting for CQL and CQL+C51 agents.}
  \label{tbl:defaultvalues_offline}
 \begin{tabular}{@{} ccc @{}}
  %\toprule
    \toprule
    & \multicolumn{2}{c}{Atari}\\
    \cmidrule(lr){2-3}
  Hyper-parameter &  CQL & CQL+C51\\
    \midrule
     Adam's($\epsilon$) & 0.0003125 & 0.00015\\
     Batch Size & 32 & 32\\
     Conv. Activation Function & ReLU & ReLU \\
     Convolutional Width & 1& 1\\
     Dense Activation Function & ReLU & ReLU\\
     Normalization & None & None \\
     Dense Width & 512 & 512 \\
     Discount Factor & 0.99 & 0.99 \\
     Learning Rate & 0.00005 & 0.0000625 \\
     Number of Atoms & 0 & 51 \\
     Number of Convolutional Layers & 3 & 3\\
     Number of Dense Layers & 2 & 2\\
     Fixed Replay Capacity & 2,500,000 steps & 2,500,000 steps \\
     Reward Clipping & True & True \\
     Update Horizon & 1 & 3 \\
     Update Period & 1 & 1\\
     Weight Decay & 0 & 0\\
     Replay Scheme & Uniform & Uniform \\
     Dueling & False & True \\
     Double DQN & False & True \\
     CQL coef & 0.1 & 0.1\\
     \bottomrule
  \end{tabular}
\end{table}

Default hyper-parameter settings for CNN architecture \citep{mnih2015humanlevel} and Impala-based ResNet \citep{espeholt2018impala} \autoref{tbl:defaultvalues_networ} shows the default values for each hyper-parameter across all the Atari games.

\begin{table}[!h]
 \centering
  \caption{Default hyper-parameters for neural networks.}
  \label{tbl:defaultvalues_networ}
 \begin{tabular}{@{} ccc @{}}
  %\toprule
    \toprule
    & \multicolumn{2}{c}{Atari}\\
    \cmidrule(lr){2-3}
  Hyper-parameter &  CNN architecture \citep{mnih2015humanlevel} & Impala-based ResNet \citep{espeholt2018impala}\\
    \midrule
    %  Grey-scaling  &True \\
    Observation down-sampling & (84, 84) & (84, 84) \\
    Frames stacked &4  &4 \\
    %  Frame skip (Action repetitions) & 4 \\
     Q-network (channels) & 32, 64, 64 & 32, 64, 64\\
     Q-network (filter size) & 8 x 8, 4 x 4, 3 x 3 & 8 x 8, 4 x 4, 3 x 3\\
     Q-network (stride)  & 4, 2, 1 & 4, 2, 1 \\
     Num blocks & - & 2 \\
     Use max pooling & False & True \\
     Skip connections & False & True \\
     Hardware & Tesla P100 GPU & Tesla P100 GPU\\
     \bottomrule
  \end{tabular}
\end{table}

\newpage
\section{Additional experiments}
Unless otherwise specified, in all experiments below we report the interquantile mean after 40 million environment steps, aggregated over 15 games with 5 seeds each; error bars indicate 95\% stratified bootstrap confidence intervals \citep{agarwal2021deep}.

\subsection{Experiments with PPO on MuJoCo}
\label{sec:ppoExperiments}

We used the PPO implementation from \citep{graesser2022state} and ran some initial experiments with MuJoCo, increasing the width by 5x. As with our SAC experiments, we see no real change in performance, with perhaps some mild gains in Humanoid-v2. One reason why we may not see performance improvements in neither SAC nor PPO is that in ALE experiments, all agents use Convolutional layers, whereas for the MuJoCo experiments (where we ran SAC and PPO) the networks only use dense layers.

Nonetheless, it is worth noting that \citet{graesser2022state} saw degradation with pruning at width=1 (Figure 16 in their paper), with almost a total collapse at 99\% sparsity. In contrast, our results with 5x width shows strong performance even at 99\% sparsity.

\begin{figure*}[!h]
    \centering
    \includegraphics[width=\textwidth]{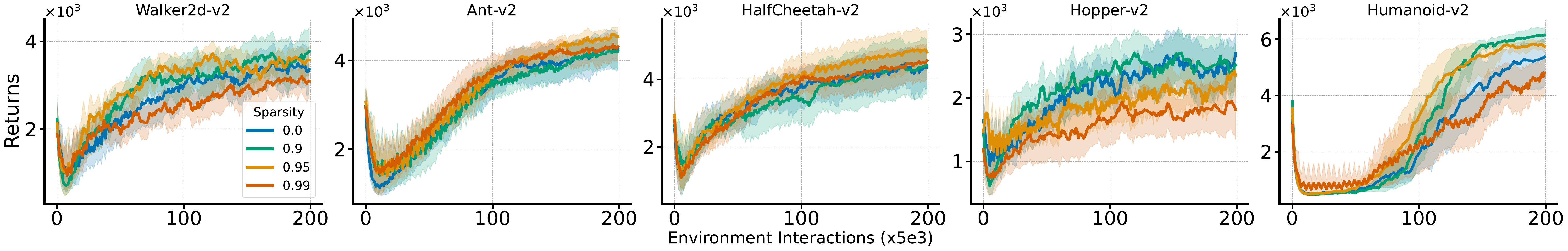}
    %\vspace{-0.1cm}
    \caption{\textbf{Proximal Policy Optimization (PPO)} \citep{schulman2017proximal} on MuJoCo environments when increasing width x5. We report returns over 10 runs for each experiment.}
    \label{fig:}
    \vspace{-0.2cm}
\end{figure*}

\subsection{Experiments with IQN and M-IQN}
\label{sec:iqn}

While Rainbow is still a competitive agent in the ALE and both DQN and Rainbow are still regularly used as baselines in recent works, exploring newer agents is a reasonable request. To address this, we ran experiments with Implicit Quantile Networks (IQN) \citep{dabney2018implicit} and Munchausen-IQN \citep{vieillard2020munchausen} with widths of 1 and 3; consistent with our submission’s findings, we observe significant gains when using pruning.

\begin{figure}[!h]
    \centering
    \includegraphics[width=0.4\textwidth]{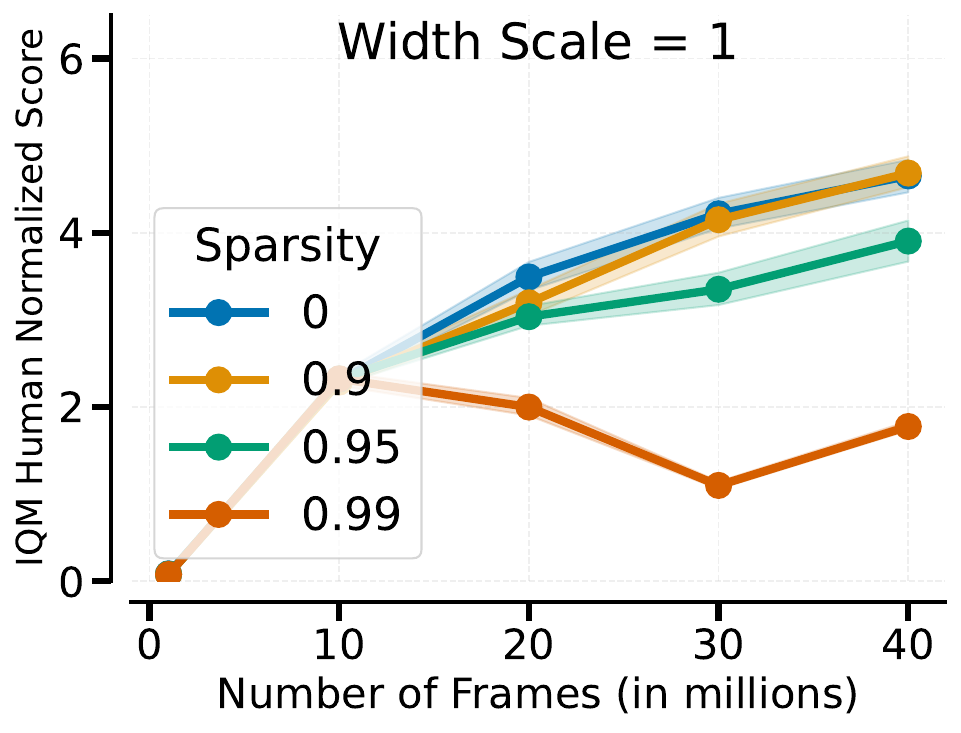}%
    \includegraphics[width=0.4\textwidth]{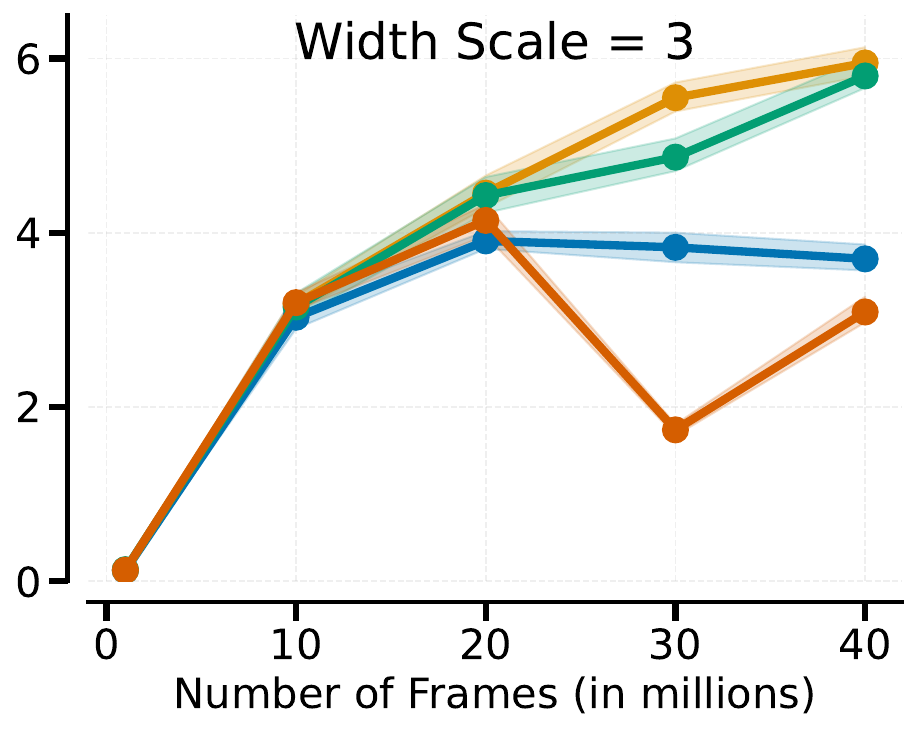}%
    \caption{\textbf{IQN (\textit{Implicit Quantile Networks})} \citep{dabney2018implicit} with ResNet architecture (with a width factor of $1$ and $3$).}
    % See \cref{sec:setup} for training details
    \label{fig:}
    \vspace{-0.2cm}
\end{figure}

\begin{figure}[!h]
    \centering
    \includegraphics[width=0.42\textwidth]{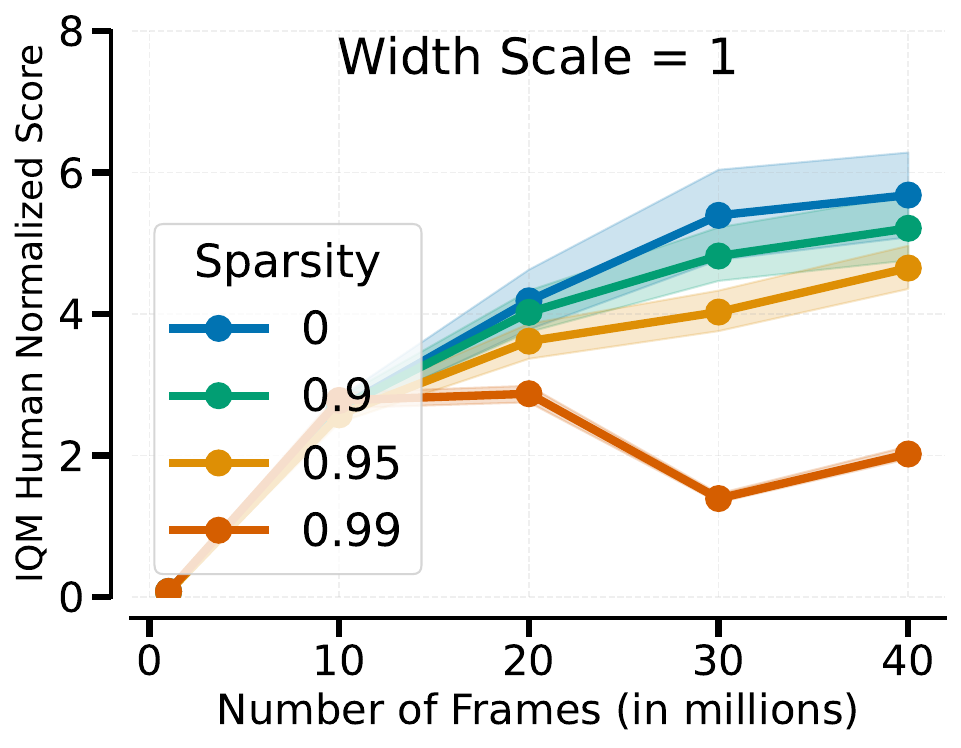}%
    \includegraphics[width=0.4\textwidth]{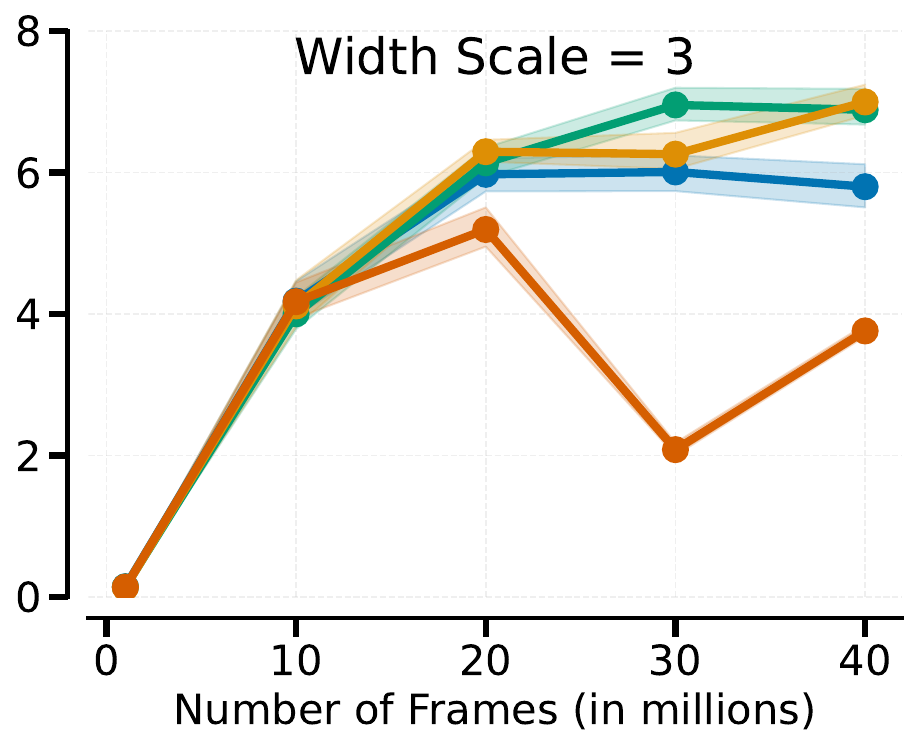}%
    %\vspace{-0.1cm}
    \caption{\textbf{M-IQN (\textit{Munchausen-Implicit Quantile Networks})} \citep{vieillard2020munchausen} with ResNet architecture (with a width factor of $1$ and $3$).}
    \label{fig:}
    \vspace{-0.2cm}
\end{figure}

\newpage
\subsection{Comparison with RigL}
\label{sec:rigl}

We have run a comparison with RigL \citep{evci2020rigging}. While RigL can be somewhat effective, it is unable to match the performance of pruning.

\begin{figure}[!h]
    \centering
    \includegraphics[width=0.42\textwidth]{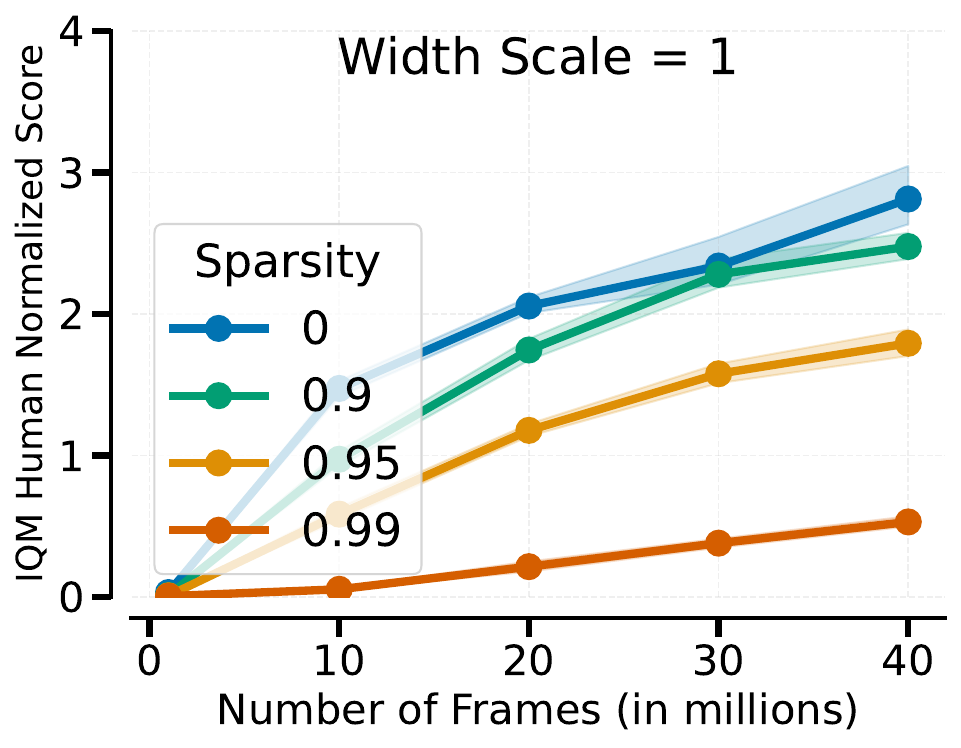}%
    \includegraphics[width=0.4\textwidth]{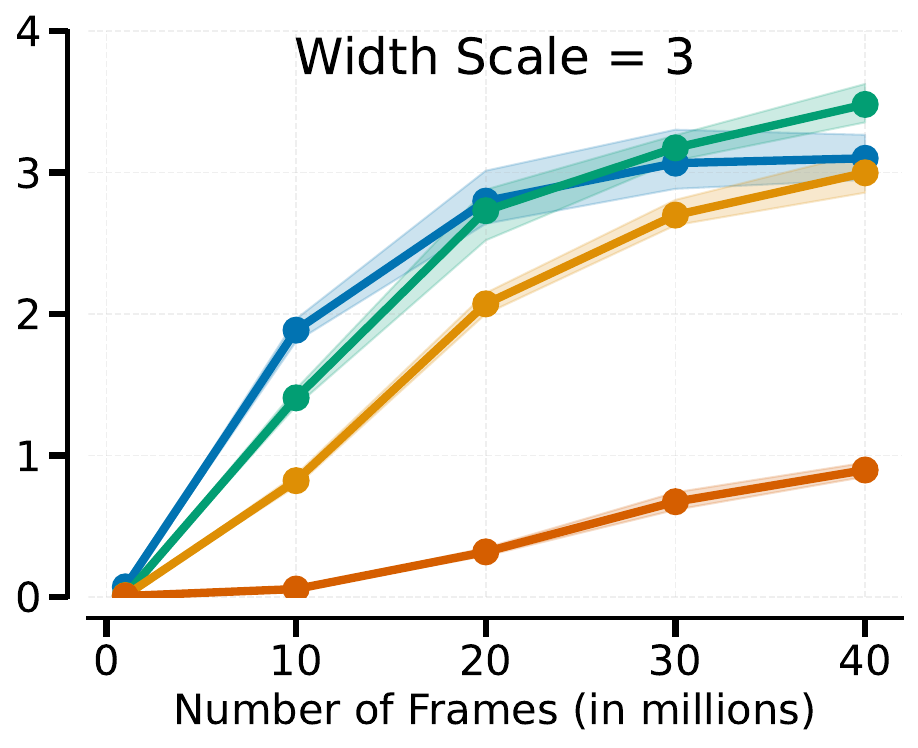}%
    % \caption{Evaluating how varying sparsity affects performance for DQN with the ResNet architecture and a width multiplier of 1 and 3, when using RigL \citep{evci2020rigging}.}
    \caption{\textbf{The Rigged Lottery (RigL)} \citep{evci2020rigging} for DQN with ResNet architecture (with a width factor of $1$ and $3$).}
    \label{fig:}
    \vspace{-0.2cm}
\end{figure}

\subsection{Varying Adam's $\epsilon$}
\label{sec:epsilonSweep}

The default value for Adam’s $\epsilon$ is $1.5e-5$; we ran experiments by dividing/multiplying this value by $3$ ( $5e-5$ and $4.5e-4$, respectively). In all these cases, pruning maintains a significant advantage over the dense baseline.

\begin{figure*}[!h]
    \centering
    \includegraphics[width=0.26\textwidth]{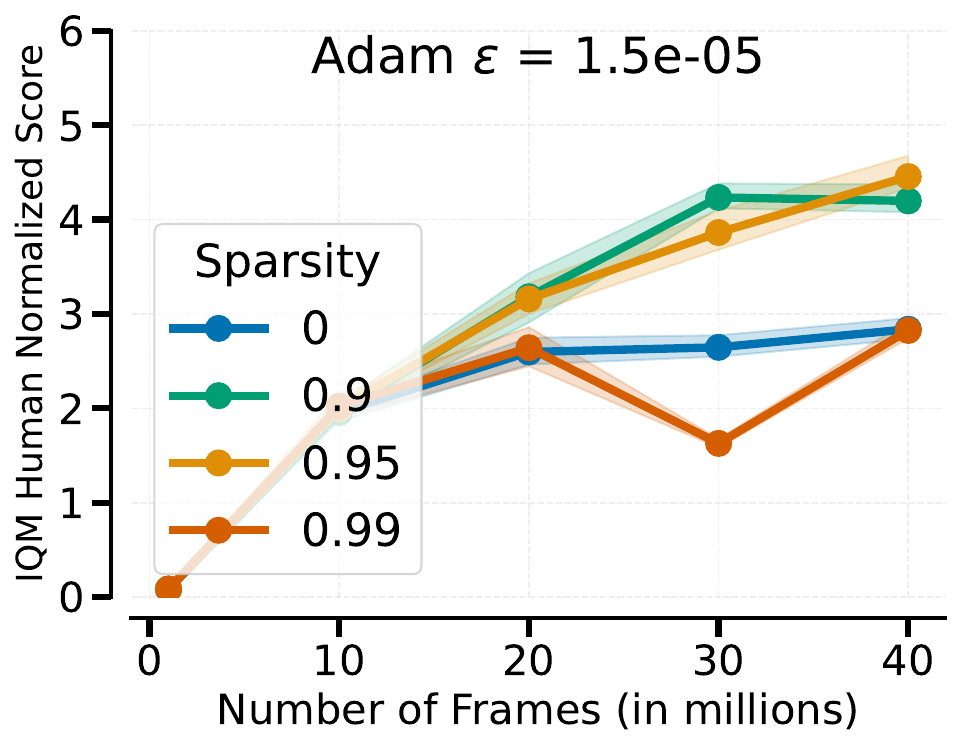}%
    \includegraphics[width=0.245\textwidth]{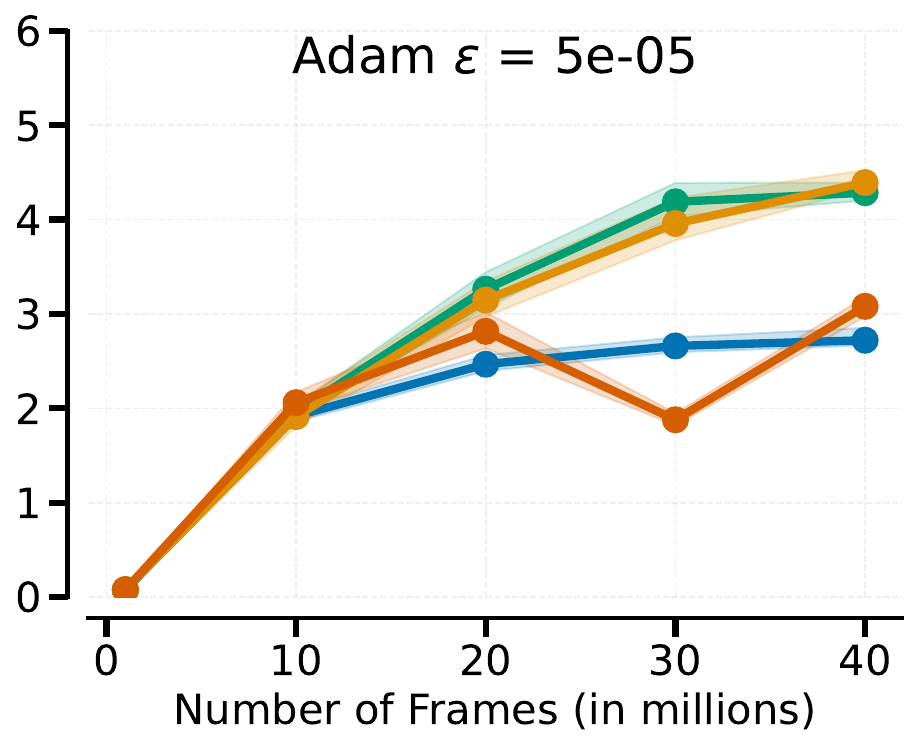}%
    \includegraphics[width=0.245\textwidth]{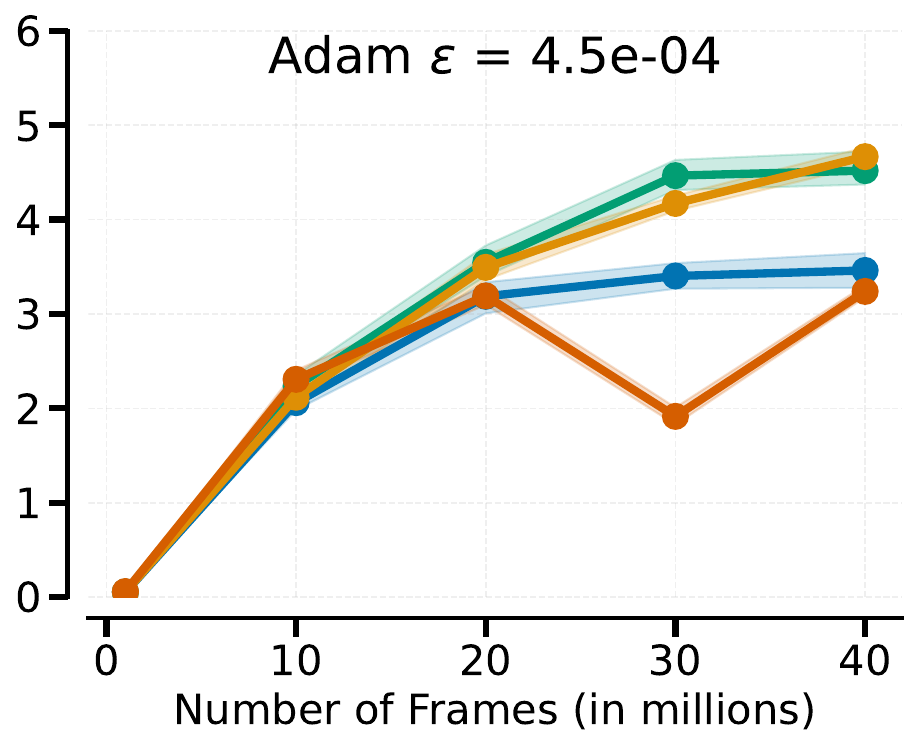}%
    %  \includegraphics[width=0.245\textwidth]{figures/DQN_wd3_epsilonD.pdf}%
    %\vspace{-0.1cm}
    \caption{\textbf{Adam's epsilon ($\epsilon$)} \citep{kingma2014adam} for DQN with ResNet architecture and a width multiplier of $3$. }
    \label{fig:}
    \vspace{-0.2cm}
\end{figure*}

\newpage

\subsection{Sweeping over ReDo threshold $\tau$}
\label{sec:redoSweep}

This parameter (introduced in Definition $3.1$ of \citep{sokar2023dormant}) defines the threshold for determining neuron dormancy. \citet{sokar2023dormant} suggested using 0.1 with the CNN network. Since we are using the Impala network architecture, we tested three additional values: ($0$, $0.025$, $0.3$). We found that $0.1$, as used in \autoref{fig:scalingWidths_learningCurves} of our submission, yields the best performance. 

\begin{figure}[!h]
    \centering
    \includegraphics[width=0.48\textwidth]{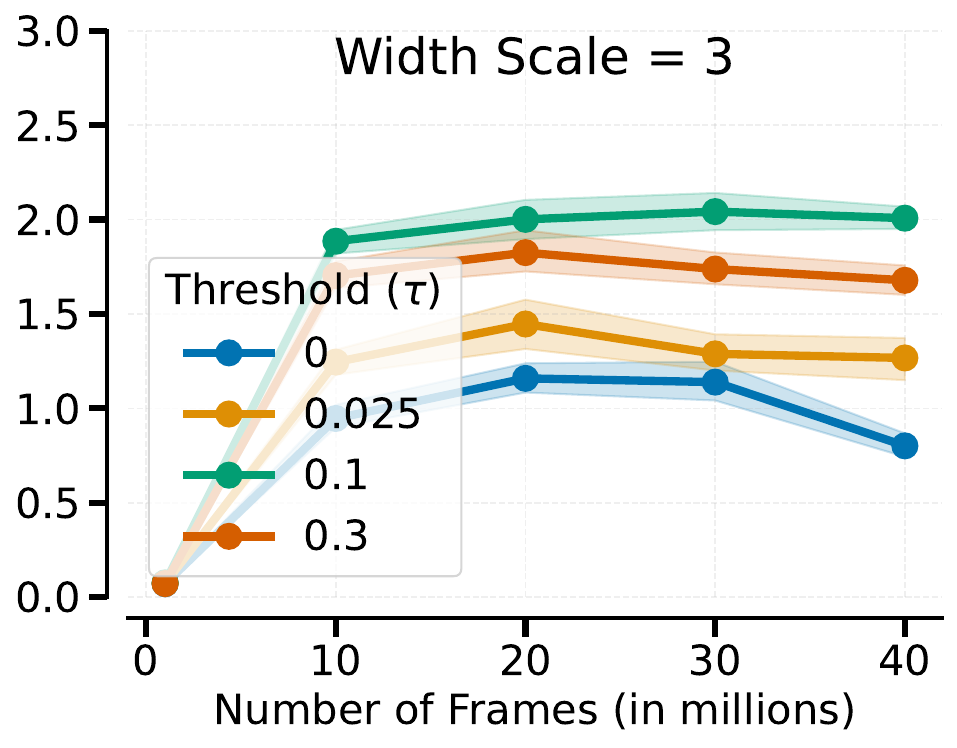}%
    %\vspace{-0.1cm}
    \caption{\textbf{Varying ReDo's $\tau$ threshold} \citep{sokar2023dormant} for DQN with ResNet architecture and a width multiplier of $3$.}
    \label{fig:}
    \vspace{-0.2cm}
\end{figure}

\subsection{Frequency of network resets}
\label{sec:freqResetSweep}

We varied the frequency of network resets \citep{nikishin22primacy} to evaluate whether this could help mitigate the performance loss when increasing the network width. While more infrequent resets (every $250000$ steps compared to the default value of $100000$) improves performance slightly, it still drastically under-performs with respect to the baseline and the pruning approach.

\begin{figure}[!h]
    \centering
    \includegraphics[width=0.48\textwidth]{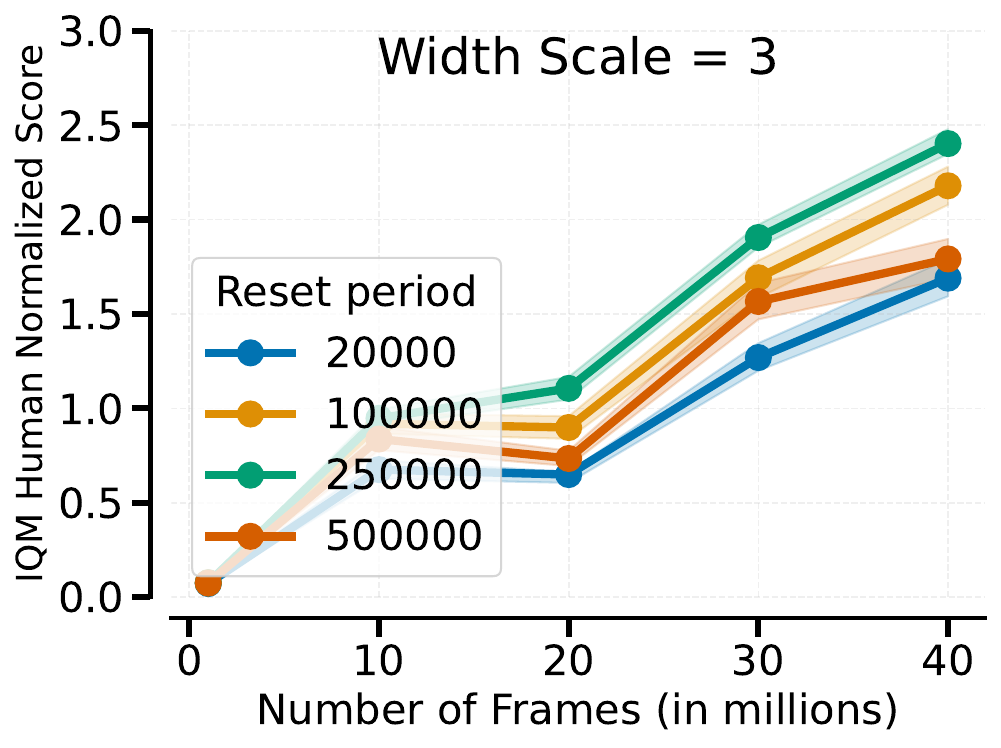}%
    %\vspace{-0.1cm}
    \caption{\textbf{Varying the reset period} \citep{nikishin22primacy} for DQN with ResNet architecture and a width multiplier of $3$.}
    \label{fig:}
    \vspace{-0.2cm}
\end{figure}

\newpage

\subsection{Layer to reset}
\label{sec:layerResetSweep}

In our paper we followed the approach of \citet{nikishin22primacy} and \citet{sokar2023dormant} of resetting only the last layer. We explored resetting different layers, but found it resulted in no significant performance difference.

\begin{figure}[!h]
    \centering
    \includegraphics[width=0.48\textwidth]{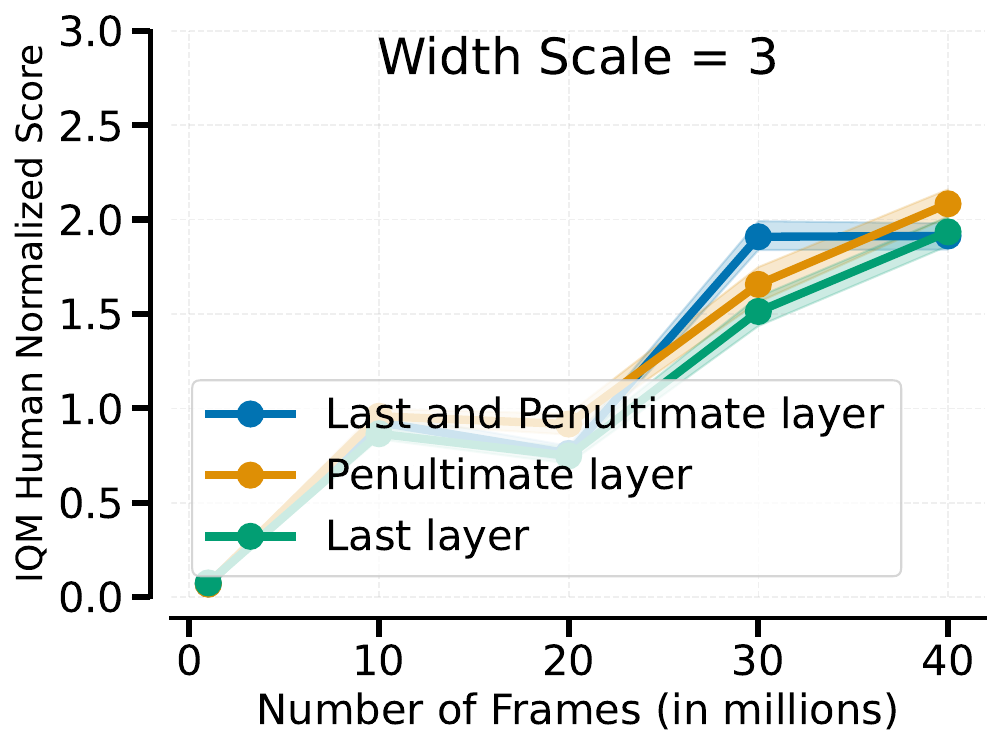}%
    %\vspace{-0.1cm}
    \caption{\textbf{Resetting different layers} \citep{nikishin22primacy}, DQN with ResNet architecture and a width multiplier of $3$.}
    \label{fig:}
    \vspace{-0.2cm}
\end{figure}

\subsection{Varying weight decay}
\label{sec:weightDecaySweep}

We ran a sweep over the following values $10e^{-6}$ , $10e^{-5}$ , $10e^{-4}$ , $10e^{-3}$, and $10e^{-2}$, using the Impala architecture with a width factor of $3$. The best performance is obtained with $10e^{-5}$, which is the value suggested by \citet{sokar2023dormant}, and the value used in \autoref{fig:scalingWidths_learningCurves} of our submission.

\begin{figure}[!h]
    \centering
    \includegraphics[width=0.48\textwidth]{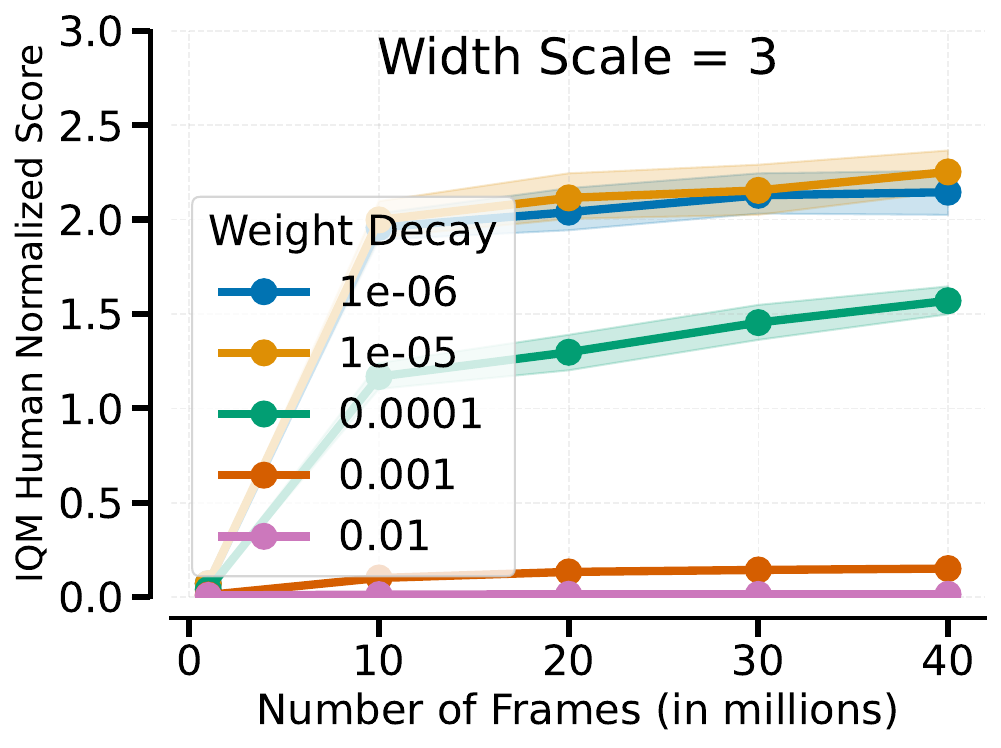}%
    %\vspace{-0.1cm}
    \caption{\textbf{Weight Decay (WD)} for DQN with ResNet architecture with a width factor of $3$.}
    \label{fig:}
    \vspace{-0.2cm}
\end{figure}

\newpage

\subsection{Varying learning rates}
\label{sec:learningRateSweep}

The default learning for DQN is $6.25e-5$. As suggested by the reviewer, we have run experiments with a learning rate divided by the width scale factor (so $2.08e-5$ for a width factor of $3$, and $1.25e-5$ for a width factor of $5$). These learning rates do improve the performance of the baseline, but it is still surpassed by pruning. These results are consistent with the thesis of the paper: \textit{pruning can serve as a drop-in mechanism for increasing agent performance.}

\begin{figure}[!h]
    \centering
    \includegraphics[width=0.42\textwidth]{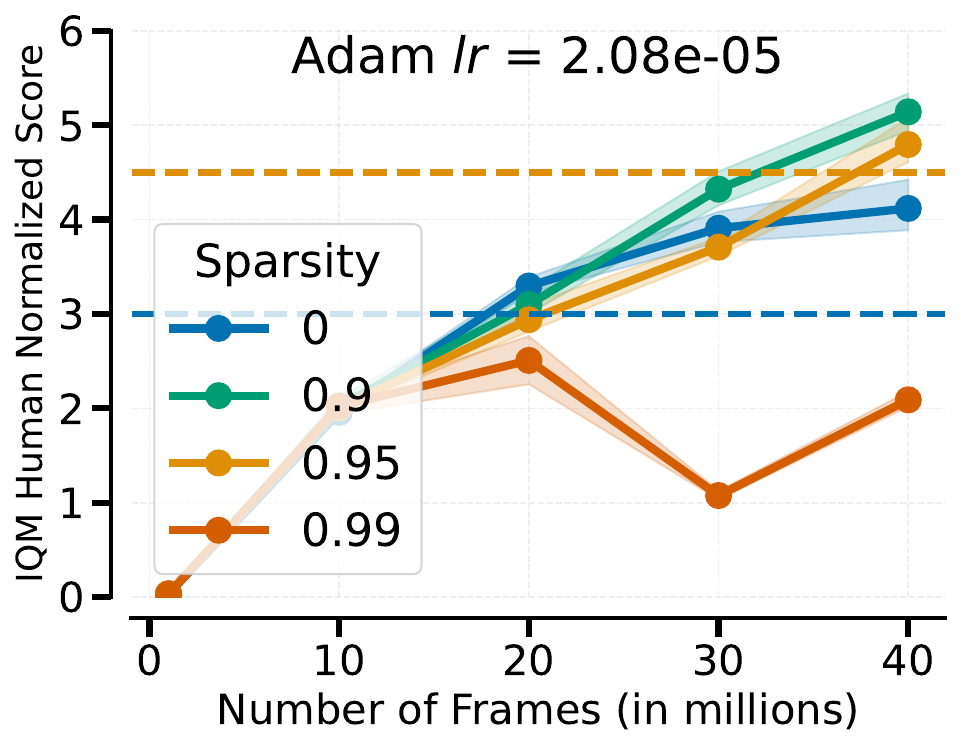}%
     \includegraphics[width=0.4\textwidth]{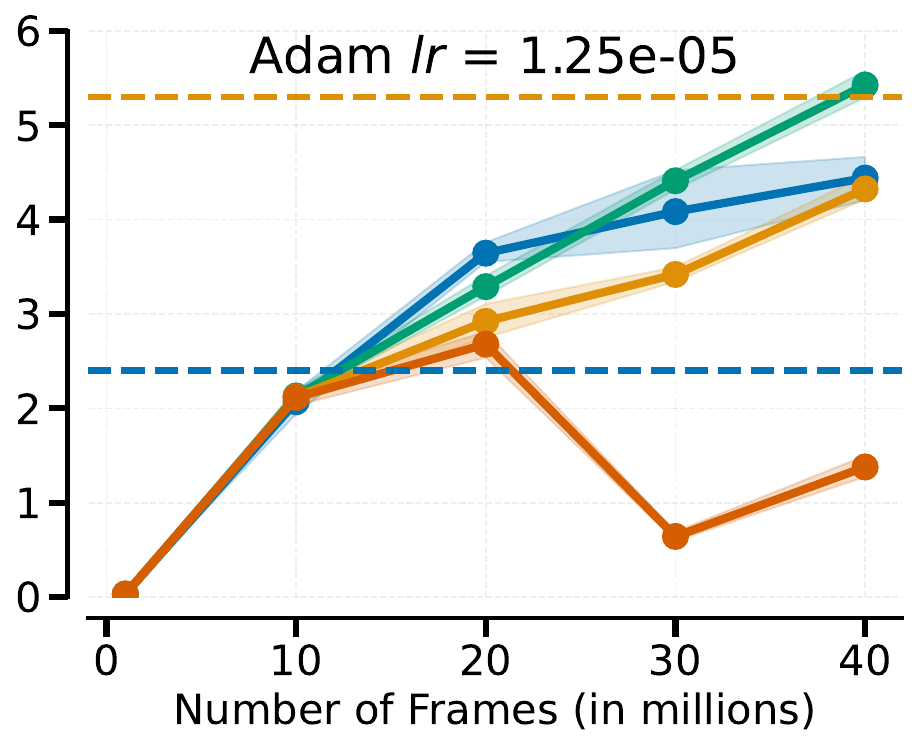}%
    \caption{\textbf{Learning rate evaluation} for DQN agent with ResNet architecture (with a width factor of $3$ {\bf left} and $5$ {\bf right}). These learning rate values correspond to dividing the default learning rate by the factor we used to amplify the size of the neural network. The dashed lines indicate the final performance for \textcolor{blue}{dense (blue)} and \textcolor{orange}{$0.95\%$ sparse (orange)} nets when using the default learning rate ( \textit{lr:} $6.25e-5$).}
    \label{fig:learning_rate}
    \vspace{-0.2cm}
\end{figure}

\subsection{Varying batch size}
\label{sec:batchSizeSweep}

The default batch size is $32$, and ran experiments with batch sizes of $16$ and $64$. In all cases, pruning maintains its strong advantage.

\begin{figure}[!h]
    \centering
    \includegraphics[width=0.4\textwidth]{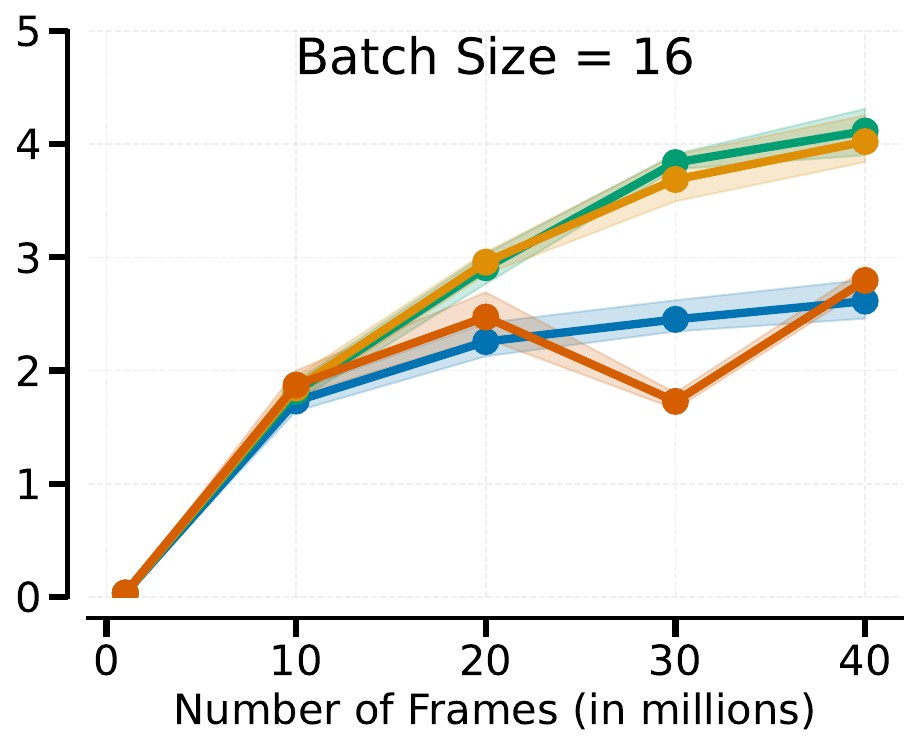}%
    \includegraphics[width=0.4\textwidth]{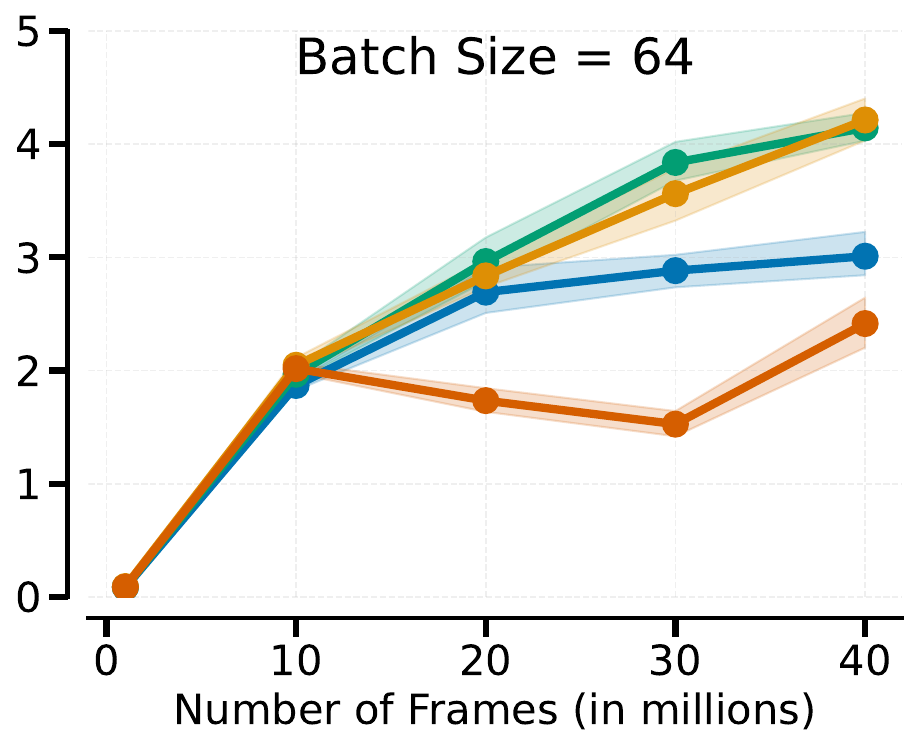}%
    %\vspace{-0.1cm}
    \caption{\textbf{Batch size} \citep{ceron2023small} for DQN with ResNet architecture and a width multiplier of $3$.}
    \label{fig:batch_size}
    \vspace{-0.2cm}
\end{figure}
}

\newpage
\subsection{Varying update horizon}
\label{sec:updateHorizonSweep}

We explored using an update horizon of $3$ for DQN (the default is $1$) and found that pruning still maintains its advantage.

\begin{figure}[!h]
    \centering
    \includegraphics[width=0.48\textwidth]{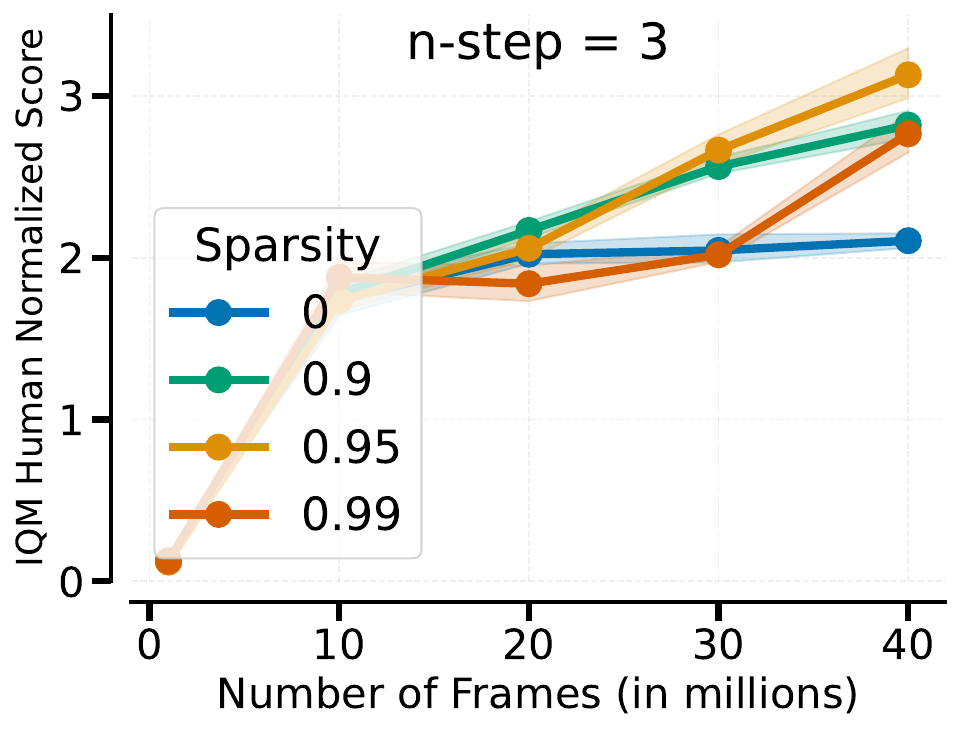}%
    %\vspace{-0.1cm}
    \caption{\textbf{Multi-step return} \citep{sutton88learning} for DQN with ResNet architecture and a width multiplier of $3$.}
    \label{fig:}
    \vspace{-0.2cm}
\end{figure}

% You can have as much text here as you want. The main body must be at most $8$ pages long.
% For the final version, one more page can be added.
% If you want, you can use an appendix like this one.  

% The $\mathtt{\backslash onecolumn}$ command above can be kept in place if you prefer a one-column appendix, or can be removed if you prefer a two-column appendix.  Apart from this possible change, the style (font size, spacing, margins, page numbering, etc.) should be kept the same as the main body.
%%%%%%%%%%%%%%%%%%%%%%%%%%%%%%%%%%%%%%%%%%%%%%%%%%%%%%%%%%%%%%%%%%%%%%%%%%%%%%%
%%%%%%%%%%%%%%%%%%%%%%%%%%%%%%%%%%%%%%%%%%%%%%%%%%%%%%%%%%%%%%%%%%%%%%%%%%%%%%%

\end{document}